%% file: article.tex
\documentclass[accepted]{uai2026}

\usepackage{siunitx}
\usepackage[acronym,nohypertypes={acronym}]{glossaries}
\usepackage{booktabs}

\usepackage{adjustbox}
\usepackage[capitalise]{cleveref}
\usepackage{multirow}
\usepackage{longtable}
\usepackage{placeins}
\crefname{figure}{Figure}{Figures}
\Crefname{figure}{Figure}{Figures}

\usepackage[american]{babel}
\usepackage{natbib}
\usepackage{amssymb}

\usepackage{mathtools}

\newacronym{scm}{SCM}{Structural Causal Model}
\newacronym{ate}{ATE}{Average Treatment Effect}
\newacronym{tabpfn}{TabPFN}{Tabular Prior-Data Fitted Network}
\newacronym{pc}{PC}{Peter-Clark}
\newacronym{dag}{DAG}{Directed Acyclic Graph}
\newacronym{cpdag}{CPDAG}{Completed Partially Directed Acyclic Graph}
\newacronym{tvd}{TVD}{Total Variation Distance}
\newacronym{kmtvd}{kMTVD}{$k$-Marginal Total Variation Distance}
\newacronym{kci}{KCI}{Kernel Conditional Independence}
\newacronym{ks}{KS}{Kolmogorov-Smirnov}
\newacronym{nnaa}{NNAA}{Nearest-Neighbor Adversarial Accuracy}
\newacronym{tabularargn}{TabularARGN}{Tabular Auto-Regressive Generative Network}
\newacronym{realtabformer}{REaLTabFormer}{Realistic Relational
and Tabular Transformer}
\newacronym{dptbart}{DP-TBART}{Differentially-Private TaBular AutoRegressive Transformer}
\newacronym{decaf}{DECAF}{DEbiasing CAusal Fairness}
\newacronym{gan}{GAN}{Generative Adversarial Network}
\newacronym{cmd}{CMD}{Correlation Matrix Difference}
\newacronym{dpsgd}{DP-SGD}{Differentially Private Stochastic Gradient Descent}
\newacronym{t1dm}{T1DM}{Type \num{1} Diabetes Mellitus}
\newacronym{vae}{VAE}{Variational Autoencoder}
\newacronym{tgan}{TGAN}{Table GAN}
\newacronym{ctgan}{CTGAN}{Conditional Tabular GAN}
\newacronym{cart}{CART}{Classification And Regression Trees}
\newacronym{fda}{FDA}{Food and Drug Administration}
\newacronym{opdag}{oracle-PDAG}{oracle partially directed acyclic graph}
\newacronym{pdag}{PDAG}{partially directed acyclic graph}

\newacronym{clb}{CLB}{CSuite Large Backdoor}
\newacronym{cmc}{CMC}{CSuite Mixed Confounding}
\newacronym{cms}{CMS}{CSuite Mixed Simpson}
\newacronym{cns}{CNS}{CSuite Nonlin Simpson}
\newacronym{css}{CSS}{CSuite Symprod Simpson}
\newacronym{cwa}{CWA}{CSuite Weak Arrows}
\newacronym{csm}{CSM}{Custom SCM}
\newacronym{sgl}{SGL}{Simglucose}

\glsdisablehyper

\title{Improving TabPFN's Synthetic Data Generation by Integrating Causal Structure}

\author[1,2]{Davide~Tugnoli}
\author[3]{Andrea~{De Lorenzo}}
\author[4]{Marco~Virgolin}
\author[5,6]{Giovanni~Cin\`{a}}

\affil[1]{%
    Department of Mathematics, Informatics and Geosciences\\
    University of Trieste, Trieste, Italy
}
\affil[2]{%
    InSilicoTrials Technologies\\
    Riva Grumula 2, 34123 Trieste, Italy
}
\affil[3]{%
    Department of Engineering and Architecture\\
    University of Trieste, Trieste, Italy
}
\affil[4]{%
    InSilicoTrials Technologies BV\\
    The Netherlands
}
\affil[5]{%
    Medical Informatics Dept., Amsterdam UMC\\
    The Netherlands
}
\affil[6]{%
    Institute for Logic, Language and Computation\\
    University of Amsterdam, The Netherlands
}
\begin{document}
\sloppy
\maketitle

\begin{abstract}
Synthetic tabular data generation addresses data scarcity and privacy constraints in a variety of domains.
\gls{tabpfn}, a recent foundation model for tabular data, has been shown capable of generating high-quality synthetic tabular data.
However, \gls{tabpfn} is autoregressive: features are generated sequentially by conditioning on the previous ones, depending on the order in which they appear in the input data.
We demonstrate that when the feature order conflicts with causal structure, the model produces spurious correlations that impair its ability to generate synthetic data and preserve causal effects. We address this limitation by integrating causal structure into \gls{tabpfn}'s generation process through two complementary approaches: \gls{dag}-aware conditioning, which samples each variable given its causal parents, and a \gls{pdag}-based strategy for scenarios with partial causal knowledge. We evaluate these approaches on controlled benchmarks and six CSuite datasets, assessing structural fidelity, distributional quality, and \gls{ate} preservation. Across most settings, \gls{dag}-aware conditioning improves the quality and stability of synthetic data relative to vanilla \gls{tabpfn}. Under partial causal knowledge, the \gls{opdag}, which orients only the edges into the colliders, shows moderate gains, while the benefit of a \gls{cpdag} discovered from data depends on how well the causal structure is recovered. These results indicate that reliable causal structure, even partial, can be injected into \gls{tabpfn} at inference time, without parameter updates, to improve synthetic data quality.
\end{abstract}
\glsresetall


\section{Introduction}

Synthetic tabular data generation is becoming an important approach to handle data scarcity and privacy concerns in many domains~\citep{shi2025comprehensive}. This is especially relevant in fields such as healthcare, finance, and policy research, where privacy regulations and ethical constraints often limit access to real data. In pharmaceutical research, for example, synthetic data can be used to simulate drug effects for safety and efficacy, while protecting patient confidentiality. For tabular data, generation methods must avoid producing close copies of training samples, especially when data are limited.

However, generating reliable synthetic tabular data is challenging because real datasets usually contain complex causal relations among variables. Generation methods that ignore such dependencies may create spurious correlations that differ from the true data-generating process. When synthetic data are used to augment limited real data or simulate clinical scenarios, misleading dependencies can propagate to treatment effect estimates. In drug development, for instance, inaccurate estimation of treatment effects from flawed synthetic data could lead to costly trials on ineffective drugs or the exclusion of promising ones.

Recent foundation models for tabular data, such as \gls{tabpfn}~\citep{hollmann2025accurate}, have shown promising results by pre-training on millions of synthetic datasets derived from \glspl{scm}~\citep{pearl2009causality}. Pre-training on such diverse datasets is particularly valuable when real data are limited, a common scenario in domains like healthcare. This advantage becomes even more important for synthetic data generation, which requires predicting all features rather than a single target variable as in traditional classification or regression tasks. Extensions of \gls{tabpfn}\footnote{\url{https://github.com/PriorLabs/tabpfn-extensions}} enable unsupervised, autoregressive generation of synthetic tabular data. However, \gls{tabpfn}'s autoregressive approach generates variables sequentially without explicitly accounting for causal structure. Although the model averages predictions over random permutations of the conditioning set, this mitigates but does not eliminate order sensitivity: during generation, each variable can only condition on features that appear earlier in the sequence. When the generation order conflicts with causal dependencies, particularly in the presence of colliders, the model may introduce spurious correlations that compromise the reliability of synthetic data.

This work studies the autoregressive nature of \gls{tabpfn}-based synthetic data generation, identifies a fundamental limitation, and proposes methods to address it. Our contributions are as follows:
\begin{enumerate}
    \item We demonstrate that \gls{tabpfn} synthetic data quality depends heavily on the order of the features in the input data due to the absence of causal reasoning, with this sensitivity persisting even at large training sizes.
    \item We propose causal conditioning strategies that leverage known, partially known, or discovered causal structure at inference time, without parameter updates. We show improvements over vanilla \gls{tabpfn} for both full \gls{dag} knowledge and the \gls{opdag} across distributional quality metrics, while structure discovered from data has no consistent effect and its potential benefit depends on how well the causal structure is recovered.
    \item We quantify how synthetic data errors propagate to treatment effect preservation, demonstrating substantial misestimation that could lead to incorrect decisions in applications such as drug development.
\end{enumerate}
All code, datasets, and detailed numerical results are publicly available.\footnote{\url{https://github.com/DavideTugnoli/tabpfn-causal-synthetic}}

\section{Related Work}
Generative approaches to synthetic tabular data range from classical statistical methods to deep models based on \glspl{gan}, \glspl{vae}, diffusion models, and autoregressive architectures~\citep{shi2025comprehensive}. 

Early deep generative models adapted architectures from image generation. TGAN~\citep{xu2018synthesizing} applied \glspl{gan} to mixed continuous--categorical tables using Long Short-Term Memory-based generation, while CTGAN~\citep{xu2019modeling} addressed class imbalance and multimodal distributions. CTAB-GAN~\citep{zhao2021ctab} further improved semantic fidelity. More recently, diffusion models have achieved strong performance~\citep{kotelnikov2023tabddpm,zhang2024mixed,shi2025tabdiff}. While effective for statistical fidelity, these methods generate all features simultaneously without explicit causal structure.

Autoregressive approaches factorize the joint distribution into sequential conditionals $p(x_j \mid x_{C(j)})$, where $C(j)$ denotes the conditioning set. Early work used \gls{cart} to generate features sequentially~\citep{reiter2005using}. GReaT treats tabular rows as text sequences for language model generation~\citep{borisov2023language}. TabularARGN~\citep{tiwald2025tabularargn} randomizes feature orderings during training to learn conditionals across different subsets. Among foundation models, \gls{tabpfn}~\citep{hollmann2022tabpfn, hollmann2025accurate} achieves state-of-the-art performance on small datasets via in-context learning. Pre-trained on millions of \gls{scm}-sampled datasets, it provides strong inductive biases without target data retraining. TabPFGen extends TabPFN to generative tasks via energy-based sampling rather than autoregressive conditioning~\citep{ma2024tabpfgen}, but neither approach explicitly conditions on causal graphs during generation.

Causal approaches integrate structural knowledge directly into generation. DECAF embeds \glspl{scm} into the architecture for interventional debiasing~\citep{van2021decaf}. Causal-TGAN structures its generator according to a causal graph~\citep{wen2022causal}. DATGAN incorporates expert \glspl{dag} to enforce structural dependencies~\citep{lederrey2022datgan}. CA-GAN combines causal discovery with reinforcement learning~\citep{nguyen2025causal}. CausalDiffTab is a diffusion model that learns a causal graph from the data and uses it to regularize generation~\citep{zhang2025causaldifftab}. Unlike TabPFN, these models are trained from scratch per dataset and explicitly use \gls{dag} structure to guide generation.

Recent work has applied transformer-based meta-learning to causal effect estimation.
Do-PFN~\citep{robertson2025pfn} adapts the PFN framework to predict conditional interventional distributions from observational data, taking covariates and treatment as input.
CausalPFN~\citep{balazadeh2025causalpfn} trains a transformer on a causal prior to estimate expected potential outcomes, also conditioning on covariates and treatment.
Both models predict outcomes under intervention but do not model the full joint distribution over all variables.
MACE-TNP~\citep{dhir2025estimating} addresses a related task, estimating Bayesian model-averaged interventional distributions from observational data via neural processes; it similarly targets interventional queries rather than joint density modeling.
To our knowledge, this is the first work to combine a foundation model with explicit causal structure for synthetic tabular data generation, addressing autoregressive ordering violations and collider bias through DAG-aware and PDAG-based conditioning strategies.

\section{Methodology}

\subsection{TabPFN Causal Extensions}
\label{sec:tabpfn-extensions}
\Gls{tabpfn} is a transformer-based model pre-trained on approximately \num{130} million synthetic datasets, each sampled from a randomly generated \gls{scm} with diverse graph structures, functional relationships, and noise distributions~\citep{hollmann2025accurate}. At inference time, it receives an entire dataset as context and produces predictions in a single forward pass via in-context learning, without retraining on the target data. For synthetic data generation, the model is applied autoregressively: features are generated sequentially following the column order of the input data.
Denoting this ordering as~$\pi$, each feature $x_{\pi(i)}$ at position $i$ in the sequence is sampled from a conditional distribution
\begin{equation}
p\!\left(x_{\pi(i)} \mid \mathcal{C}(\pi(i))\right),
\label{eq:vanilla}
\end{equation}
where the conditioning context $\mathcal{C}(\pi(i)) = \{x_{\pi(0)}, \dots, x_{\pi(i-1)}\}$ consists of all features preceding position~$i$ in the ordering.
When no context is available ($i=0$), the model conditions on random noise to approximate the marginal distribution.

\citet{hollmann2025accurate} recognize that the autoregressive nature induces a specific bias due to the local ordering of variables within the conditioning set $\mathcal{C}(\pi(i))$.
To mitigate this issue, they propose averaging predictions over multiple random permutations of the conditioning set at inference time.
However, this averaging strategy does not address the limitation imposed by the global ordering~$\pi$.
When~$\pi$ does not respect the causal structure, the model may still condition on descendant variables when generating their ancestors, inducing spurious correlations.

For instance, consider a collider structure $X \to Z \leftarrow Y$, where the causes $X$ and $Y$ are marginally independent. 
If the global ordering places the common effect $Z$ before its parents, the model may, e.g., generate $X$ conditioned on $Z$, then $Y$ conditioned on both $Z$ and $X$.
Conditioning on the collider $Z$ makes $X$ and $Y$ conditionally dependent. Sampling from exact conditionals would still recover their marginal independence, but estimation errors in the learned conditionals propagate this dependence to the synthetic data, introducing spurious correlations.

To address this limitation, we introduce causal conditioning strategies that align \gls{tabpfn}'s generative process with known or partially known causal structures.
These strategies act only on \gls{tabpfn}'s input and differ only in the conditioning set. For each variable, we present the model with the target column and its conditioning columns and query it through the standard in-context interface, leaving the pre-trained network unchanged.
Causal relationships between variables can be formally represented using a \gls{dag} $\mathcal{G}=(V,E)$, where nodes $V$ correspond to variables and directed edges $E$ indicate direct causal influences~\citep{pearl2009causality}.
When such a graph is available, we condition each variable only on its causal parents rather than on all previously generated features:
\begin{equation}
\mathcal{C}_{\mathrm{DAG}}(x_{\pi(i)}) 
= \{ x_j : x_j \to x_{\pi(i)} \text{ in } \mathcal{G} \}.
\label{eq:dag}
\end{equation}
Variables are generated following a topological ordering of~$\mathcal{G}$, ensuring that all parents have been generated before each child.\footnote{A topological ordering alone does not imply DAG-aware conditioning: the vanilla strategy of Eq.~\eqref{eq:vanilla} can also follow a topological ordering, but conditions on all predecessors rather than on causal parents only.}

However, in most real-world scenarios, the full \gls{dag} is unknown, and a causal discovery algorithm can recover only a partial graph from data.
Such a graph can be represented by a \gls{cpdag}~\citep{chickering2002optimal}, which includes directed edges where the orientation is uniquely determined and undirected edges where multiple orientations remain compatible with the data. The discovered \gls{cpdag} represents a realistic scenario, though its recovered orientations may contain errors. As a correct-by-construction reference, we also evaluate the \gls{opdag}, which orients only the edges into the true \gls{dag}'s colliders. Derived directly from the true graph, it isolates the effect of these orientations from finite-sample discovery errors.

Due to the partially directed structure of either partial graph, Equation~\eqref{eq:dag} cannot be readily applied.
We define a generation ordering~$\sigma$ that places variables with known causal parents before the remaining ones, and propose the following hybrid conditioning strategy, applied to both graphs by letting $\mathcal{G}^\ast$ denote the discovered \gls{cpdag} or the \gls{opdag}:
\begin{equation}
\mathcal{C}(x_{\sigma(i)}) \!=\!
\begin{cases}
\mathrm{pa}_{\mathcal{G}^\ast}(x_{\sigma(i)}) 
  & \text{if fully directed,} \\[4pt]
\{x_{\sigma(0)}, \dots, x_{\sigma(i\!-\!1)}\} 
  & \text{otherwise,}
\end{cases}
\label{eq:cpdag}
\end{equation}
where ``fully directed'' means that $x_{\sigma(i)}$ has at least one directed adjacent edge and no undirected adjacent edges in~$\mathcal{G}^\ast$. Variables satisfying this condition are generated according to their causal parents, while the rest revert to sequential conditioning on all predecessors in the ordering. When every edge in $\mathcal{G}^\ast$ is directed, Eq.~\eqref{eq:cpdag} recovers the conditioning set of Eq.~\eqref{eq:dag} for all non-isolated variables; when no edge is directed, it reduces to the conditioning set of Eq.~\eqref{eq:vanilla}.
Because the ordering places oriented parents before their children, this fallback still conditions on them. It drops only the restriction of conditioning on the parents alone, without assuming an orientation for the undirected edges.

\section{Experimental Design}

\subsection{Evaluation Metrics and Controls}
We evaluate synthetic data quality using metrics widely adopted in benchmark frameworks such as SynthEval~\citep{lautrup2025syntheval} and the Differential Privacy Synthetic Data Challenge organized by the National Institute of Standards and Technology~\citep{ridgeway2021challenge}. We focus on dependency structure, measured by the \gls{cmd}, and use the \gls{kmtvd} as a robustness check on pairwise fidelity. We additionally report the \gls{nnaa}~\citep{yale2020generation}, a nearest-neighbor resemblance metric used in privacy-oriented evaluations of synthetic data, which measures how distinguishable synthetic and real data are, defined in \cref{sec:nnaa_definition} and reported with each comparison in the appendix.
The \gls{cmd} quantifies how well the overall pairwise dependency structure among variables is preserved.
We compute mixed correlation matrices combining Cramér's~$V$ for categorical--categorical pairs, the correlation ratio~$\eta$ for categorical--numerical pairs, and Spearman's rank correlation for numerical--numerical pairs. We replace Pearson (the SynthEval default) with Spearman to capture monotonic relationships in datasets with nonlinear dependencies.
The metric is computed as the Frobenius norm of the difference between real and synthetic correlation matrices, i.e., $\mathrm{CMD}=\lVert C_{\text{real}}-C_{\text{synthetic}}\rVert_F$.

The \gls{kmtvd} with $k=2$ measures pairwise distributional fidelity. 
Continuous variables are discretized into $B=\num{20}$ quantile-based bins.
For two empirical distributions $P$ and $Q$, the total variation distance is $\mathrm{TVD} = \tfrac{1}{2}\sum_x \lvert P(x)-Q(x)\rvert$, and the metric is the mean $\mathrm{TVD}$ across all variable pairs.

We assess the statistical significance of differences between conditioning strategies using the Wilcoxon signed-rank test with Pratt's method for handling tied observations, applying Holm correction for prespecified comparisons. Effect sizes are quantified using the Hodges--Lehmann estimator, the median of pairwise averages of differences~\citep{hodges1963estimates}.

\subsection{Datasets}

We conduct experiments on three dataset classes, ranging from fully controlled hand-crafted settings to public benchmark datasets and realistic clinical scenarios (\Cref{tab:datasets}).

\begin{table*}[!htbp]
\centering
\caption{Overview of the datasets used in our experiments, from controlled hand-crafted \glspl{scm} to CSuite benchmarks and a realistic clinical simulator.}
\label{tab:datasets}
\begin{tabular}{llcll}
\toprule
Dataset & Abbr. & Vars. & Structural role & Causal info \\
\midrule
Custom collider SCM & CSM & 4 & Collider & Known true DAG \\
CSuite Nonlin Simpson & CNS & 4 & Simpson's paradox, continuous & Known true DAG \\
CSuite Symprod Simpson & CSS & 4 & Nonlinear product interaction & Known true DAG \\
CSuite Mixed Simpson & CMS & 4 & Simpson's paradox, mixed-type & Known true DAG \\
CSuite Mixed Confounding & CMC & 12 & Complex confounding, mixed-type & Known true DAG \\
CSuite Large Backdoor & CLB & 9 & Backdoor paths & Known true DAG \\
CSuite Weak Arrows & CWA & 9 & Weak causal effects & Known true DAG \\
Simglucose & SGL & 38 & Physiological (static T1DM) & Partial (full DAG unknown) \\
\bottomrule
\end{tabular}
\end{table*}

\subsubsection{Custom Collider SCM}
We design a four-variable \gls{scm} containing a collider to evaluate \gls{tabpfn}'s sensitivity to causal structure under fully controlled conditions.
Colliders (variables with multiple parent edges) pose a fundamental challenge for autoregressive generators because they induce conditional dependencies between otherwise independent variables.

The DAG of this SCM is defined as $X_0 \to X_1 \leftarrow X_2 \leftarrow X_3$, where $X_1$ is the collider node. 
Although $X_0$ and $X_2$ are marginally independent ($X_0 \perp\!\!\!\perp X_2$), conditioning on their common child $X_1$ induces dependence between them ($X_0 \not\!\perp\!\!\!\perp X_2 \mid X_1$). 
When an autoregressive generator conditions on $X_1$ while generating its parents, it correctly learns that $X_0$ and $X_2$ are dependent given $X_1$. However, sequential sampling from these conditionals propagates this dependence into the marginal distribution, creating spurious marginal correlation between variables that should be independent.
We construct the \gls{scm} with near-deterministic linear relationships to amplify this collider bias effect and enable clear detection of spurious associations introduced by inappropriate conditioning strategies. Full structural equations appear in \cref{sec:appendix-custom-scm}.

\subsubsection{CSuite Benchmark Datasets}
Next, we consider the Microsoft CSuite benchmark~\citep{geffner2022deep}, a collection of hand-crafted \glspl{scm} designed for evaluating causal discovery and inference under known ground-truth graphs.

We select six datasets with at least four nodes and distinct structures, covering a range of causal scenarios: 
\gls{cns} and \gls{cms} (four nodes each) are variants of Simpson's paradox with continuous or mixed-type variables; \gls{css} (four nodes) is a nonlinear benchmark with a product interaction.
\gls{cmc} (twelve nodes) combines mixed data types and complex confounding structures, while \gls{clb} and \gls{cwa} (nine nodes each) have large topologies with backdoor paths and weak causal effects, respectively.
Each dataset includes predefined interventions for treatment effect estimation.

\subsubsection{Simglucose Type 1 Diabetes Mellitus Simulator}
Finally, we evaluate on a realistic dataset derived from the UVA/Padova \gls{t1dm} Simulator\footnote{\url{https://github.com/jxx123/simglucose}}~\citep{doi:10.1177/193229680900300107}, a \gls{fda}-approved model of glucose-insulin dynamics.
The original simulator integrates a system of ordinary differential equations over time.
Following the approach in~\citet{pmlr-v275-esponera25a}, we use an acyclic reformulation in a static setting by removing temporal dependencies.

The dataset contains \num{38} variables, including physiological states, patient-specific parameters, and exogenous actions.
Although the physiological equations define a clear causal progression among the endogenous states, the relations among patient-specific parameters are not known, i.e., we do not know the full DAG.
These parameters are sampled from distributions based on clinical data and represent heterogeneous virtual patients.
Because only partial causal knowledge is available, we test whether partial topological ordering still improves generation quality.

\subsection{Synthetic Data Quality}
To ensure reproducibility, we conduct all experiments using the official implementation of \gls{tabpfn} (version~2.1.0).
We compare three conditioning strategies to assess how causal structure affects synthetic data quality: \emph{vanilla} \gls{tabpfn} (which conditions sequentially on predecessors, Equation~\eqref{eq:vanilla}), \gls{dag}-aware generation (which conditions on causal parents), and \gls{pdag}-based generation (which combines directed and undirected causal relationships).
We conduct experiments on the \gls{csm}, the six CSuite benchmark datasets, and \gls{sgl}, using the default value of \num{3} permutations for the internal averaging mechanism described in Section~\ref{sec:tabpfn-extensions}.

To assess robustness across different data availability scenarios, we test multiple training sizes $N \in \{20, 50, 100, 200, 500\}$\footnote{$N$ denotes the number of samples provided to TabPFN as in-context learning data.} with \num{100} resampling iterations per configuration. 
We strictly separate training and test data by creating fixed test sets of \num{2000} samples for each dataset. 
For the \gls{csm} and \gls{sgl}, we generate \num{6000} samples total, select test samples once using a fixed random seed, and sample training data from the remaining \num{4000} samples. 
For CSuite benchmarks, we use the predefined test sets and sample training data from the remaining pool.

To study the susceptibility of TabPFN-based data generation to the input ordering, we evaluate three column orderings: \emph{original ordering} (as provided in the dataset, or a random permutation if it coincides with topological or reverse topological), \emph{topological ordering} (parents before children), and \emph{reverse topological ordering} (children before parents), which maximizes causal violations, illustrating a potential worst-case scenario. We use the original ordering as the default-use baseline, since it reflects how the model is typically applied without causal knowledge, and verify its robustness to random column permutations in \Cref{sec:appendix_random_order}.

For \gls{pdag}-based generation, we consider two settings: a realistic one, where \glspl{cpdag} are \emph{discovered} from the training data via the order-independent \gls{pc}-stable algorithm ($\alpha=0.05$)~\citep{colombo2014order}; and an ablation with full prior knowledge, the \gls{opdag}, which orients only the edges into the colliders of the true \gls{dag} and leaves the other edges undirected. The latter applies no further orientation rules (e.g., Meek's rules~\citep{meek1995causal}), so in general it is not a \gls{cpdag}.
For the custom \gls{scm}, we apply the Fisher--Z conditional independence test to assess edge dependencies within the \gls{pc} algorithm. 
For CSuite datasets with categorical or mixed-type variables, we use a hybrid strategy: \gls{kci} test for continuous pairs, $G^2$ test for discrete pairs, and $k$-Nearest Neighbors Conditional Mutual Information ($k=5$, \num{500} permutations) for mixed pairs.
For all strategies, we generate synthetic datasets of the same size as the test sets and evaluate them using \gls{cmd}, \gls{kmtvd}, and \gls{nnaa}.

\subsection{Treatment Effect Preservation}

We evaluate how well synthetic data preserve causal effects by measuring \gls{ate} preservation.
This addresses a practical scenario: given a small interventional dataset (e.g., from a pilot trial), we generate synthetic data and assess whether the treatment effect is preserved.

For all datasets, we construct fixed balanced test sets by sampling equally from both intervention arms with a fixed random seed, using generated interventional data for \gls{csm} and \gls{sgl} and benchmark-provided data for CSuite.
As both arms are sampled under the intervention, the \gls{ate} is read directly as the difference in their mean outcomes. For \gls{dag}-aware and \gls{pdag}-based generation, we apply a do-surgery disconnecting the treatment variable from its parents.

We test training sizes $N \in \{20, 50, 100, 200, 500, 1000\}$ with \num{100} resampling iterations per configuration, extending to larger sizes as effect estimation generally benefits from more samples. 
For each training size, we construct the training set by combining equal numbers of samples from both intervention arms into a single dataset, which is then provided to \gls{tabpfn} as in-context data.

The ground-truth \gls{ate} on the test set is computed as $\text{ATE}_{\text{test}} = \mathbb{E}[Y \mid \text{do}(X = x_1)] - \mathbb{E}[Y \mid \text{do}(X = x_0)]$, where $x_1$ and $x_0$ denote the two prespecified intervention values and $Y$ the predefined outcome variable for each dataset.
Synthetic data are then generated using the three conditioning strategies, and the corresponding \gls{ate} on the synthetic data is computed as $\text{ATE}_{\text{synthetic}} = \mathbb{E}[\hat{Y} \mid \text{do}(X = x_1)] - \mathbb{E}[\hat{Y} \mid \text{do}(X = x_0)],$ where the expectation here is taken over the synthetic data. Our evaluation metric is the absolute \gls{ate} difference, $\Delta_{\text{ATE}} = \left| \text{ATE}_{\text{test}} - \text{ATE}_{\text{synthetic}} \right|.$
Lower values indicate better preservation of causal effects. In comparative analyses, we test whether one method produces significantly lower absolute errors than another.

\section{Results}

\subsection{Synthetic Data Quality}

\paragraph{Topological Column Ordering Improves Vanilla TabPFN}
Vanilla \gls{tabpfn} is sensitive to feature ordering across all datasets, with sensitivity decreasing at larger training sizes (\cref{fig:order_sensitivity_combined}).
Topological ordering yields significant improvements in \gls{cmd} for \num{26} out of \num{40} combinations of dataset and training size (\Cref{fig:forest_vanilla_topological_ordering}), with \num{3} degradations on \gls{css} at training sizes $N \leq 100$.
Datasets \gls{csm} and \gls{sgl} show the greatest improvements in \gls{cmd}: largest at small training sizes for \gls{csm}, and increasing with training size for \gls{sgl}.
The \gls{kmtvd} and \gls{nnaa} results are consistent with these trends (\cref{fig:forest_vanilla_topo_kmtvd_nnaa}).
\Cref{tab:spurious_correlations_main} reports correlation coefficients for variable pairs that should be independent in \gls{csm}, confirming that vanilla \gls{tabpfn} introduces spurious correlations up to $|\rho| \approx 0.15$ while \gls{dag}-aware generation does not.

\begin{table}[!htbp]
\centering
\caption{Mean Pearson correlation coefficients for independent variable pairs in the custom collider \gls{scm}, at the smallest and largest training sizes. Values close to zero indicate correct independence preservation; the best value per column and training size is in \textbf{bold}. Standard deviations across \num{100} repetitions in parentheses. Intermediate training sizes are reported in \Cref{tab:spurious_correlations}.}
\label{tab:spurious_correlations_main}
\begin{tabular}{l r@{\,}l r@{\,}l}
\toprule
Method & \multicolumn{2}{c}{$\rho(X_0, X_3)$} & \multicolumn{2}{c}{$\rho(X_0, X_2)$} \\
\midrule
\multicolumn{5}{l}{\textit{Train size $N = 20$}} \\
Vanilla original & -0.149 & (0.203) & -0.150 & (0.202) \\
Vanilla topological & -0.018 & (0.110) & -0.018 & (0.109) \\
Vanilla reverse top. & -0.127 & (0.195) & -0.123 & (0.197) \\
\gls{dag}-aware & \textbf{0.004} & (0.021) & \textbf{0.004} & (0.021) \\
\Gls{opdag} & -0.018 & (0.109) & -0.018 & (0.110) \\
Discovered \gls{cpdag} & -0.141 & (0.212) & -0.141 & (0.210) \\
\midrule
\multicolumn{5}{l}{\textit{Train size $N = 500$}} \\
Vanilla original & -0.028 & (0.043) & -0.028 & (0.043) \\
Vanilla topological & -0.003 & (0.025) & -0.003 & (0.025) \\
Vanilla reverse top. & -0.033 & (0.040) & -0.032 & (0.040) \\
\gls{dag}-aware & \textbf{0.001} & (0.023) & \textbf{0.001} & (0.023) \\
\Gls{opdag} & -0.003 & (0.025) & -0.003 & (0.025) \\
Discovered \gls{cpdag} & -0.013 & (0.096) & -0.012 & (0.100) \\
\midrule
Test set & 0.022 & {--} & 0.022 & {--} \\
\bottomrule
\end{tabular}
\end{table}

Conversely, reverse topological ordering predominantly degrades performance across metrics (\cref{sec:appendix_reverse}). This confirms that column ordering significantly impacts TabPFN's data generation quality.

\paragraph{DAG-Aware Generation Outperforms Vanilla TabPFN}
\gls{dag}-aware generation shows significant improvements in \gls{cmd} for \num{24} of \num{35} configurations (\cref{fig:forest_dag_cpdag_opdag_cmd}, left), compared to vanilla \gls{tabpfn} with original ordering, with \num{2} degradations on \gls{css} at $N = 20$ and $N = 100$.
Improvements are strongest on \gls{csm}, \gls{cwa}, and \gls{cmc}, with improvement sizes decreasing at larger training sizes. 
The \gls{kmtvd} and \gls{nnaa} results also show widespread improvements across datasets (\cref{fig:forest_dag_vanilla_kmtvd_nnaa}), with one degradation on \gls{clb} in \gls{kmtvd}.
We also compare against five external tabular generators, all trained per dataset, including two causal-aware baselines given the true \gls{dag}. On the custom \gls{scm}, \gls{dag}-aware \gls{tabpfn} still outperforms all five (\cref{sec:appendix_external_baselines}).

Comparing vanilla \gls{tabpfn} and \gls{dag}-aware generation under the same topological ordering isolates the contribution of causal conditioning from the choice of feature ordering. \Gls{dag}-aware generation still significantly improves \gls{cmd} on \num{20} of \num{35} configurations with no degradations (\cref{fig:forest_dag_topo_vanilla_topo_cmd_main}), while \gls{kmtvd} and \gls{nnaa} improve less but rarely degrade (\cref{fig:forest_dag_topo_vanilla_topo_combined}).

\paragraph{\gls{pdag}-Based Generation Improves Over Vanilla TabPFN When Sufficiently Oriented}
The \gls{opdag} shows \num{15} significant \gls{cmd} improvements and \num{3} significant degradations (\cref{fig:forest_dag_cpdag_opdag_cmd}, right), two on \gls{clb} at $N = 20$ and $N = 50$, and one on \gls{css} at $N = 20$.
Conversely, the discovered \gls{cpdag} shows no significant improvements and \num{7} significant degradations in \gls{cmd} (\cref{fig:forest_cpdag_discovered_cmd}). Three of the seven degradations occur on \gls{cmc}, where PC orients most discovered edges with low accuracy (\cref{tab:pc_discovery_metrics}).
Most configurations (\num{28} of \num{35}) show no significant difference from vanilla \gls{tabpfn}.
The \gls{kmtvd} and \gls{nnaa} results are consistent with these patterns (\cref{fig:forest_cpdag_combined_2marg,fig:forest_cpdag_combined_nnaa}).
\input{figures_main.tex}

\subsection{Treatment Effect Preservation}

\paragraph{Vanilla TabPFN with Topological Ordering Improves ATE Preservation}
Topological ordering yields \num{18} significant improvements in \gls{ate} preservation across \num{42} configurations (\Cref{fig:forest_vanilla_topological_ate}), with \num{2} degradations, one on \gls{cmc} and one on \gls{css}, at \num{20} samples.
\gls{csm} shows the greatest improvements, particularly at smaller training sizes.
On \gls{sgl}, topological ordering produces \num{5} significant improvements out of \num{6} training sizes (\Cref{fig:forest_vanilla_topological_ate_sgl}), with larger effect sizes than other datasets.
Reverse topological ordering produces large degradations on \gls{csm} (\Cref{fig:forest_vanilla_worst_ate}).

\paragraph{DAG-Aware Generation Improves ATE Preservation}
\gls{dag}-aware generation shows \num{26} significant improvements across \num{42} configurations with \num{2} degradations, one on \gls{cmc} and one on \gls{css}, at \num{20} samples (\Cref{fig:forest_dag_cpdag_opdag_ate}, left), compared to \num{18} improvements for vanilla topological ordering.
\gls{csm} and \gls{cms} show significant improvements across all training sizes, with the largest reductions at \num{20} samples.
\gls{cns} shows significant improvements at most training sizes.
Under the same topological ordering, \gls{dag}-aware generation still achieves \num{15} of \num{42} significant improvements in \gls{ate} preservation, with no degradations (\cref{fig:forest_dag_topo_vanilla_topo_ate}).

\paragraph{\gls{pdag}-Based Generation Improves ATE Preservation When Sufficiently Oriented}
The \gls{opdag} shows \num{23} significant improvements across \num{42} configurations, with one degradation on \gls{cmc} at $N=20$ and one on \gls{css} at $N=200$ (\Cref{fig:forest_dag_cpdag_opdag_ate}, right).
Improvements are strongest on \gls{cns} (across all training sizes), and on \gls{cms} and \gls{csm} (at smaller training sizes).
The discovered \gls{cpdag} shows no consistent effects, with one improvement and one degradation across \num{42} configurations (\Cref{fig:forest_cpdag_discovered_ate,tab:pc_discovery_metrics_ate}).

\section{Discussion}

We demonstrated that vanilla \gls{tabpfn}, which generates features autoregressively in the input column order, is sensitive to that ordering. When the column order conflicts with the causal structure, the model conditions on descendants or colliders. While this is not a problem in the population limit, for small sample sizes the estimation error in learned conditionals introduces spurious correlations that degrade both synthetic data quality and treatment effect preservation. We addressed this limitation by injecting causal structure into the generation process, using the full \gls{dag} when it is available and partial causal knowledge otherwise. For the partial case we evaluated both the \gls{opdag} and \glspl{cpdag} discovered from data. \Gls{dag}-aware generation yields the most consistent improvements, the \gls{opdag} shows moderate gains, and a discovered \gls{cpdag} has no consistent effect, with a potential benefit that depends on how well the causal structure is recovered.

Reordering columns topologically improves both distributional quality and treatment effect preservation, while reversing causal direction worsens both.
\gls{dag}-aware generation conditions each variable only on its causal parents rather than on all previously generated features; this produces higher-quality synthetic data with minimal degradations across datasets and metrics.
The single degradation in \gls{kmtvd} occurs on \gls{clb}, whose sparse causal structure (most nodes have a single parent) limits the conditioning context available to the model.

\Glspl{opdag} orienting only edges into colliders show improvements in synthetic data quality and \gls{ate} preservation.
However, \glspl{cpdag} discovered from data often leave many edges unoriented, causing the model to fall back to vanilla sequential conditioning; when the recovered graph is inaccurate, they can even degrade performance. When \gls{pc} recovers enough correct structure from the data, by contrast, discovered-\gls{cpdag} generation improves over vanilla \gls{tabpfn} (\Cref{sec:appendix_recoverable}).
Beyond sufficient edge orientation, the position of colliders within the causal graph also matters.
When these are central (as in \gls{cmc}), generating oriented variables first allows the remaining variables to follow the causal direction, producing improvements. When they are at the end of a causal chain (as in \gls{clb}), generating oriented variables first effectively reverses the generation order relative to the causal direction, leading to degradations.

Causal conditioning preserves causal effects better than vanilla \gls{tabpfn}, particularly with limited training data. Under interventions, it also preserves more of the real data's empirical conditional independencies (\Cref{sec:appendix_ci_preservation}).
On \gls{csm}, \gls{dag}-aware generation reduces the median absolute \gls{ate} error from \num{1.23} for vanilla \gls{tabpfn} to \num{0.186} units at \num{20} samples, with improvements holding at larger training sizes. \Cref{tab:ate_scale_reference} reports the absolute \gls{ate} scale in native units for every dataset and training size.
Since the \gls{csm} is near-deterministic ($\sigma = 10^{-5}$), we verify that the benefits of causal conditioning hold under higher noise ($\sigma = 10^{-2}$, \cref{sec:appendix_noise}).
The \gls{cmc} degradation at $N = 20$ in the \gls{ate} experiments also occurs with vanilla \gls{tabpfn} under topological ordering, which does not modify the generation mechanism. This indicates that the effect is a model property rather than a consequence of causal conditioning.

Our findings extend to \gls{sgl}, a realistic dataset derived from an \gls{fda}-approved \gls{t1dm} simulator whose partial causal knowledge comes from validated physiological mechanisms. On this dataset, partial topological ordering improves synthetic data quality (\Cref{fig:forest_vanilla_topological_ordering}). Unlike the other datasets, these improvements increase with training size, which we attribute to the interplay between sample size and data dimensionality: with \num{38} variables, the model needs sufficient context to benefit from correct ordering. In the smaller \glspl{scm} (\numrange{4}{12} variables), the model already benefits from correct ordering at small $N$, but this advantage diminishes as data increases, because vanilla \gls{tabpfn} can then estimate the relevant conditionals well enough on its own.
On \gls{sgl}, the same ordering also better preserves treatment effects (\Cref{fig:forest_vanilla_topological_ate_sgl}), which highlights the relevance to real-world medical applications, where accurate treatment effect preservation from synthetic data can support early-stage research while reducing the need to share sensitive patient data.

\subsection{Limitations and Future Work}
This work is subject to the following limitations. First, \gls{dag}-aware generation assumes full knowledge of the causal structure, which is rarely available in practice. 
\gls{pdag}-based generation shows that partial causal information can still be valuable, but its effectiveness depends on collider placement in ways that cannot be predicted without knowledge of the true graph.
Second, causal discovery relied solely on \gls{pc}-stable. A preliminary evaluation with ReX~\citep{renero2026rex}, which returns fully oriented DAGs, yielded worse synthetic data quality since incorrect orientations propagate into the conditioning strategy (\cref{sec:appendix_rex}).
Third, \gls{cmd} assumes monotonic dependencies and is uninformative for \glspl{scm} with multiplicative equations, as in \gls{css}.
Fourth, the interventional analysis focused exclusively on the \gls{ate}; other causal estimands or downstream tasks may respond differently to causal conditioning.
Finally, our study focused primarily on \gls{tabpfn}. 
Future work should explore ordering strategies for incomplete causal knowledge, evaluate a broader range of causal discovery algorithms, select evaluation metrics based on dependency structure, and test these findings on other neural autoregressive architectures, such as TabularARGN~\citep{tiwald2025tabularargn}.

\begin{acknowledgements}
We thank the anonymous reviewers for their constructive comments, which helped improve the paper, and Ameen Abu-Hanna, Otto Nyberg, and Juliette Ortholand for their feedback on an earlier version of this manuscript.
We acknowledge ISCRA for awarding this project access to the LEONARDO supercomputer, owned by the EuroHPC Joint Undertaking, hosted by CINECA (Italy).
We also thank the University of Trieste for providing additional computational resources.
\end{acknowledgements}

\bibliography{references}

\newpage
\onecolumn
\title{Improving TabPFN's Synthetic Data Generation by Integrating Causal Structure\\(Supplementary Material)}
\begingroup
\renewcommand{\maketitlehooka}{\vbox\bgroup}
\maketitle
\endgroup
\appendix
\crefalias{section}{appendix}
\crefalias{subsection}{appendix}
\crefalias{subsubsection}{appendix}
\renewcommand{\thefigure}{A\arabic{figure}}
\renewcommand{\theHfigure}{A\arabic{figure}}
\setcounter{figure}{0}
\renewcommand{\thetable}{A\arabic{table}}
\renewcommand{\theHtable}{A\arabic{table}}
\setcounter{table}{0}
\input{appendix}

\end{document}

%% file: figures_main.tex
\begin{figure}[!htbp]
    \centering
    \includegraphics[width=\linewidth]{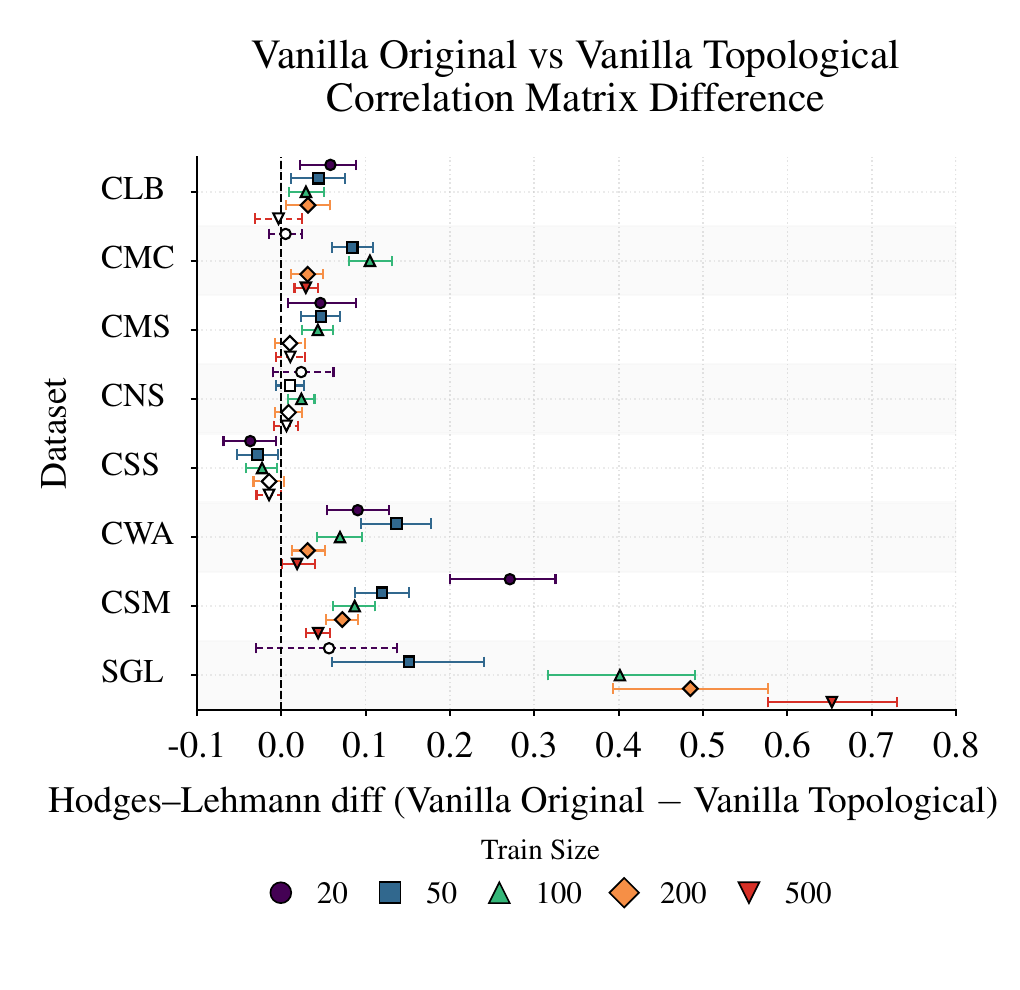}
    \caption{Hodges--Lehmann estimates comparing vanilla \gls{tabpfn} with original ordering versus topological ordering in \gls{cmd}. 
    Positive values indicate that topological ordering achieves lower metric values (i.e., better synthetic data quality).
    Filled markers with solid error bars indicate significance at $p<0.05$ (Holm correction).}
    \label{fig:forest_vanilla_topological_ordering}
\end{figure}

\begin{figure*}[!htbp]
  \centering
  \begin{adjustbox}{max width=\textwidth, max totalheight=0.48\textheight}
  \includegraphics{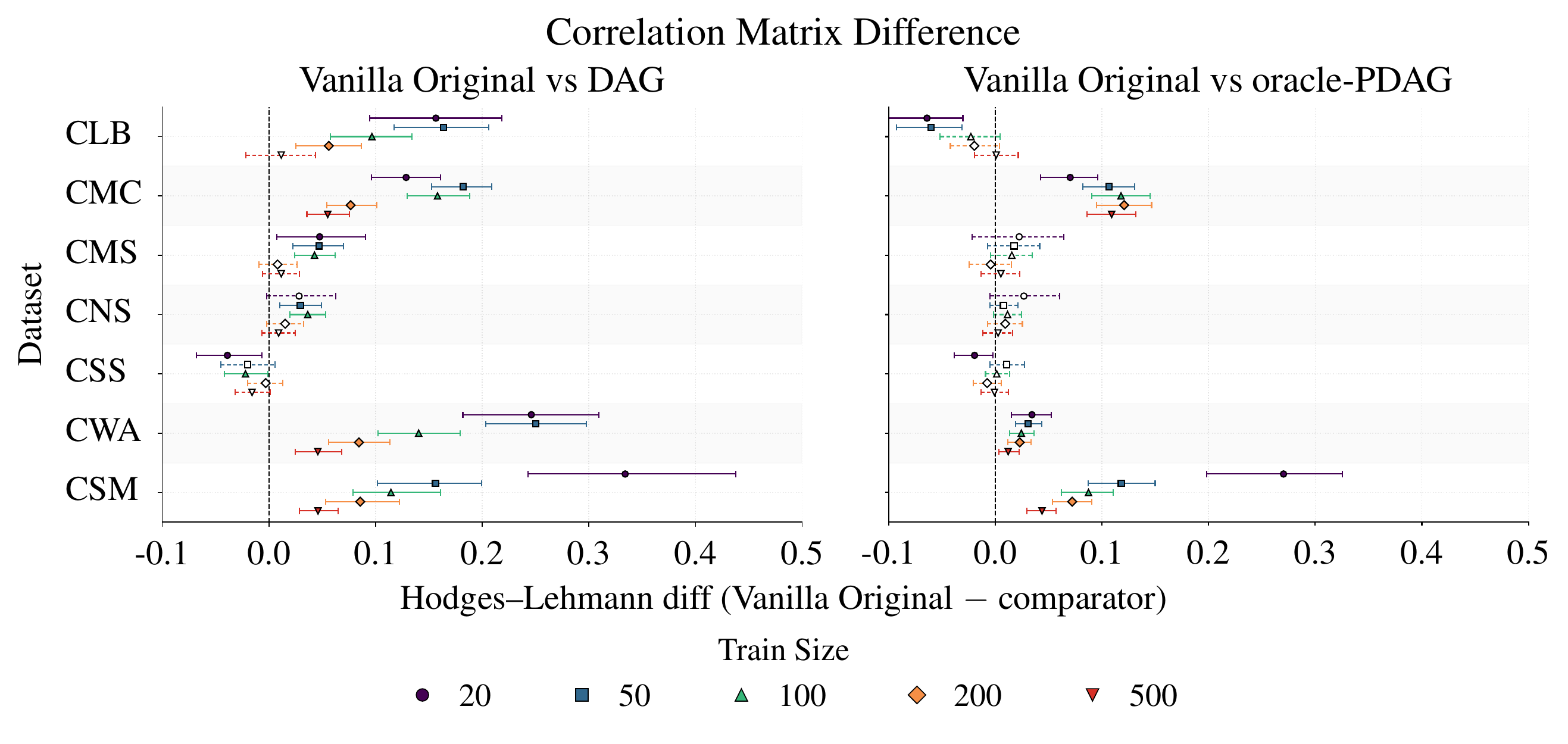}
  \end{adjustbox}
  \caption{Hodges--Lehmann estimates comparing vanilla \gls{tabpfn} with original ordering versus \gls{dag}-aware generation (left) and versus the \gls{opdag} (right) in \gls{cmd}. 
  Positive values indicate that the respective method achieves lower metric values (i.e., better synthetic data quality).
  Filled markers with solid error bars indicate significance at $p<0.05$ (Holm correction).}
  \label{fig:forest_dag_cpdag_opdag_cmd}
\end{figure*}

\begin{figure}[!htbp]
    \centering
    \includegraphics[width=\linewidth]{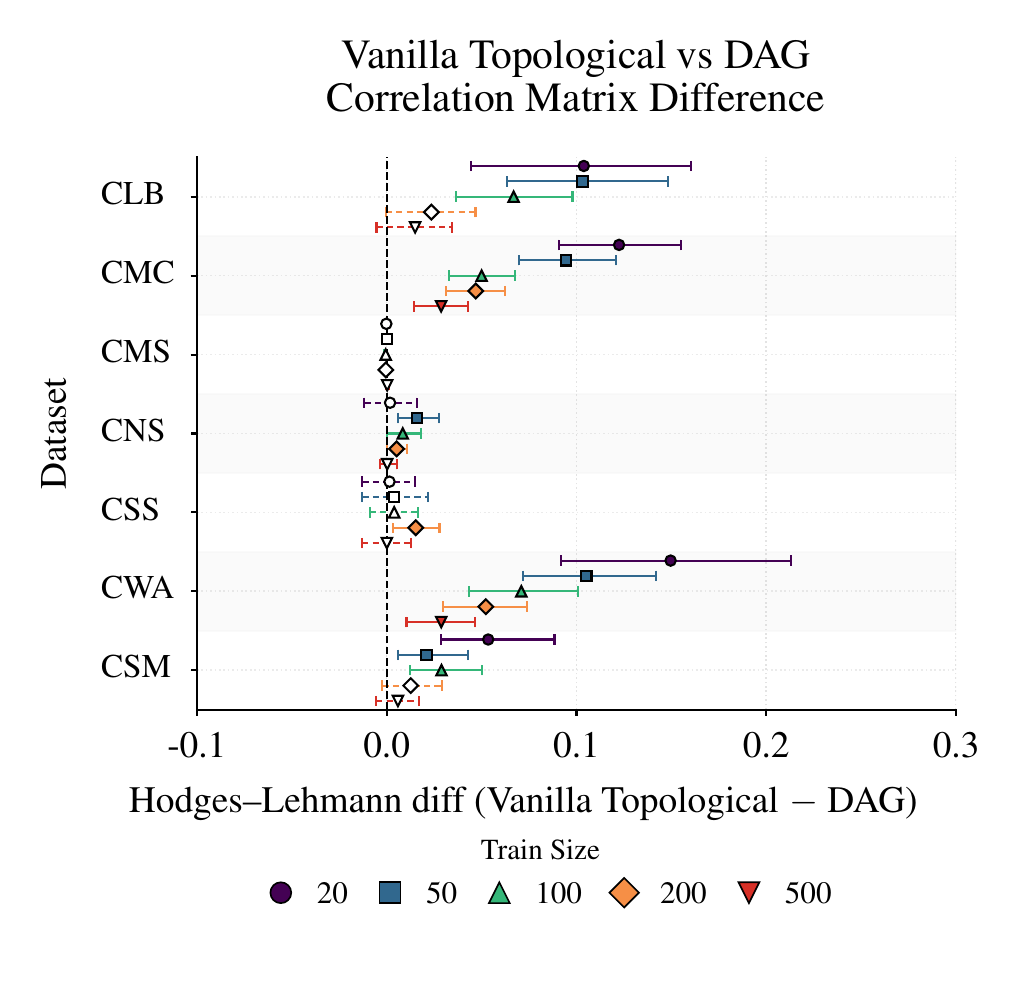}
    \caption{Hodges--Lehmann estimates comparing \gls{dag}-aware generation and vanilla \gls{tabpfn} when both use topological ordering, in \gls{cmd}. With feature ordering held fixed, this isolates the contribution of parent-based conditioning from the ordering itself. Positive values indicate that \gls{dag}-aware generation achieves lower metric values (i.e., better synthetic data quality). Filled markers with solid error bars indicate significance at $p<0.05$ (Holm correction).}
    \label{fig:forest_dag_topo_vanilla_topo_cmd_main}
\end{figure}

\begin{figure}[!htbp]
    \centering
    \includegraphics[width=\linewidth]{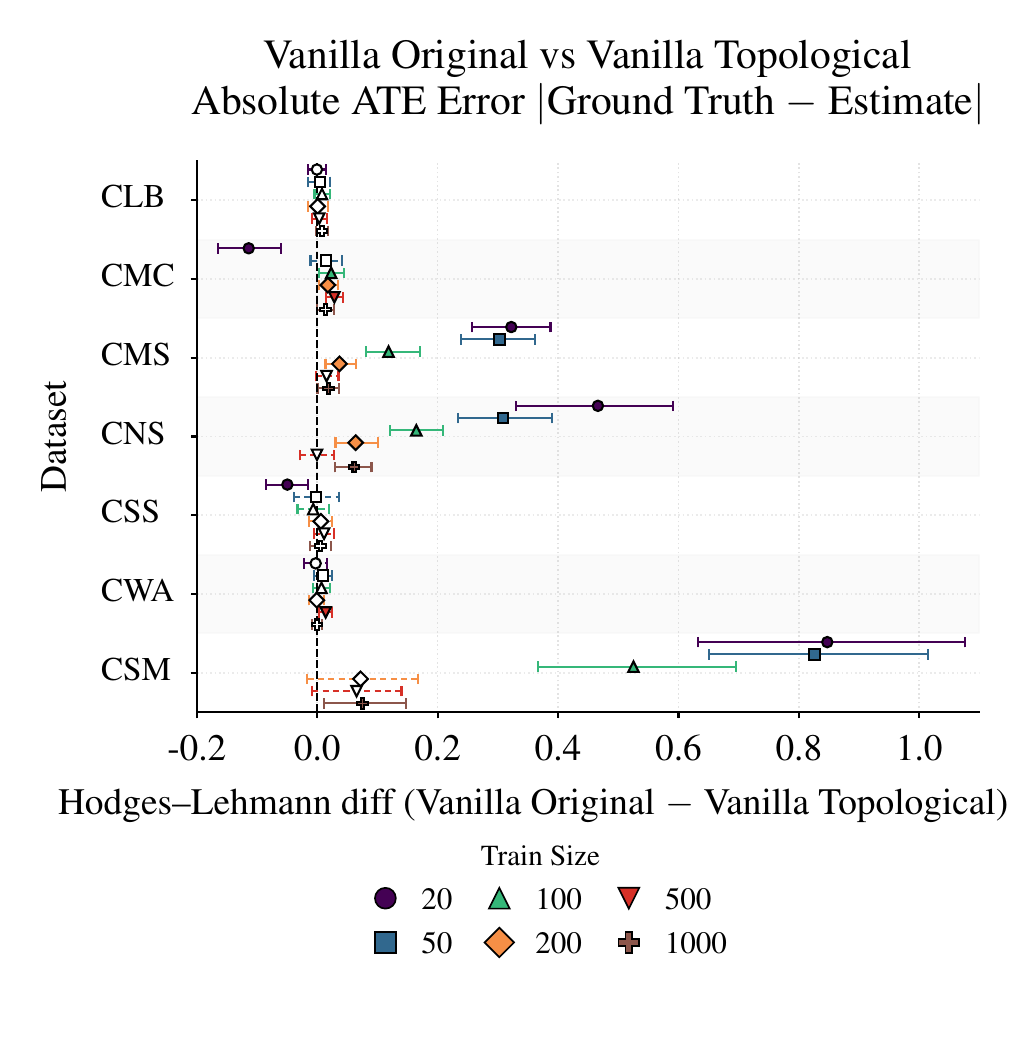}
    \caption{Hodges--Lehmann estimates of the reduction in absolute \gls{ate} error ($\Delta_{\text{ATE}}$) when comparing vanilla \gls{tabpfn} with original ordering versus topological ordering on CSuite and \gls{csm} datasets.
    Positive values indicate smaller errors (closer to ground truth) for topological ordering; negative values indicate larger errors.
    Filled markers with solid error bars indicate significance at $p<0.05$ (Holm correction).}
    \label{fig:forest_vanilla_topological_ate}
\end{figure}


\begin{figure*}[!htbp]
  \centering
  \begin{adjustbox}{max width=\textwidth, max totalheight=0.48\textheight}
    \includegraphics{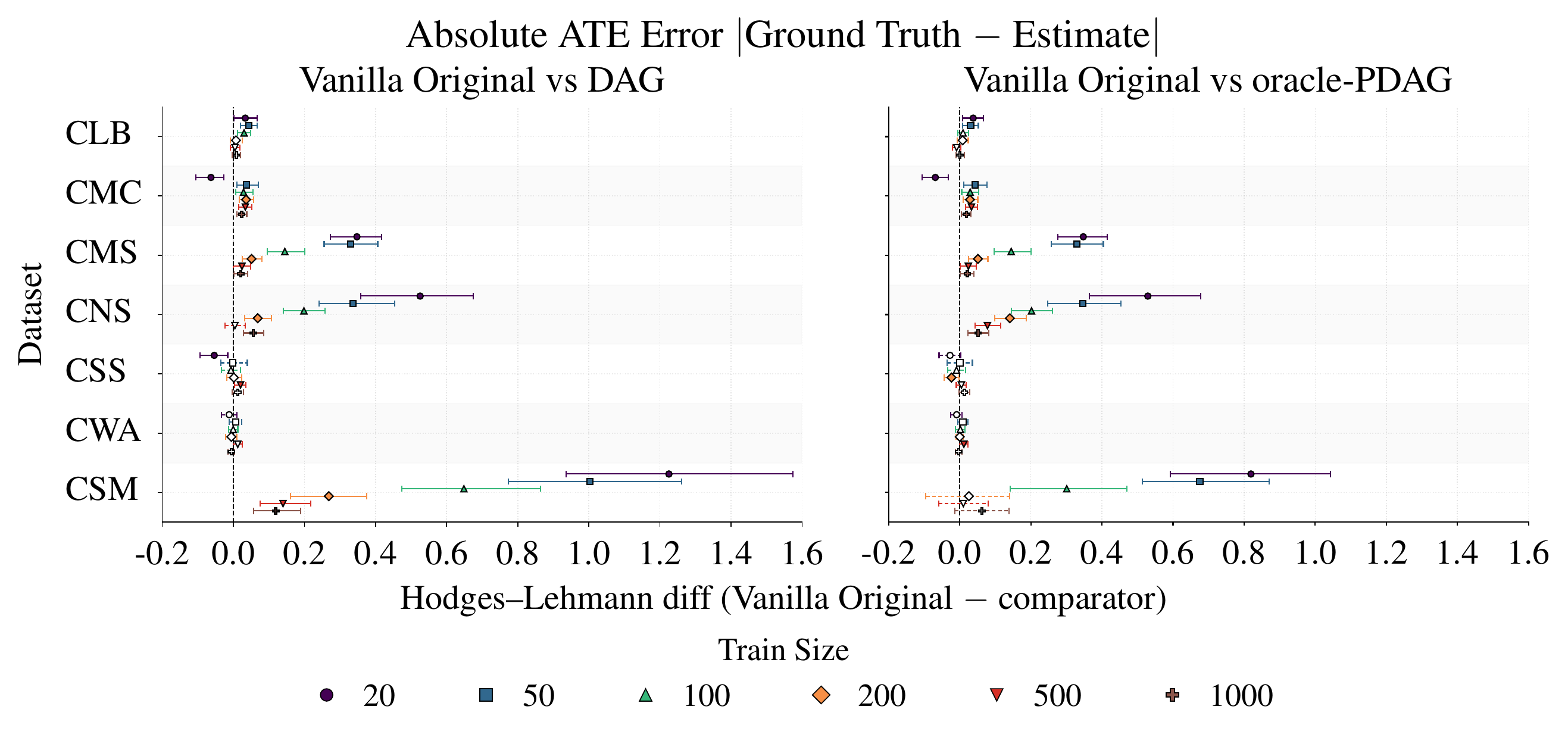}
  \end{adjustbox}
  \caption{Hodges--Lehmann estimates of the reduction in absolute \gls{ate} error ($\Delta_{\text{ATE}}$) when comparing vanilla \gls{tabpfn} with original ordering versus \gls{dag}-aware generation (left) and versus the \gls{opdag} (right).
  Positive values indicate smaller errors (closer to ground truth) for the respective method (\gls{dag}-aware or \gls{opdag}); negative values indicate larger errors.
  Filled markers with solid error bars indicate significance at $p<0.05$ (Holm correction).}
  \label{fig:forest_dag_cpdag_opdag_ate}
\end{figure*}

%% file: appendix.tex
\section{Nearest-Neighbor Adversarial Accuracy}
\label{sec:nnaa_definition}
The \gls{nnaa}~\citep{yale2020generation} is a nearest-neighbor metric used in privacy-oriented evaluations of synthetic data. It quantifies the distinguishability between synthetic and real data based on nearest-neighbor distances. By construction, unlike the \gls{cmd}, which compares pairwise association, and the \gls{kmtvd} with $k=2$, which compares two-way joint distributions, the \gls{nnaa} assesses similarity at the level of individual records in the full joint space, capturing aspects of fidelity that neither pairwise metric can detect.
We use SynthEval's implementation with the Gower distance~\citep{lautrup2025syntheval}.
For each real sample~$i$, the distance $d_{TS}(i)$ to its nearest synthetic neighbor is compared with the distance $d_{TT}(i)$ to its nearest other real sample.
Similarly, for each synthetic sample~$j$, the distance $d_{ST}(j)$ to its nearest real neighbor is compared with the distance $d_{SS}(j)$ to its nearest other synthetic sample.
The adversarial accuracy is
\begin{equation}
\begin{split}
\mathrm{NNAA} = \frac{1}{2} \biggl[
  &\frac{1}{n} \sum_{i=1}^{n} \mathbf{1}\bigl(d_{TS}(i) > d_{TT}(i)\bigr) \\
  + &\frac{1}{n} \sum_{j=1}^{n} \mathbf{1}\bigl(d_{ST}(j) > d_{SS}(j)\bigr)
\biggr],
\end{split}
\end{equation}
where $\mathbf{1}(\cdot)$ denotes the indicator function, and $n$ is the number of samples in both the real and synthetic datasets.
Values near $0.5$ indicate that synthetic and real data are hard to distinguish. A value above $0.5$ means they are easy to tell apart. A value below $0.5$ means the synthetic samples are unusually close to the real ones. Since our interest is indistinguishability, we treat both sides equally and report each result as the distance from this ideal value, $\lvert\mathrm{NNAA} - 0.5\rvert$. Lower values mean greater indistinguishability. The raw \gls{nnaa} values are available in the code repository for analyses that separate the two cases. In our experiments, the median \gls{nnaa} lies above $0.5$ in every dataset and generation method. Because we measure \gls{nnaa} against a held-out test set that the model never sees during generation, this closeness reflects fidelity to the unseen real distribution.

\begin{figure}[!htbp]
\centering
\includegraphics[width=\linewidth, height=0.48\textheight, keepaspectratio]{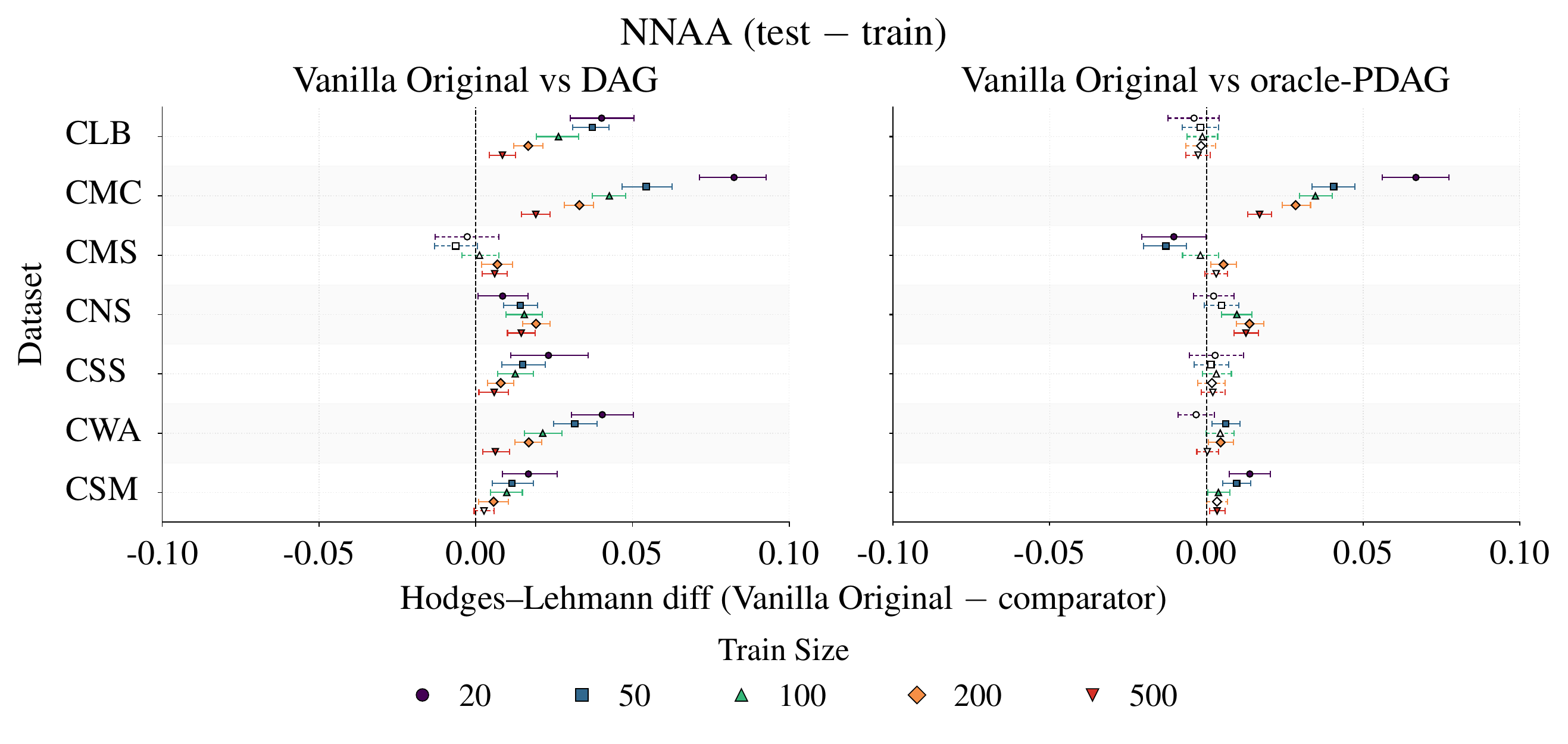}
\caption{Hodges--Lehmann estimates comparing vanilla \gls{tabpfn} with \gls{dag}-aware generation (left) and the \gls{opdag} (right) in the privacy loss (the difference between \gls{nnaa} on the held-out test set and \gls{nnaa} on the training set). Positive values indicate that the method's synthetic data are no closer to the training records than vanilla's. Filled markers with solid error bars indicate significance at $p<0.05$ (Holm correction).}
\label{fig:nnaa_traintest_dag_opdag}
\end{figure}

\begin{figure}[!htbp]
\centering
\includegraphics[width=0.72\linewidth, height=0.40\textheight, keepaspectratio]{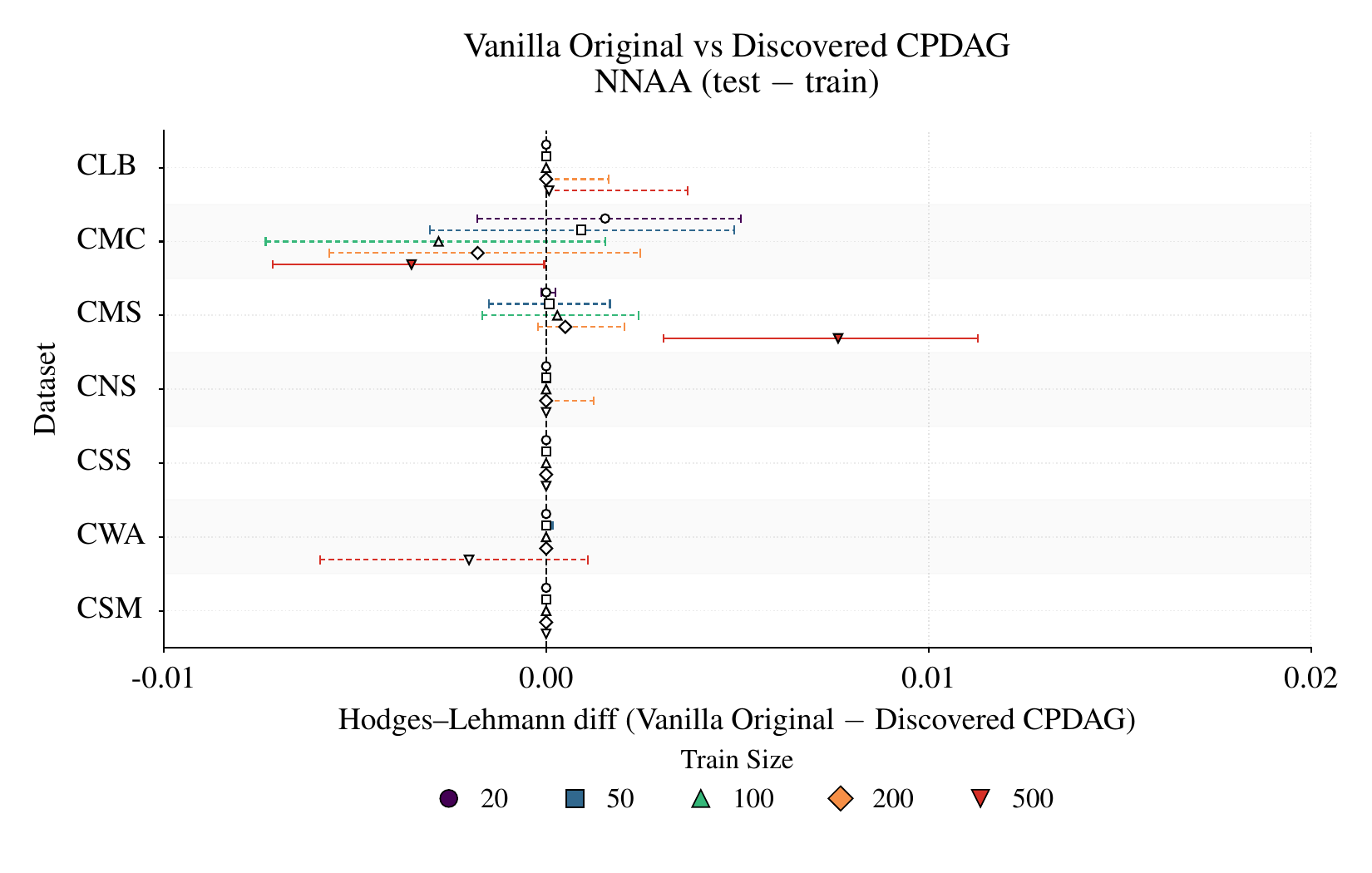}
\caption{Hodges--Lehmann estimates comparing vanilla \gls{tabpfn} with the discovered \gls{cpdag} in the privacy loss (the difference between \gls{nnaa} on the held-out test set and \gls{nnaa} on the training set). Positive values indicate that the synthetic data are no closer to the training records than vanilla's. Filled markers with solid error bars indicate significance at $p<0.05$ (Holm correction).}
\label{fig:nnaa_traintest_discovered}
\end{figure}

However, the model does see the training records. We therefore ask whether causal conditioning brings the synthetic data closer to them than vanilla \gls{tabpfn} does. To test this, we compute the same \gls{nnaa} against the training set. The difference between the two values is the privacy loss~\citep{yale2020generation}. A value near $0$ means the synthetic data are similarly close to the training and held-out records, while positive values indicate greater proximity to the training records. \Cref{fig:nnaa_traintest_dag_opdag,fig:nnaa_traintest_discovered} report it as vanilla \gls{tabpfn} minus each method. \Gls{dag}-aware generation lowers it in most settings and never significantly increases it. The \gls{opdag} also lowers it in many settings, though less consistently. The discovered \gls{cpdag} shows no clear effect.

\section{Custom SCM Definition}
\label{sec:appendix-custom-scm}

The custom collider \gls{scm} has causal structure $X_0 \to X_1 \leftarrow X_2 \leftarrow X_3$, with structural equations:
\begin{align}
X_0 &\sim \mathcal{N}(0, 1) \\
X_3 &\sim \mathcal{N}(0, 1) \\
X_2 &= 0.5 \, X_3 + \epsilon_2, \quad \epsilon_2 \sim \mathcal{N}(0, \sigma^2) \\
X_1 &= 5.0 \, X_0 + 10.0 \, X_2 + \epsilon_1, \quad \epsilon_1 \sim \mathcal{N}(0, \sigma^2)
\end{align}
where $\sigma = 10^{-5}$.
Variables $X_0$ and $X_3$ are independent root nodes, and $X_0$ and $X_2$ are d-separated since $X_2$ depends only on $X_3$.

\section{Order Sensitivity Across Datasets}
\label{sec:appendix_order_sensitivity}

To quantify how much feature ordering affects synthetic data quality, we measure the variability of each metric across the three orderings (original, topological, reverse topological).

For each dataset and training size, we compute:
\begin{itemize}
\item The metric value under each of the three orderings
\item The range: maximum value minus minimum value
\item Bootstrap confidence intervals for the median range (\num{1000} resamples, $\alpha = 0.05$)
\end{itemize}

A large range indicates that the model is highly sensitive to feature ordering. A small range indicates consistent performance regardless of ordering.

\begin{figure}[!htbp]
    \centering
    \includegraphics[width=\linewidth, height=0.48\textheight, keepaspectratio]{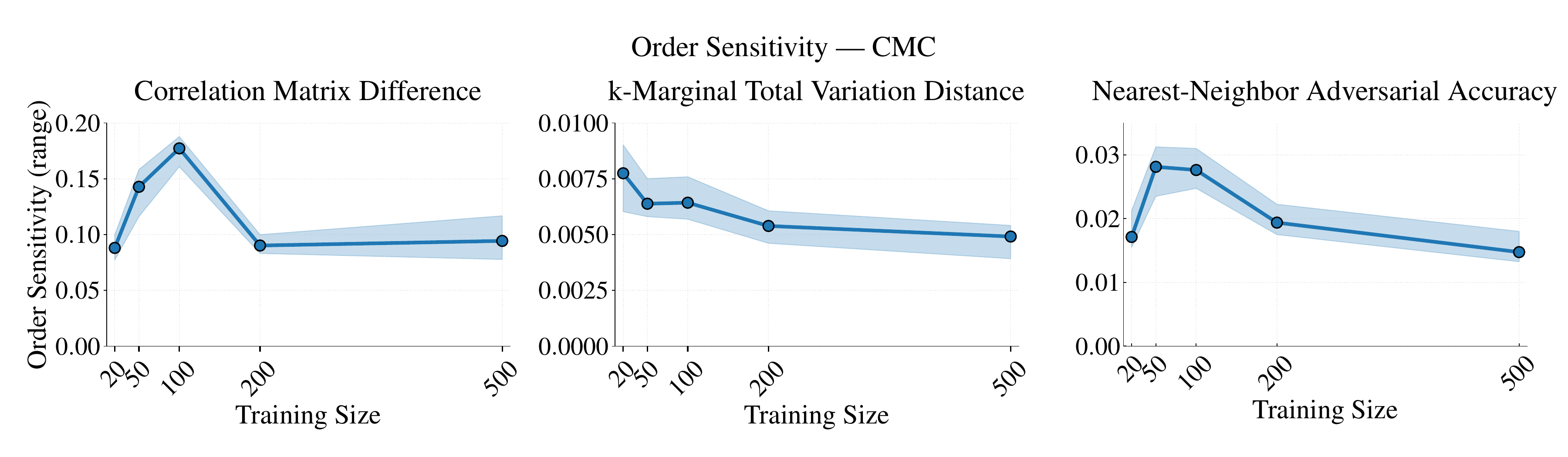}
    \caption{Order sensitivity for vanilla \gls{tabpfn} on \gls{cmc}.
    Y-axis shows the range of \gls{cmd} (left), \gls{kmtvd} ($k=2$, center), and \gls{nnaa} (right) values across three feature orderings: original, topological, and reverse topological. 
    Shaded regions show \SI{95}{\percent} bootstrap confidence intervals.}
    \label{fig:order_sensitivity_combined}
\end{figure}

\Cref{fig:order_sensitivity_combined} shows results for \gls{cmc}. This dataset exemplifies a general pattern observed across all evaluated datasets: ordering sensitivity decreases with training size but persists even at larger sample sizes. Results for additional datasets are available in the code repository.

\section{Random-Order Sensitivity of the Vanilla Baseline}
\label{sec:appendix_random_order}
Throughout the paper we use each dataset's original column ordering as the vanilla baseline, since it matches the common use of \gls{tabpfn} for synthetic data generation, where the practitioner feeds the data as-is without a causally informed ordering. As an additional check that our conclusions do not depend on this particular ordering, we also compare against vanilla \gls{tabpfn} under random column orderings. For each dataset we sample a fixed pool of \num{10} random permutations and cyclically assign one to each of the \num{100} paired repetitions. Vanilla \gls{tabpfn} generates with the assigned ordering on the same cached training split, and the generated data are evaluated with the paper's paired protocol (Wilcoxon signed-rank with Pratt ties and Holm correction).

\begin{figure}[!htbp]
\centering
\begin{adjustbox}{max width=\linewidth, max totalheight=0.48\textheight}
\includegraphics{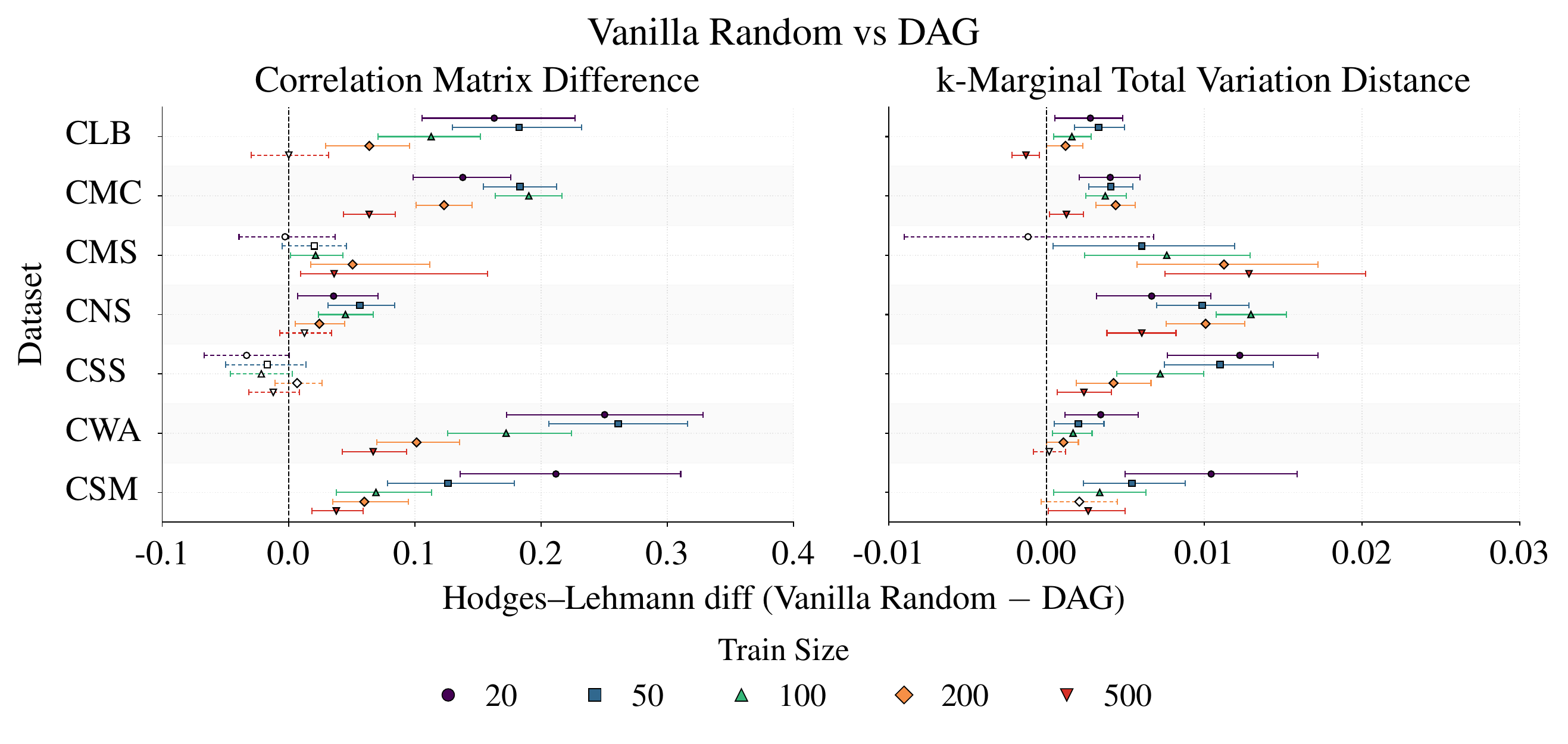}
\end{adjustbox}
\caption{Hodges--Lehmann estimates comparing vanilla \gls{tabpfn} with random column orderings and \gls{dag}-aware generation with topological ordering, in \gls{cmd} (left) and \gls{kmtvd} ($k=2$, right).
Positive values indicate that \gls{dag}-aware generation achieves lower metric values (i.e., better synthetic data quality).
Filled markers with solid error bars indicate significance at $p<0.05$ (Holm correction).}
\label{fig:forest_random_order_combined}
\end{figure}

\begin{figure}[!htbp]
\centering
\includegraphics[width=\linewidth, height=0.48\textheight, keepaspectratio]{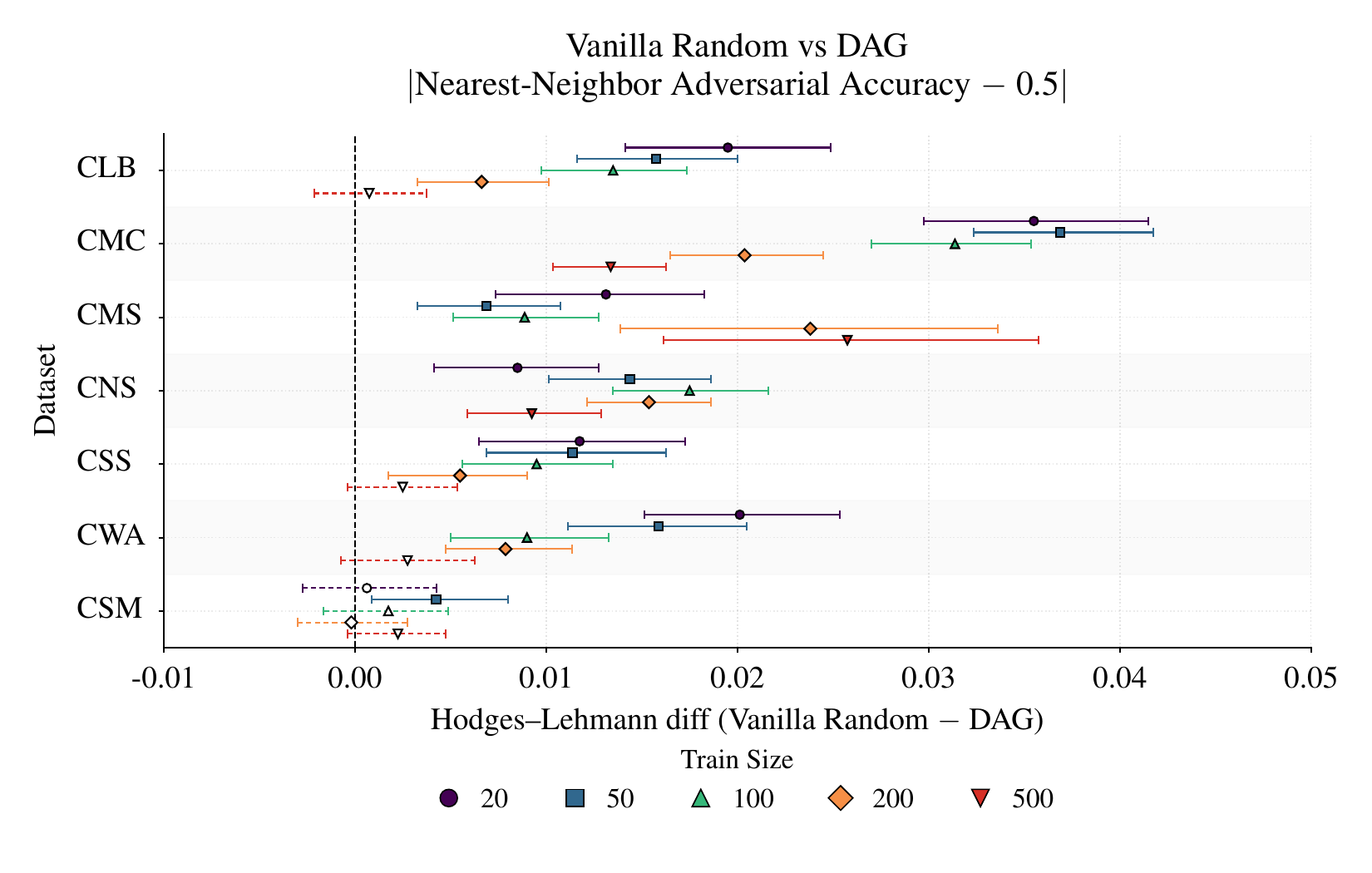}
\caption{Hodges--Lehmann estimates comparing vanilla \gls{tabpfn} with random column orderings and \gls{dag}-aware generation with topological ordering, in \gls{nnaa} (reported as the distance from the ideal value, $\lvert\mathrm{NNAA} - 0.5\rvert$).
Positive values indicate that \gls{dag}-aware generation achieves lower values (i.e., greater indistinguishability between synthetic and real data).
Filled markers with solid error bars indicate significance at $p<0.05$ (Holm correction).}
\label{fig:forest_random_order_nnaa}
\end{figure}

\Cref{fig:forest_random_order_combined} compares \gls{dag}-aware generation against vanilla \gls{tabpfn} with random orderings on the seven datasets of the main comparison (\num{35} dataset--size combinations). \Gls{dag}-aware generation achieves \num{26} significant improvements and no degradations in \gls{cmd}, and \num{31} improvements with \num{1} degradation (\gls{clb} at $N = 500$) in \gls{kmtvd}. In \gls{nnaa}, it achieves \num{28} improvements with no degradations (\Cref{fig:forest_random_order_nnaa}). These results confirm that the advantage of \gls{dag}-aware generation holds whether the vanilla baseline uses the original ordering or random ones.

\section{Vanilla TabPFN with Topological Ordering Results}
\label{sec:appendix_topological}
\begin{figure}[!htbp]
\centering
\includegraphics[width=\linewidth, height=0.48\textheight, keepaspectratio]{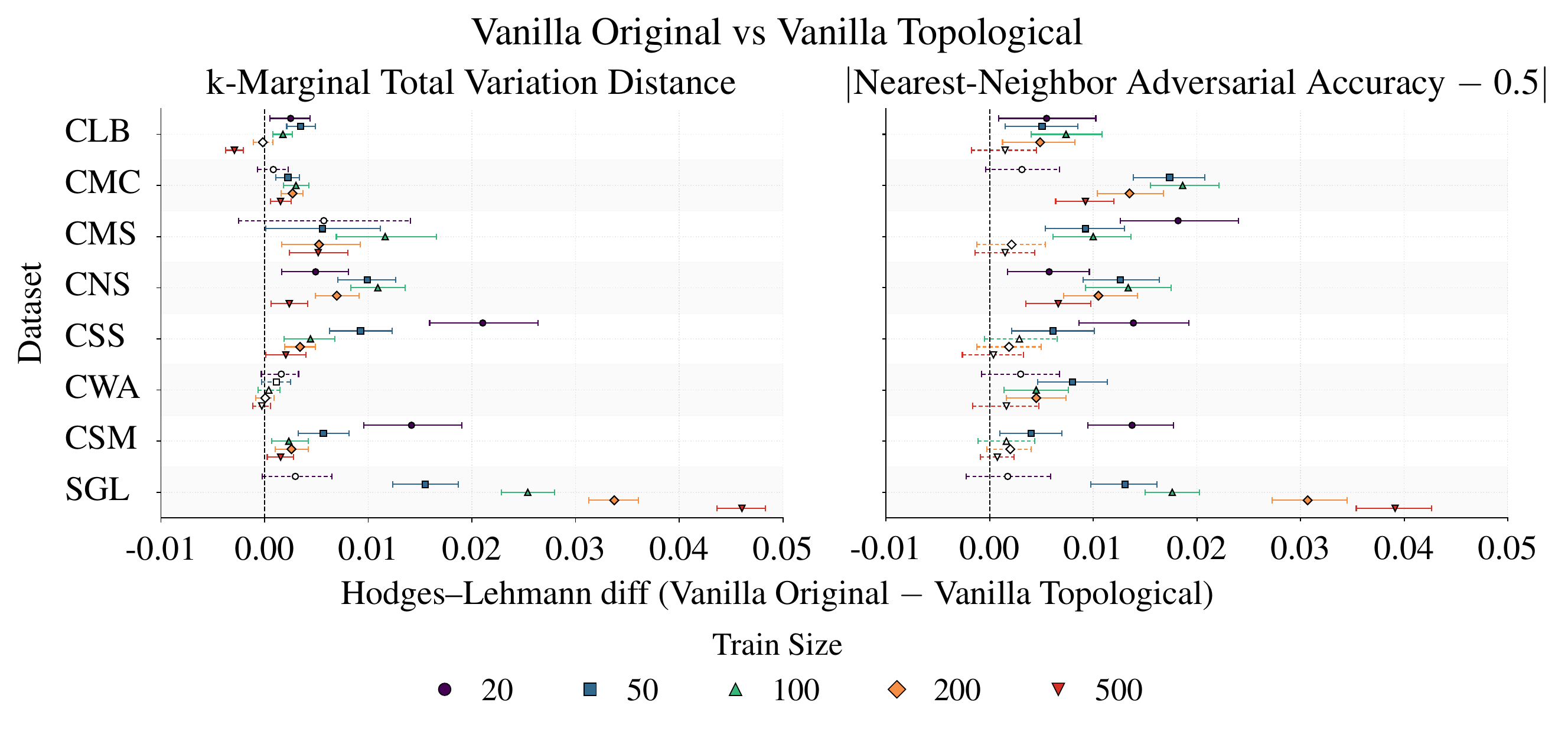}
\caption{Hodges--Lehmann estimates comparing vanilla \gls{tabpfn} with original ordering versus topological ordering in \gls{kmtvd} ($k=2$, left) and \gls{nnaa} (right), reported as the distance from the ideal value, $\lvert\mathrm{NNAA} - 0.5\rvert$.
Positive values indicate that topological ordering achieves lower \gls{kmtvd} (i.e., better pairwise fidelity) and lower $\lvert\mathrm{NNAA} - 0.5\rvert$ (i.e., greater indistinguishability between synthetic and real data).
Filled markers with solid error bars indicate significance at $p<0.05$ (Holm correction).}
\label{fig:forest_vanilla_topo_kmtvd_nnaa}
\end{figure}

\Cref{fig:forest_vanilla_topo_kmtvd_nnaa} reports \gls{kmtvd} and \gls{nnaa} results for vanilla \gls{tabpfn} with topological ordering. In \gls{kmtvd}, topological ordering yields \num{30} significant improvements and \num{1} degradation on \gls{clb} at \num{500} samples (left panel). In \gls{nnaa}, it yields \num{27} significant improvements with no degradations (right panel).

\section{Spurious Correlations in Custom SCM}
\label{sec:appendix-spurious-corr}

\Cref{tab:spurious_correlations} reports Pearson correlation coefficients for variable pairs that should be independent according to the custom \gls{scm} causal structure (\cref{sec:appendix-custom-scm}).
The test set baseline confirms the expected independence with correlations close to zero ($\rho \approx 0.02$).
Here we report the intermediate training sizes $N \in \{50, 100, 200\}$; $N = 20, 500$ are in the main text (\Cref{tab:spurious_correlations_main}).
Vanilla \gls{tabpfn} with original and reverse topological orderings introduces substantial spurious correlations, particularly at small training sizes ($|\rho| \approx 0.15$ at $N=20$).
\gls{dag}-aware generation maintains near-zero correlations across all training sizes.
\Gls{opdag} performs similarly to vanilla topological, while discovered \gls{cpdag} tracks vanilla original at the smallest training size (\Cref{tab:spurious_correlations_main}) and shows attenuated but more variable correlations at the sizes reported here, reflecting how undirected edges are resolved based on column order.

\begin{table}[!htbp]
\centering
\caption{Mean Pearson correlation coefficients for independent variable pairs in the custom collider \gls{scm}.
Values close to zero indicate correct independence preservation; the best value per column and training size is in \textbf{bold}.
Standard deviations across \num{100} repetitions in parentheses.}
\label{tab:spurious_correlations}
\begin{tabular}{l
  S[table-format=-1.3,detect-weight,detect-family] @{\hspace{1mm}} l
  S[table-format=-1.3,detect-weight,detect-family] @{\hspace{1mm}} l}
\toprule
Method & {$\rho(X_0, X_3)$} & & {$\rho(X_0, X_2)$} & \\
\midrule
\multicolumn{5}{l}{\textit{Train size $N = 50$}} \\
Vanilla original & -0.023 & (0.143) & -0.025 & (0.143) \\
Vanilla topological & 0.005 & (0.075) & 0.006 & (0.076) \\
Vanilla reverse top. & -0.030 & (0.148) & -0.031 & (0.146) \\
\gls{dag}-aware & \bfseries 0.000 & (0.023) & \bfseries 0.000 & (0.024) \\
\Gls{opdag} & 0.006 & (0.077) & 0.005 & (0.076) \\
Discovered \gls{cpdag} & -0.001 & (0.178) & -0.002 & (0.180) \\
\midrule
\multicolumn{5}{l}{\textit{Train size $N = 100$}} \\
Vanilla original & -0.042 & (0.114) & -0.043 & (0.114) \\
Vanilla topological & -0.017 & (0.064) & -0.017 & (0.063) \\
Vanilla reverse top. & -0.047 & (0.101) & -0.045 & (0.101) \\
\gls{dag}-aware & \bfseries 0.000 & (0.020) & \bfseries 0.000 & (0.020) \\
\Gls{opdag} & -0.017 & (0.063) & -0.017 & (0.064) \\
Discovered \gls{cpdag} & -0.029 & (0.140) & -0.029 & (0.141) \\
\midrule
\multicolumn{5}{l}{\textit{Train size $N = 200$}} \\
Vanilla original & -0.029 & (0.104) & -0.035 & (0.080) \\
Vanilla topological & -0.012 & (0.042) & -0.012 & (0.042) \\
Vanilla reverse top. & -0.036 & (0.082) & -0.028 & (0.112) \\
\gls{dag}-aware & \bfseries -0.001 & (0.025) & \bfseries -0.001 & (0.025) \\
\Gls{opdag} & -0.006 & (0.077) & -0.012 & (0.042) \\
Discovered \gls{cpdag} & \bfseries -0.001 & (0.152) & -0.005 & (0.147) \\
\midrule
Test set & 0.022 & {--} & 0.022 & {--} \\
\bottomrule
\end{tabular}
\end{table}

\section{Vanilla TabPFN with Reverse Topological Ordering Results}
\label{sec:appendix_reverse}
\begin{figure}[!htbp]
\centering
\begin{adjustbox}{max width=\linewidth, max totalheight=0.48\textheight}
\includegraphics{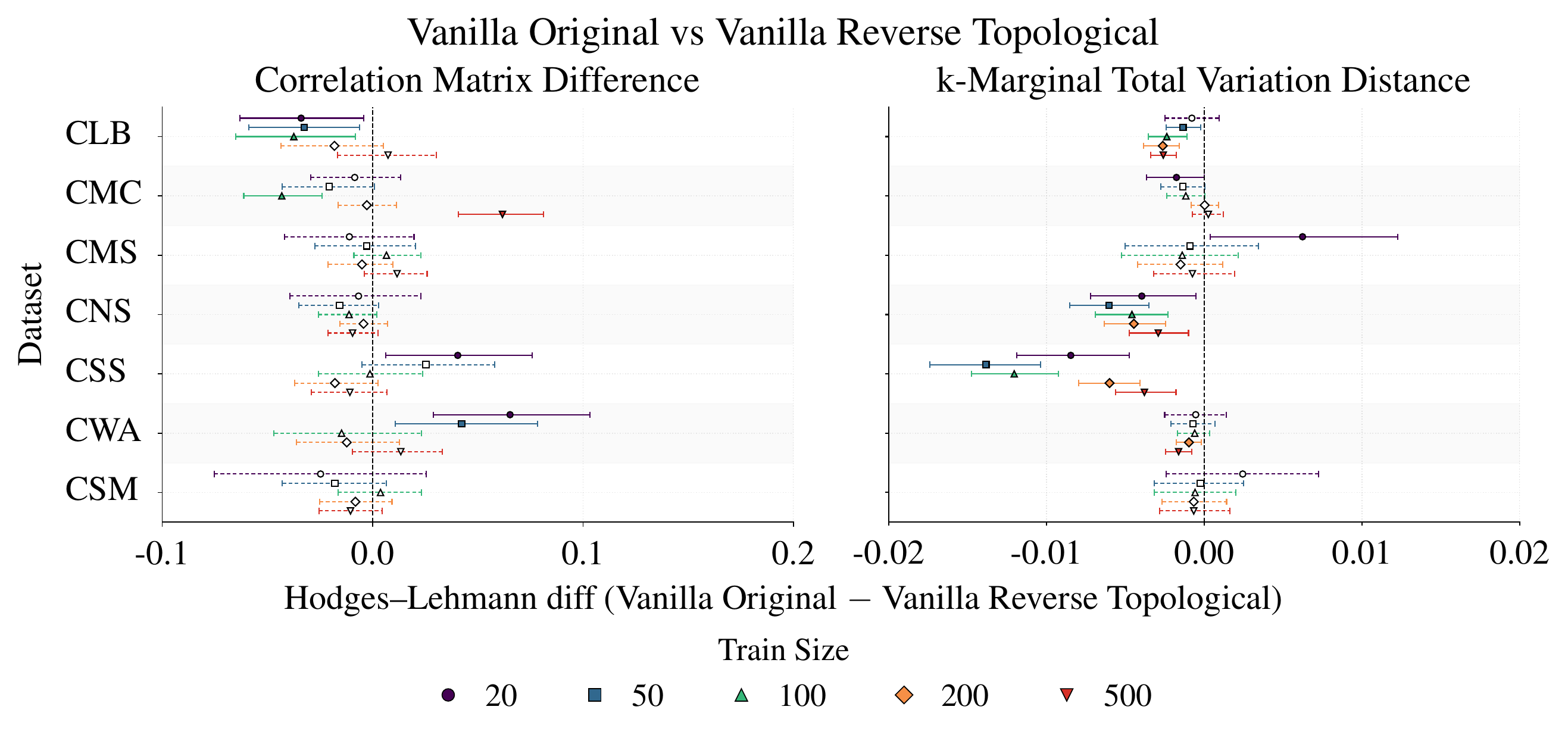}
\end{adjustbox}
\caption{Hodges--Lehmann estimates comparing vanilla \gls{tabpfn} with original ordering versus reverse topological ordering in \gls{cmd} (left) and \gls{kmtvd} ($k=2$, right).
Positive values indicate that reverse topological ordering achieves lower metric values (i.e., better synthetic data quality).
Filled markers with solid error bars indicate significance at $p<0.05$ (Holm correction).}
\label{fig:forest_vanilla_worst_combined}
\end{figure}

\begin{figure}[!htbp]
\centering
\includegraphics[width=\linewidth, height=0.48\textheight, keepaspectratio]{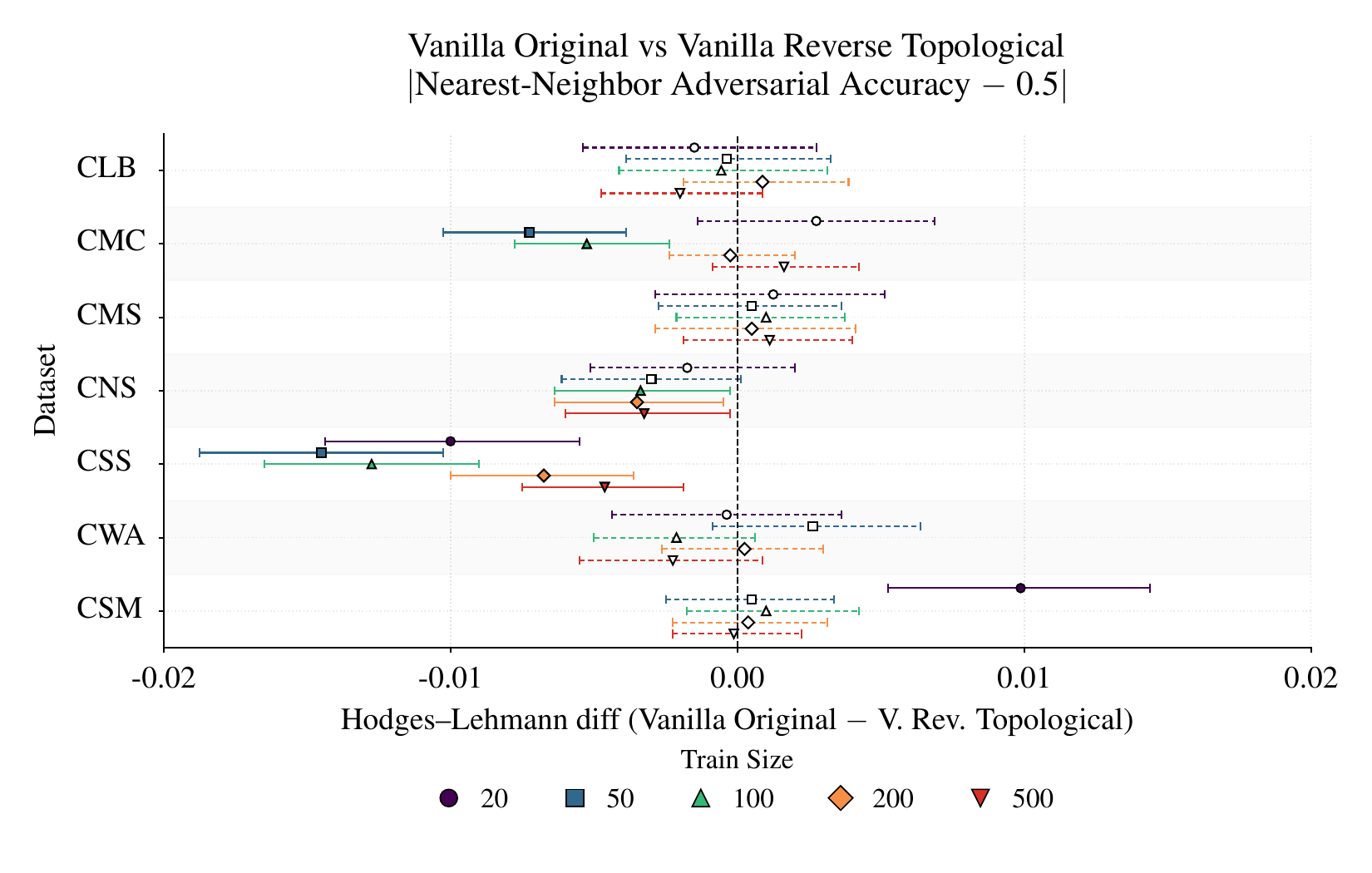}
\caption{Hodges--Lehmann estimates comparing vanilla \gls{tabpfn} with original ordering versus reverse topological ordering in \gls{nnaa} (reported as the distance from the ideal value, $\lvert\mathrm{NNAA} - 0.5\rvert$).
Positive values indicate that reverse topological ordering achieves lower values (i.e., greater indistinguishability between synthetic and real data).
Filled markers with solid error bars indicate significance at $p<0.05$ (Holm correction).}
\label{fig:forest_vanilla_worst_nnaa}
\end{figure}

This section reports results for vanilla \gls{tabpfn} with reverse topological ordering.
Reverse topological ordering shows predominantly negative effects across metrics, with \num{4} significant degradations in \gls{cmd} and \num{17} in \gls{kmtvd} (\Cref{fig:forest_vanilla_worst_combined}, left and right, respectively), against \num{4} significant improvements in \gls{cmd} and \num{1} in \gls{kmtvd}. \gls{nnaa} confirms this pattern with \num{10} significant degradations and \num{1} significant improvement (\Cref{fig:forest_vanilla_worst_nnaa}).

\FloatBarrier
\section{DAG-Aware Generation vs Vanilla: \gls{kmtvd} and \gls{nnaa} Results}
\label{sec:appendix_dag_vs_vanilla_nnaa}

\begin{figure}[!htbp]
\centering
\includegraphics[width=\linewidth, height=0.48\textheight, keepaspectratio]{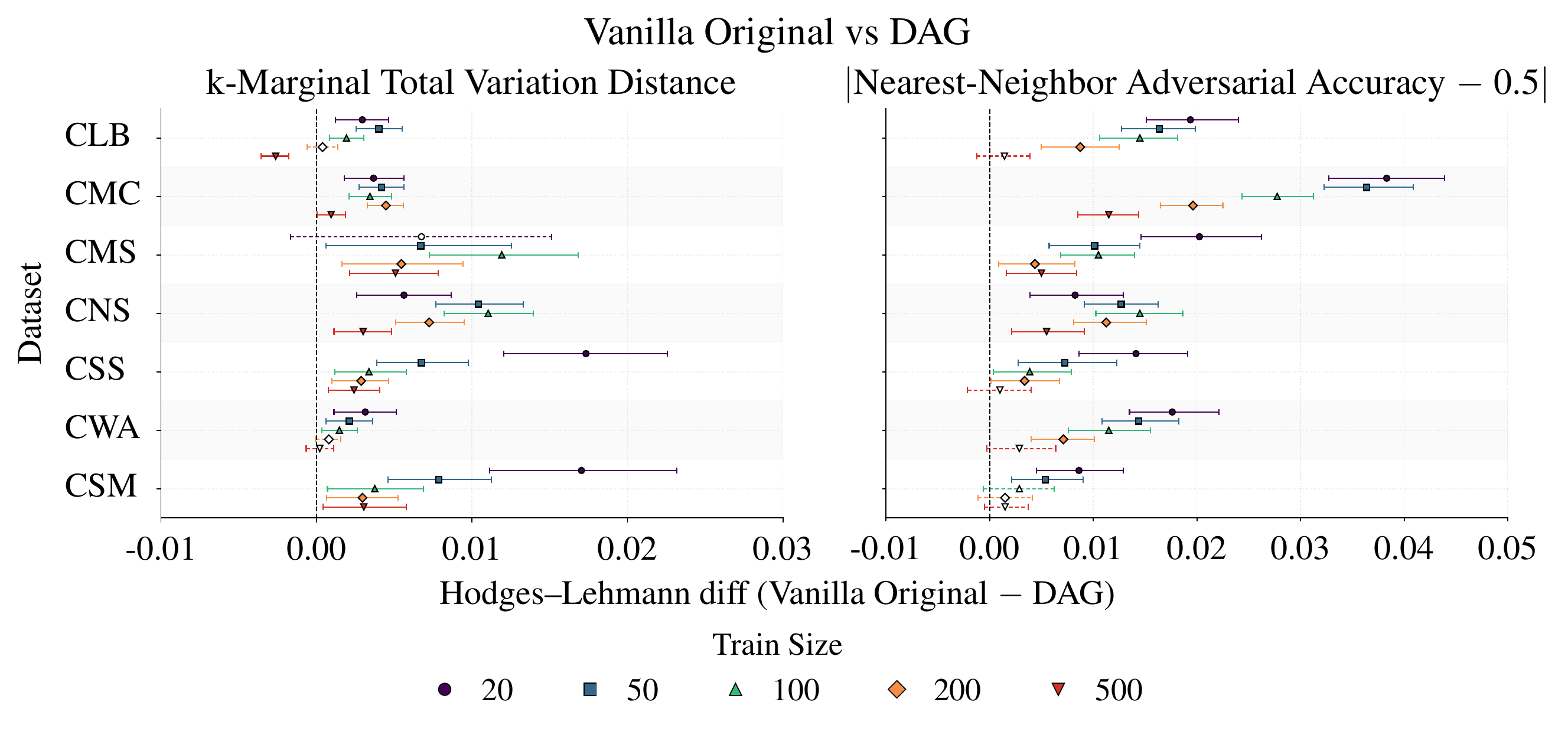}
\caption{Hodges--Lehmann estimates comparing vanilla \gls{tabpfn} with original ordering versus \gls{dag}-aware generation in \gls{kmtvd} ($k=2$, left) and \gls{nnaa} (right), reported as the distance from the ideal value, $\lvert\mathrm{NNAA} - 0.5\rvert$.
Positive values indicate that \gls{dag}-aware generation achieves lower \gls{kmtvd} (i.e., better pairwise fidelity) and lower $\lvert\mathrm{NNAA} - 0.5\rvert$ (i.e., greater indistinguishability between synthetic and real data).
Filled markers with solid error bars indicate significance at $p<0.05$ (Holm correction).}
\label{fig:forest_dag_vanilla_kmtvd_nnaa}
\end{figure}

\Cref{fig:forest_dag_vanilla_kmtvd_nnaa} reports \gls{kmtvd} and \gls{nnaa} results comparing vanilla \gls{tabpfn} with original ordering versus \gls{dag}-aware generation. In \gls{kmtvd}, \gls{dag}-aware generation yields \num{30} significant improvements and \num{1} degradation on \gls{clb} at \num{500} samples (left panel). In \gls{nnaa}, it yields \num{29} significant improvements with no degradations (right panel).

\FloatBarrier
\section{DAG-Aware Generation vs Vanilla under Topological Ordering}
\label{sec:appendix_dag_vs_vanilla_topo}

We compare \gls{dag}-aware generation and vanilla \gls{tabpfn} when both use topological ordering, isolating the contribution of causal parent-based conditioning from the feature ordering itself.

\begin{figure}[!htbp]
  \centering
  \begin{adjustbox}{max width=\linewidth, max totalheight=0.48\textheight}
    \includegraphics{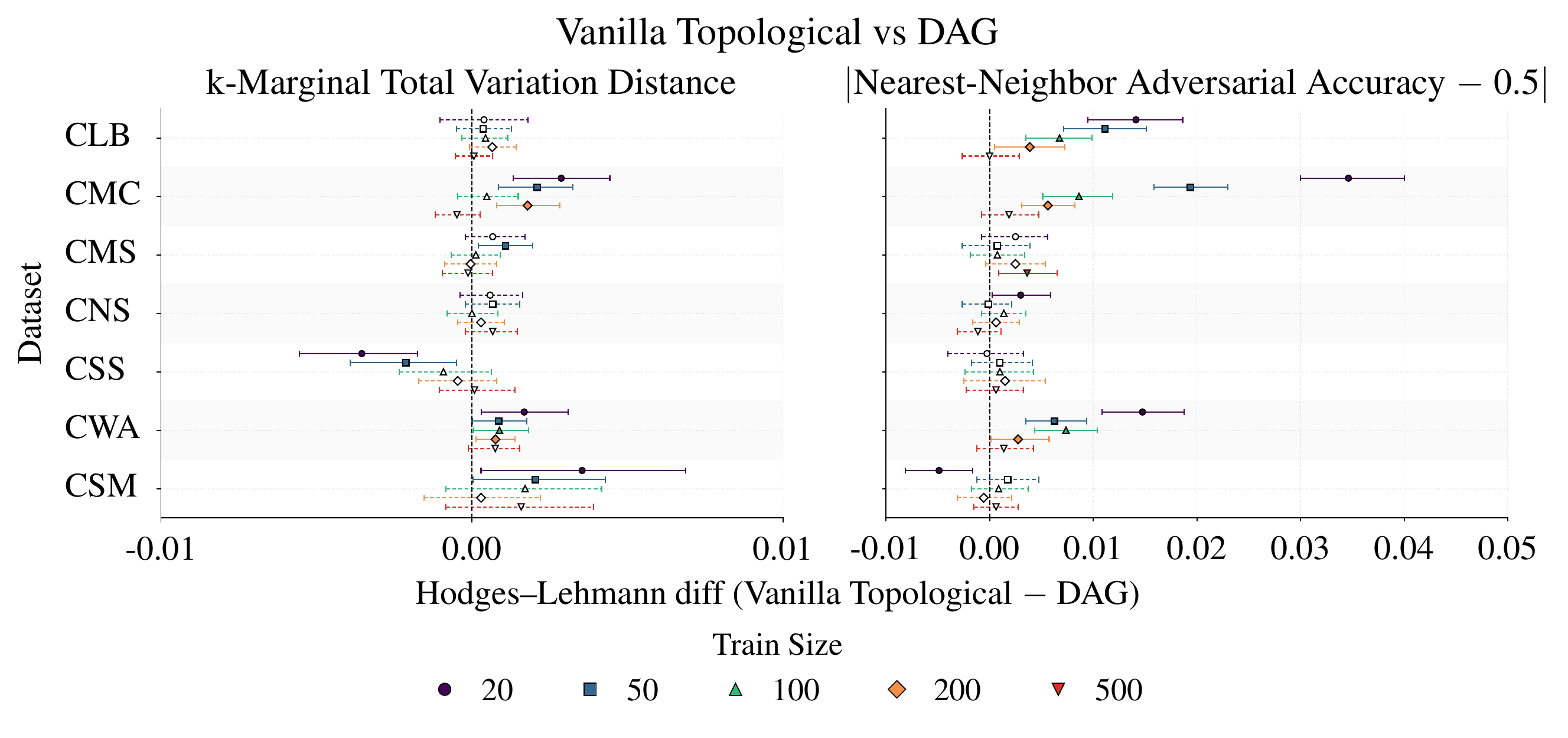}
  \end{adjustbox}
  \caption{Hodges--Lehmann estimates comparing \gls{dag}-aware generation and vanilla \gls{tabpfn}, both using topological ordering, in \gls{kmtvd} ($k=2$, left) and \gls{nnaa} (right), reported as the distance from the ideal value, $\lvert\mathrm{NNAA} - 0.5\rvert$. Positive values indicate that \gls{dag}-aware generation achieves lower \gls{kmtvd} (i.e., better pairwise fidelity) and lower $\lvert\mathrm{NNAA} - 0.5\rvert$ (i.e., greater indistinguishability between synthetic and real data).
  Filled markers with solid error bars indicate significance at $p<0.05$ (Holm correction).}
  \label{fig:forest_dag_topo_vanilla_topo_combined}
\end{figure}

\gls{dag}-aware generation shows \num{10} improvements and \num{2} degradations in \gls{kmtvd} and \num{14} improvements and \num{1} degradation in \gls{nnaa} (\Cref{fig:forest_dag_topo_vanilla_topo_combined}).

\FloatBarrier
\section{\gls{pdag}-Based Generation: \gls{cmd}, \gls{kmtvd}, and \gls{nnaa} Results}
\label{sec:appendix_cpdag}

\begin{figure}[!htbp]
  \centering
  \includegraphics[width=\linewidth, height=0.48\textheight, keepaspectratio]{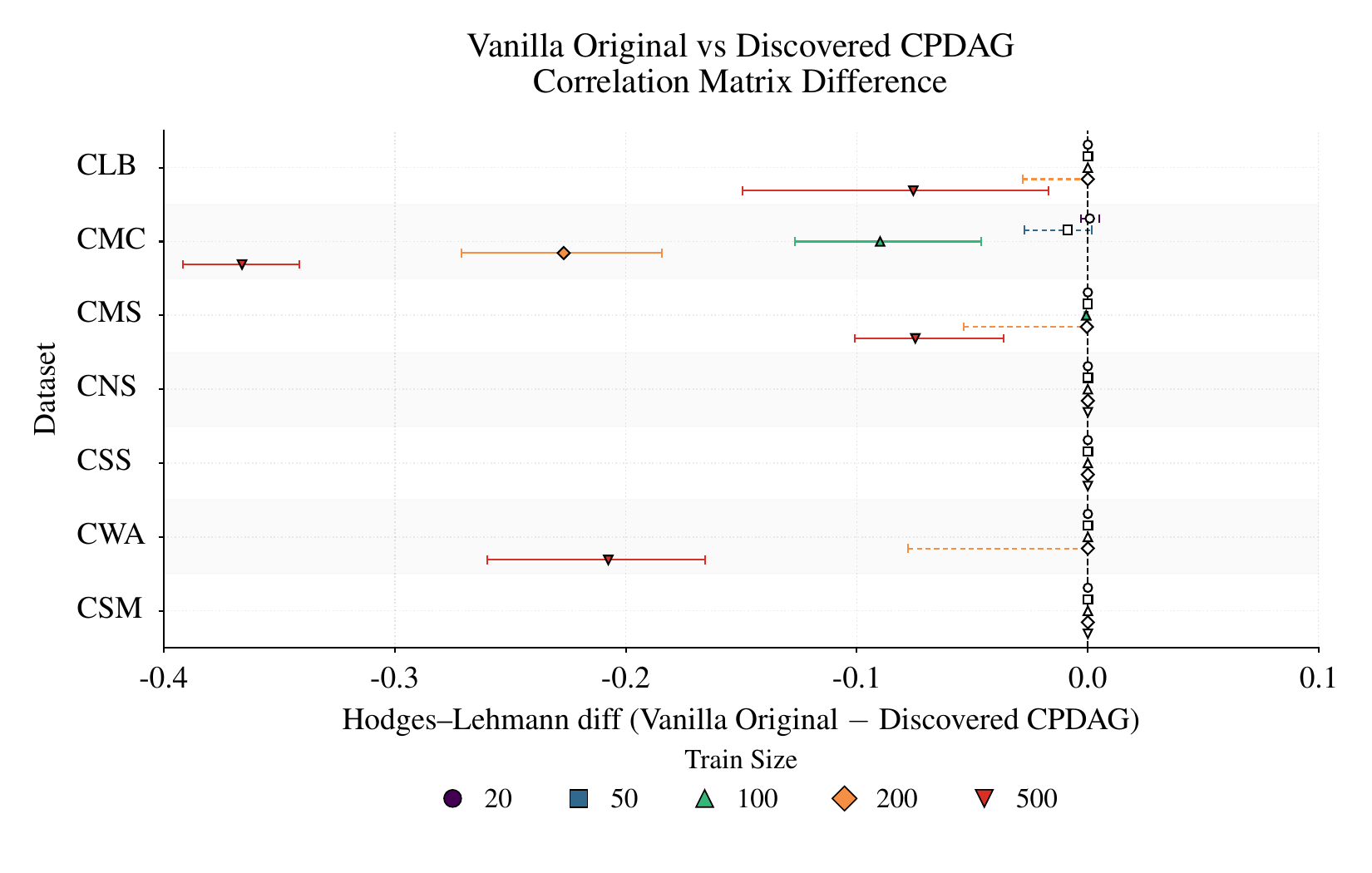}
  \caption{Hodges--Lehmann estimates comparing generation with the discovered \gls{cpdag} and vanilla \gls{tabpfn} with original ordering in \gls{cmd}.
  Positive values indicate that it achieves lower metric values (i.e., better synthetic data quality).
  Filled markers with solid error bars indicate significance at $p<0.05$ (Holm correction).}
  \label{fig:forest_cpdag_discovered_cmd}
\end{figure}

\begin{figure}[!htbp]
  \centering
  \begin{adjustbox}{max width=\linewidth, max totalheight=0.48\textheight}
  \includegraphics{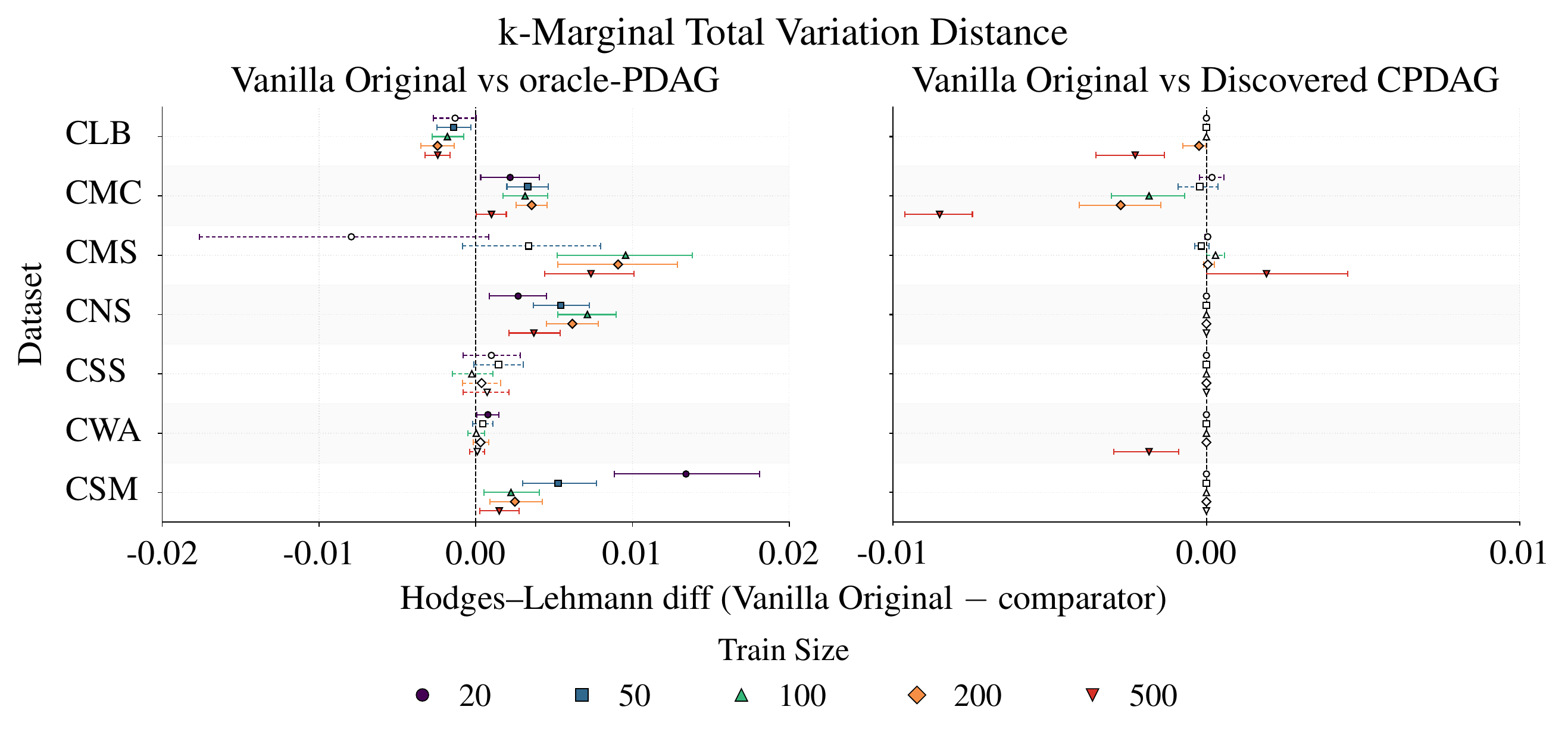}
  \end{adjustbox}
  \caption{Hodges--Lehmann estimates comparing \gls{pdag}-based generation and vanilla \gls{tabpfn} with original ordering in \gls{kmtvd} ($k=2$), with \gls{opdag} (left) and discovered \gls{cpdag} (right).
  Positive values indicate that \gls{pdag}-based generation achieves lower metric values (i.e., better synthetic data quality).
  Filled markers with solid error bars indicate significance at $p<0.05$ (Holm correction).}
  \label{fig:forest_cpdag_combined_2marg}
\end{figure}

\begin{figure}[!htbp]
  \centering
  \begin{adjustbox}{max width=\linewidth, max totalheight=0.48\textheight}
  \includegraphics{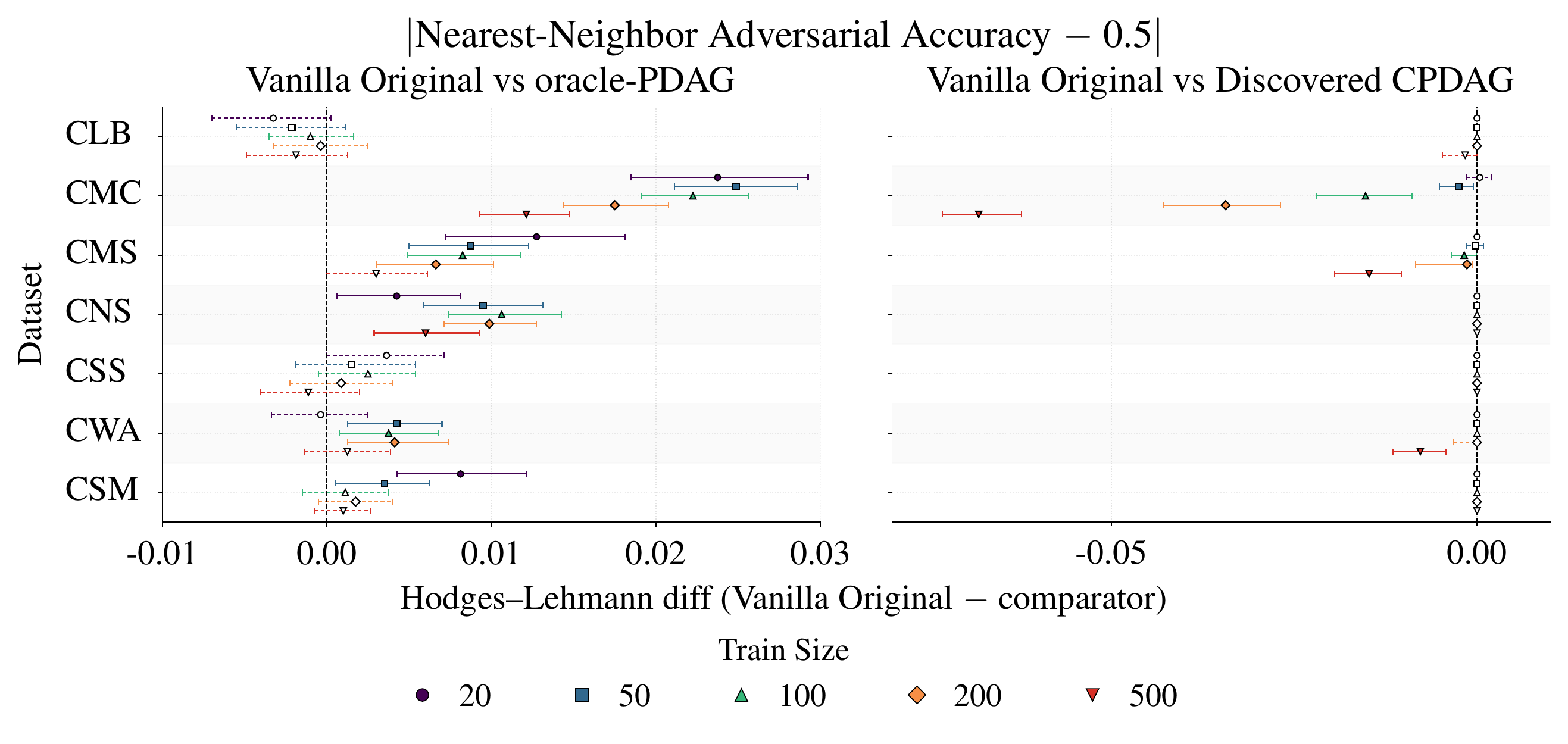}
  \end{adjustbox}
  \caption{Hodges--Lehmann estimates comparing \gls{pdag}-based generation and vanilla \gls{tabpfn} with original ordering in \gls{nnaa} (reported as the distance from the ideal value, $\lvert\mathrm{NNAA} - 0.5\rvert$), with \gls{opdag} (left) and discovered \gls{cpdag} (right).
  Positive values indicate that \gls{pdag}-based generation achieves lower values (i.e., greater indistinguishability between synthetic and real data).
  Filled markers with solid error bars indicate significance at $p<0.05$ (Holm correction).}
  \label{fig:forest_cpdag_combined_nnaa}
\end{figure}

\Cref{fig:forest_cpdag_discovered_cmd} reports \gls{cmd} results for discovered \gls{cpdag}, showing \num{7} degradations with no improvements. Results for \gls{opdag} in \gls{cmd} appear in \cref{fig:forest_dag_cpdag_opdag_cmd}.
\Cref{fig:forest_cpdag_combined_2marg} reports \gls{kmtvd} results for \gls{pdag}-based generation. In \gls{kmtvd}, \gls{opdag} shows \num{19} significant improvements and \num{4} degradations, all on \gls{clb}; discovered \gls{cpdag} shows \num{1} improvement and \num{6} degradations.
The single improvement occurs on \gls{cms} at $N = 500$, where PC orients \SI{52}{\percent} of discovered edges with \num{0.67} direction precision (\cref{tab:pc_discovery_metrics}). Three of the six degradations occur on \gls{cmc}, where PC orients most discovered edges with low accuracy.
\Cref{fig:forest_cpdag_combined_nnaa} reports \gls{nnaa} results.
In \gls{nnaa}, \gls{opdag} shows \num{19} improvements with no degradations; discovered \gls{cpdag} shows no improvements and \num{8} degradations.

\Cref{tab:pc_discovery_metrics} reports graph discovery quality for the \gls{pc}-stable algorithm. Direction recall does not exceed \num{0.32} across all datasets and training sizes. On most datasets, PC leaves the majority of edges undirected, causing the hybrid strategy to fall back to vanilla conditioning. On \gls{cmc}, PC orients up to \SI{86}{\percent} of discovered edges, but direction precision drops to \num{0.19} at $N = 500$, leading the strategy to condition on incorrect parents and producing the degradations observed in the main results.

\begin{table}[!htbp]
\centering
\caption{PC-stable graph discovery quality on CSuite datasets and the custom SCM, averaged over \num{100} repetitions. Skeleton recall measures the fraction of true edges recovered. Direction recall measures the fraction of true edge orientations correctly identified. Oriented fraction reports the proportion of discovered edges that PC orients. Direction precision measures the fraction of oriented edges with correct orientation ({--} indicates no oriented edges).}
\label{tab:pc_discovery_metrics}
\begin{tabular}{l r S[table-format=1.2] S[table-format=1.2] S[table-format=1.2] S[table-format=1.2]}
\toprule
Dataset & {$N$} & {Skel.\ rec.} & {Dir.\ rec.} & {Orient.\ frac.} & {Dir.\ prec.} \\
\midrule
\multirow{5}{*}{CSM}
  &  20 & 0.66 & 0.00 & 0.02 & 0.00 \\
  &  50 & 0.68 & 0.00 & 0.05 & 0.00 \\
  & 100 & 0.68 & 0.00 & 0.03 & 0.00 \\
  & 200 & 0.69 & 0.00 & 0.06 & 0.00 \\
  & 500 & 0.68 & 0.00 & 0.03 & 0.00 \\
\midrule
\multirow{5}{*}{CLB}
  &  20 & 0.08 & 0.00 & 0.01 & 0.00 \\
  &  50 & 0.11 & 0.00 & 0.05 & 0.00 \\
  & 100 & 0.11 & 0.00 & 0.11 & 0.02 \\
  & 200 & 0.12 & 0.00 & 0.18 & 0.01 \\
  & 500 & 0.11 & 0.00 & 0.19 & 0.01 \\
\midrule
\multirow{5}{*}{CMC}
  &  20 & 0.08 & 0.01 & 0.15 & 0.60 \\
  &  50 & 0.19 & 0.07 & 0.42 & 0.63 \\
  & 100 & 0.28 & 0.07 & 0.56 & 0.31 \\
  & 200 & 0.34 & 0.09 & 0.72 & 0.25 \\
  & 500 & 0.40 & 0.11 & 0.86 & 0.19 \\
\midrule
\multirow{5}{*}{CMS}
  &  20 & 0.33 & 0.00 & 0.01 & 0.50 \\
  &  50 & 0.56 & 0.01 & 0.03 & 0.75 \\
  & 100 & 0.59 & 0.06 & 0.13 & 0.81 \\
  & 200 & 0.67 & 0.15 & 0.25 & 0.89 \\
  & 500 & 0.81 & 0.32 & 0.52 & 0.67 \\
\midrule
\multirow{5}{*}{CNS}
  &  20 & 0.24 & 0.00 & 0.00 & {--} \\
  &  50 & 0.26 & 0.00 & 0.03 & 0.00 \\
  & 100 & 0.28 & 0.00 & 0.06 & 0.04 \\
  & 200 & 0.29 & 0.00 & 0.07 & 0.00 \\
  & 500 & 0.44 & 0.00 & 0.00 & {--} \\
\midrule
\multirow{5}{*}{CSS}
  &  20 & 0.16 & 0.00 & 0.00 & {--} \\
  &  50 & 0.30 & 0.00 & 0.04 & 0.00 \\
  & 100 & 0.31 & 0.00 & 0.01 & 0.00 \\
  & 200 & 0.33 & 0.00 & 0.00 & {--} \\
  & 500 & 0.56 & 0.02 & 0.05 & 0.40 \\
\midrule
\multirow{5}{*}{CWA}
  &  20 & 0.05 & 0.00 & 0.00 & {--} \\
  &  50 & 0.24 & 0.01 & 0.10 & 0.34 \\
  & 100 & 0.33 & 0.01 & 0.06 & 0.43 \\
  & 200 & 0.38 & 0.02 & 0.12 & 0.35 \\
  & 500 & 0.40 & 0.07 & 0.23 & 0.48 \\
\bottomrule
\end{tabular}
\end{table}

\FloatBarrier

\section{Comparison with External Tabular Generators}
\label{sec:appendix_external_baselines}
We compare \gls{tabpfn}-based generation against five external tabular generators. TabularARGN is an autoregressive generator trained over randomized feature orderings, CTGAN is a GAN-based generator, DATGAN and DECAF are causal-aware generators to which we supply the true custom \gls{scm} \gls{dag}, and CausalDiffTab is a causal-aware diffusion model~\citep{zhang2025causaldifftab} run with its official defaults. Each baseline is trained per dataset on the same cached training splits used by the comparison experiments and evaluated with the same metrics on the same global test sets, with \num{100} repetitions per combination paired one-to-one with the \gls{tabpfn} runs. \Gls{tabpfn}-based generation, by contrast, requires no training. The comparison targets the small-sample setting ($N \leq 500$) in which our method operates.

TabularARGN, the closest practical autoregressive baseline, was run on five datasets and five training sizes: \gls{clb}, \gls{cmc}, \gls{css}, the custom \gls{scm}, and its noise variant CSMn2 ($\sigma = 10^{-2}$, \Cref{sec:appendix_noise}). \Gls{tabpfn} achieves better median \gls{cmd}, \gls{kmtvd}, and $\lvert\mathrm{NNAA} - 0.5\rvert$ in all \num{25} dataset--size combinations, for both vanilla generation with original ordering and \gls{dag}-aware generation, and every paired comparison is significant (Wilcoxon-Pratt with Holm correction).

We run the full five-generator comparison on the custom \gls{scm}, the controlled collider setting that isolates the failure mode, while the broader multi-dataset check above is limited to TabularARGN, the only autoregressive baseline and thus the most directly comparable to \gls{tabpfn}. \Cref{tab:external_baselines_custom_scm} reports the custom \gls{scm} comparison for all five baselines. \Gls{dag}-aware \gls{tabpfn} achieves significantly lower values in all \num{75} paired contrasts (five baselines, three metrics, five training sizes), and the paired median difference favors \gls{tabpfn} in all \num{150} contrasts overall. Against vanilla \gls{tabpfn}, \num{68} of \num{75} contrasts are significant. The seven non-significant ones all concern \gls{cmd} for the two causal-aware baselines that approach vanilla's correlation quality at small and medium training sizes (DATGAN at $N \leq 200$ and CausalDiffTab at $50 \leq N \leq 200$). Even when supplied with the true \gls{dag} and trained on the target data, no external generator outperforms vanilla \gls{tabpfn} in any of these paired comparisons. Our repository provides modular code for testing further models.

\begin{table}[!htbp]
\centering
\caption{Median synthetic data quality on the custom \gls{scm} over \num{100} repetitions per combination, paired one-to-one with the \gls{tabpfn} runs. Lower is better for all metrics. \Gls{nnaa} is reported as the distance from the ideal value, $\lvert\mathrm{NNAA} - 0.5\rvert$. \dag\ marks generators that receive the true custom \gls{scm} \gls{dag}. CausalDiffTab estimates causal structure internally (official defaults). TabularARGN and CTGAN receive no causal information. All external baselines are trained on each training split, whereas the \gls{tabpfn} rows use inference-time conditioning only. The best value per column is in \textbf{bold}.}
\label{tab:external_baselines_custom_scm}
\begin{tabular}{l S[table-format=1.3,detect-weight,detect-family] S[table-format=1.3,detect-weight,detect-family] S[table-format=1.3,detect-weight,detect-family] S[table-format=1.3,detect-weight,detect-family] S[table-format=1.3,detect-weight,detect-family]}
\toprule
& \multicolumn{5}{c}{Training size $N$} \\
\cmidrule(lr){2-6}
Method & {20} & {50} & {100} & {200} & {500} \\
\midrule
\multicolumn{6}{l}{\textit{\gls{cmd}}} \\
TabularARGN & 2.202 & 2.210 & 2.209 & 2.212 & 2.214 \\
CTGAN & 1.841 & 1.828 & 1.868 & 2.063 & 2.128 \\
DATGAN\textsuperscript{\dag} & 0.544 & 0.278 & 0.240 & 0.180 & 0.261 \\
DECAF\textsuperscript{\dag} & 2.216 & 2.259 & 3.494 & 3.485 & 2.796 \\
CausalDiffTab & 1.048 & 0.313 & 0.231 & 0.195 & 0.207 \\
\addlinespace
\gls{tabpfn} vanilla (original) & 0.471 & 0.285 & 0.196 & 0.154 & 0.120 \\
\gls{tabpfn} \gls{dag}-aware\textsuperscript{\dag} & \bfseries 0.130 & \bfseries 0.135 & \bfseries 0.085 & \bfseries 0.086 & \bfseries 0.074 \\
\midrule
\multicolumn{6}{l}{\textit{\gls{kmtvd}}} \\
TabularARGN & 0.862 & 0.631 & 0.522 & 0.472 & 0.449 \\
CTGAN & 0.517 & 0.496 & 0.502 & 0.598 & 0.593 \\
DATGAN\textsuperscript{\dag} & 0.503 & 0.459 & 0.446 & 0.435 & 0.413 \\
DECAF\textsuperscript{\dag} & 0.997 & 0.996 & 0.992 & 0.992 & 0.990 \\
CausalDiffTab & 0.524 & 0.343 & 0.285 & 0.265 & 0.231 \\
\addlinespace
\gls{tabpfn} vanilla (original) & 0.277 & 0.244 & 0.225 & 0.221 & 0.214 \\
\gls{tabpfn} \gls{dag}-aware\textsuperscript{\dag} & \bfseries 0.260 & \bfseries 0.235 & \bfseries 0.222 & \bfseries 0.218 & \bfseries 0.213 \\
\midrule
\multicolumn{6}{l}{\textit{$\lvert\mathrm{NNAA} - 0.5\rvert$}} \\
TabularARGN & 0.491 & 0.481 & 0.476 & 0.474 & 0.472 \\
CTGAN & 0.469 & 0.470 & 0.471 & 0.475 & 0.475 \\
DATGAN\textsuperscript{\dag} & 0.457 & 0.447 & 0.431 & 0.409 & 0.419 \\
DECAF\textsuperscript{\dag} & 0.500 & 0.500 & 0.500 & 0.500 & 0.500 \\
CausalDiffTab & 0.466 & 0.237 & 0.136 & 0.096 & 0.058 \\
\addlinespace
\gls{tabpfn} vanilla (original) & 0.055 & 0.023 & 0.013 & 0.011 & 0.007 \\
\gls{tabpfn} \gls{dag}-aware\textsuperscript{\dag} & \bfseries 0.046 & \bfseries 0.016 & \bfseries 0.009 & \bfseries 0.010 & \bfseries 0.006 \\
\bottomrule
\end{tabular}
\end{table}

\clearpage

\section{Discovered \gls{cpdag} in a PC-Recoverable Regime}
\label{sec:appendix_recoverable}
The custom \gls{scm} was designed to expose \gls{tabpfn}'s autoregressive vulnerability. As a side effect, its near-deterministic mechanisms ($\sigma = 10^{-5}$) make the \gls{dag}-implied conditional independencies hard to detect with PC's Fisher--Z tests. Each variable is almost a deterministic function of its parents, so correlations among connected variables approach one and the partial correlations underlying the test degenerate at finite samples. Its weak discovered-\gls{cpdag} results therefore reflect a setting where causal discovery fails, not the quality of the generation itself. To isolate what happens when PC succeeds, we keep the same \gls{dag} and structural coefficients and increase only the \gls{scm} noise to $\sigma = 0.2$ (denoted CSMr in the figures), restoring the conditional variability the test requires while leaving the causal structure unchanged. In a separate discovery screening over \num{120} runs, PC recovers the exact \gls{cpdag} in \SI{98.3}{\percent} of runs at $N = 200$ and \SI{99.2}{\percent} at $N = 500$, and recovery at these training sizes remains high at $\sigma = 0.1$ and $\sigma = 0.5$.

\begin{figure}[!htbp]
\centering
\includegraphics[width=\linewidth, height=0.48\textheight, keepaspectratio]{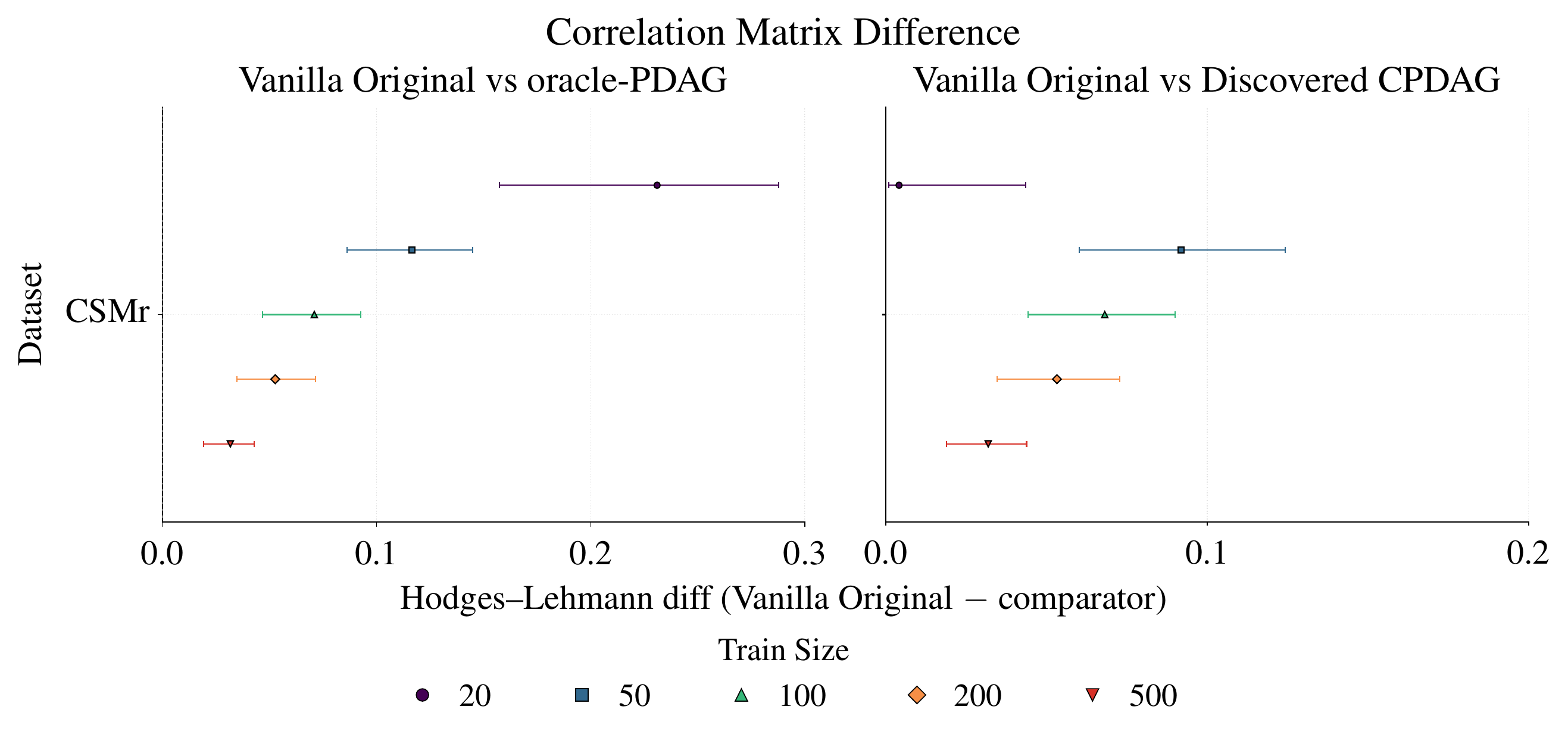}
\caption{Hodges--Lehmann estimates comparing vanilla \gls{tabpfn} with original ordering versus \gls{opdag} (left) and discovered \gls{cpdag} (right) on the PC-recoverable custom \gls{scm} variant (CSMr, $\sigma = 0.2$), in \gls{cmd}.
Positive values indicate that \gls{pdag}-based generation achieves lower metric values (i.e., better synthetic data quality).
Filled markers with solid error bars indicate significance at $p<0.05$ (Holm correction).}
\label{fig:forest_recoverable_cmd}
\end{figure}

\begin{figure}[!htbp]
\centering
\includegraphics[width=\linewidth, height=0.48\textheight, keepaspectratio]{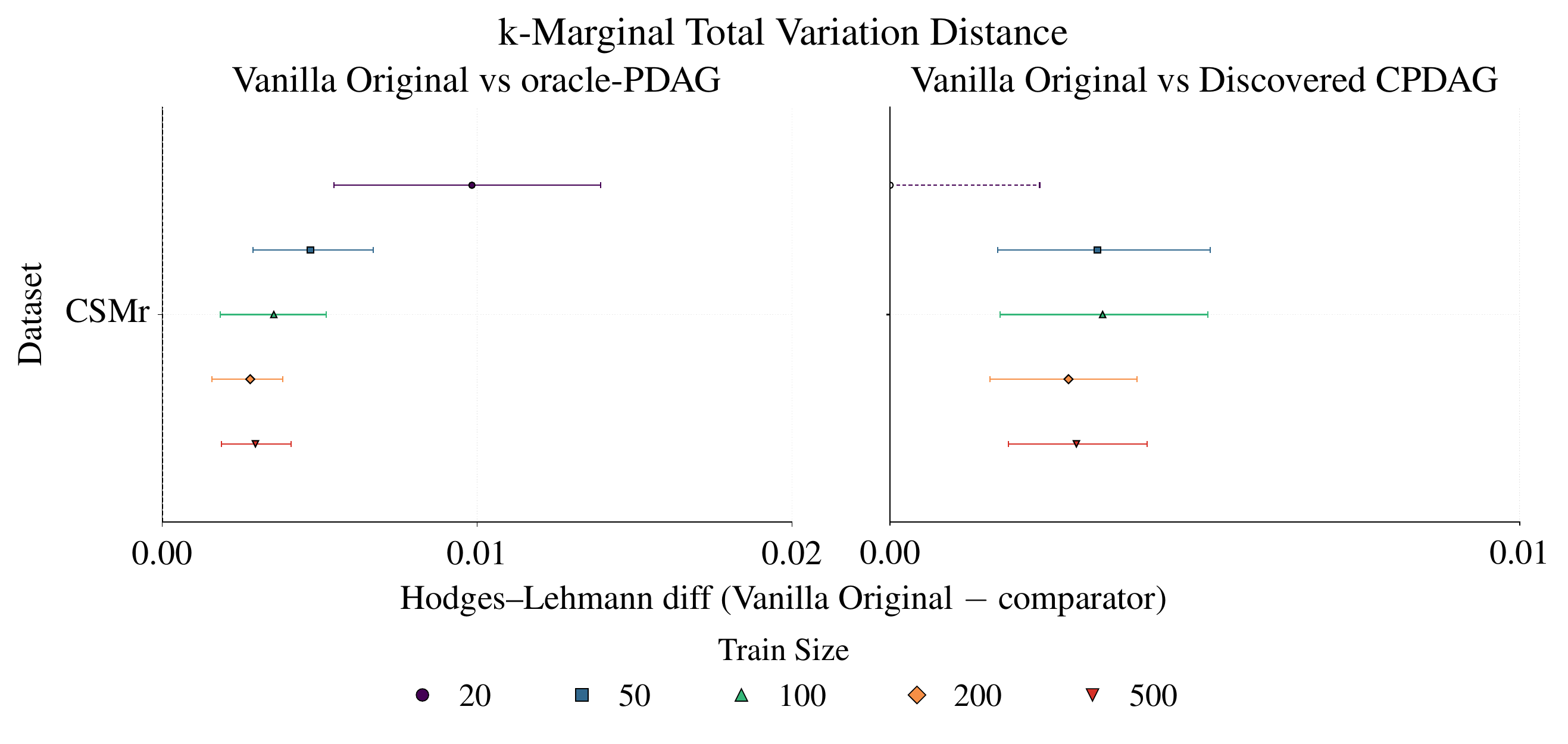}
\caption{Hodges--Lehmann estimates comparing vanilla \gls{tabpfn} with original ordering versus \gls{opdag} (left) and discovered \gls{cpdag} (right) on the PC-recoverable custom \gls{scm} variant (CSMr, $\sigma = 0.2$), in \gls{kmtvd} ($k=2$).
Positive values indicate that \gls{pdag}-based generation achieves lower metric values (i.e., better synthetic data quality).
Filled markers with solid error bars indicate significance at $p<0.05$ (Holm correction).}
\label{fig:forest_recoverable_kmtvd}
\end{figure}

\begin{figure}[!htbp]
\centering
\includegraphics[width=\linewidth, height=0.48\textheight, keepaspectratio]{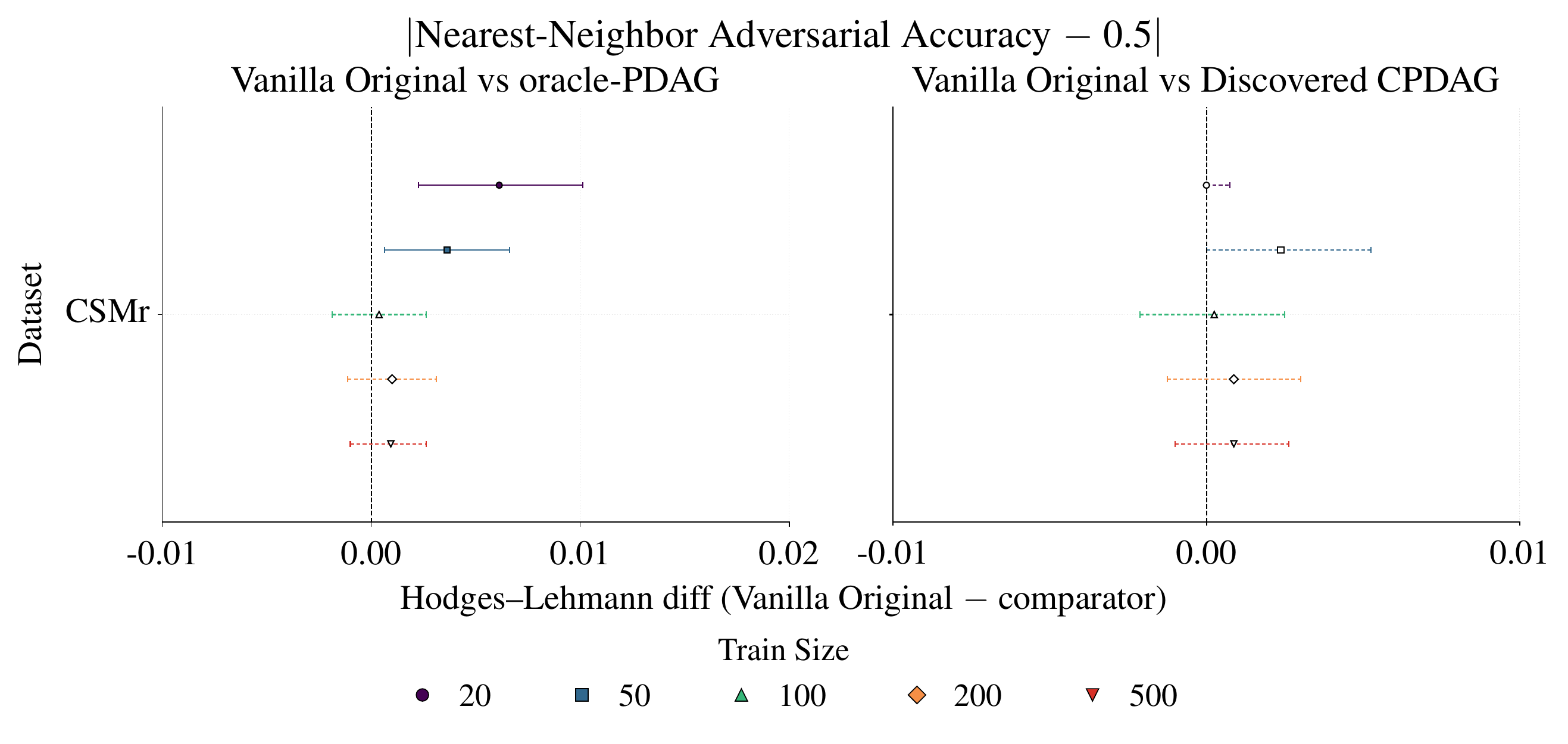}
\caption{Hodges--Lehmann estimates comparing vanilla \gls{tabpfn} with original ordering versus \gls{opdag} (left) and discovered \gls{cpdag} (right) on the PC-recoverable custom \gls{scm} variant (CSMr, $\sigma = 0.2$), in \gls{nnaa} (reported as the distance from the ideal value, $\lvert\mathrm{NNAA} - 0.5\rvert$).
Positive values indicate that \gls{pdag}-based generation achieves lower values (i.e., greater indistinguishability between synthetic and real data).
Filled markers with solid error bars indicate significance at $p<0.05$ (Holm correction).}
\label{fig:forest_recoverable_nnaa}
\end{figure}

\begin{figure}[!htbp]
\centering
\includegraphics[width=\linewidth, height=0.48\textheight, keepaspectratio]{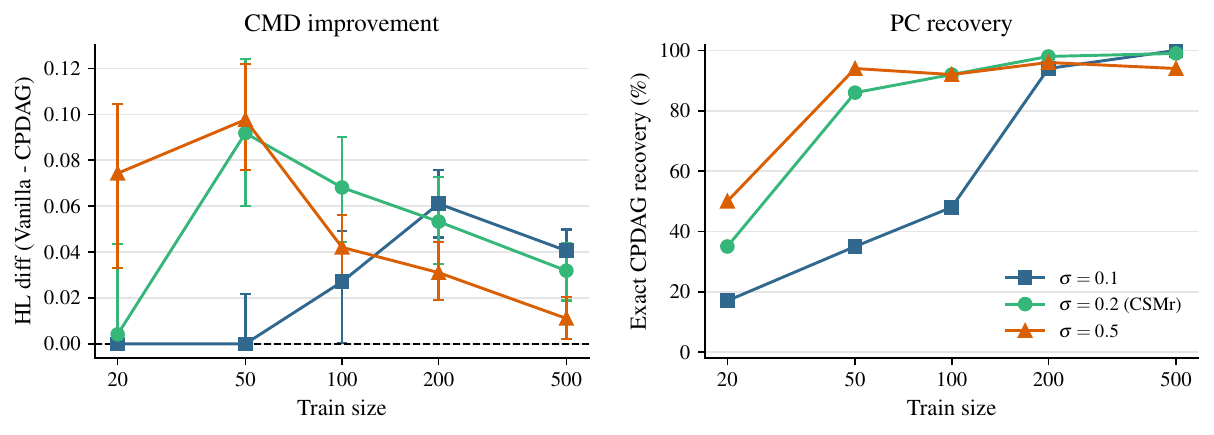}
\caption{Discovered-\gls{cpdag} generation versus PC recovery on the custom \gls{scm} at noise levels $\sigma \in \{0.1, 0.2, 0.5\}$. The left panel shows the paired Hodges--Lehmann \gls{cmd} improvement of discovered-\gls{cpdag} generation over vanilla \gls{tabpfn} with original ordering as a function of training size, over \num{100} paired repetitions per combination. Values above zero favor the discovered \gls{cpdag}. Error bars show Holm-adjusted confidence intervals. The right panel shows the fraction of the same runs in which PC recovers the exact \gls{cpdag}.}
\label{fig:recoverable_robustness}
\end{figure}

In this setting causal discovery helps synthetic data generation. Discovered-\gls{cpdag} generation significantly improves over vanilla original in \gls{cmd} at all five training sizes and in \gls{kmtvd} at four of five ($N \geq 50$), closely matching the \gls{opdag} reference for $N \geq 100$ (\Cref{fig:forest_recoverable_cmd,fig:forest_recoverable_kmtvd,fig:forest_recoverable_nnaa}). \Gls{nnaa} differences are never significant. To verify that this result is not specific to $\sigma = 0.2$, we repeat the same paired comparison at $\sigma = 0.1$ and $\sigma = 0.5$. \Cref{fig:recoverable_robustness} shows the \gls{cmd} improvement of the discovered \gls{cpdag} over vanilla original alongside the fraction of runs in which PC recovers the exact \gls{cpdag}. The improvement rises with the recovery rate at every noise level. At $\sigma = 0.1$ with $N \leq 50$, where PC recovers the exact \gls{cpdag} in about a third of the runs or fewer, the improvement is close to zero. Where recovery is high, the improvement is clearly positive, and it shrinks at large $N$ as vanilla \gls{tabpfn} improves on its own.

\FloatBarrier

\clearpage

\section{Vanilla TabPFN with Topological Ordering Results on \gls{ate} Preservation: \gls{sgl}}
\label{sec:appendix_topological_ate}

\begin{figure}[!htbp]
    \centering
    \includegraphics[width=\linewidth, height=0.48\textheight, keepaspectratio]{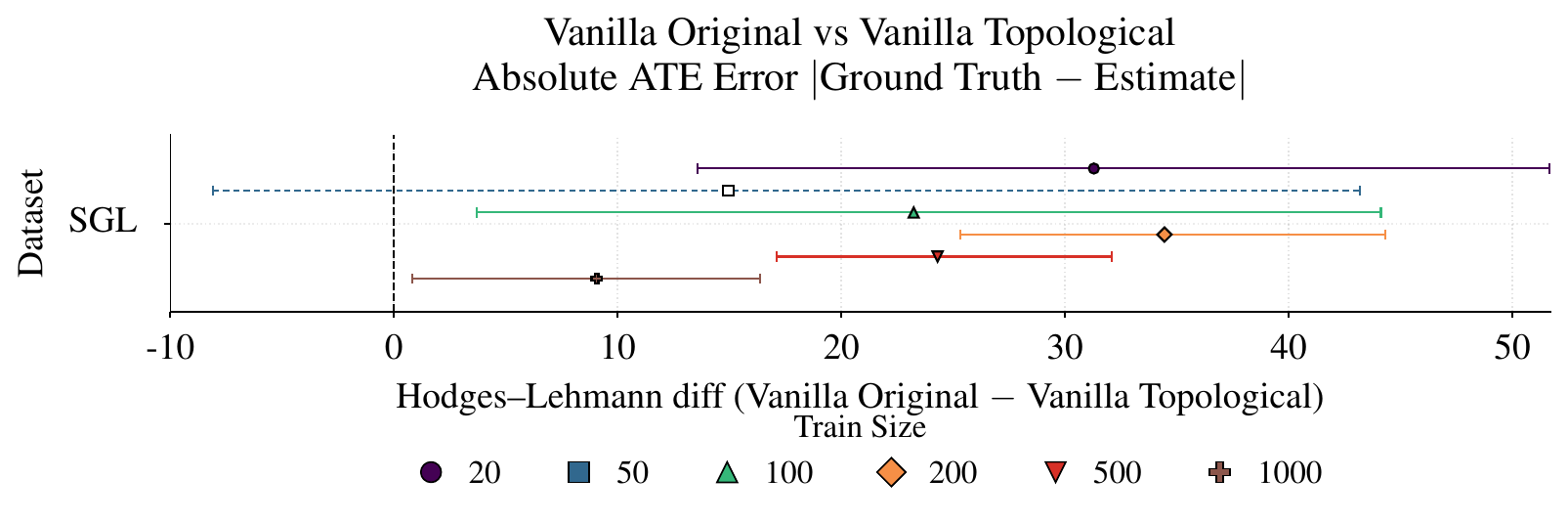}
    \caption{Hodges--Lehmann estimates of the reduction in absolute \gls{ate} error ($\Delta_{\text{ATE}}$) when comparing vanilla \gls{tabpfn} with original ordering versus topological ordering on \gls{sgl}.
    Positive values indicate smaller errors (closer to ground truth) for topological ordering; negative values indicate larger errors.
    Filled markers with solid error bars indicate significance at $p<0.05$ (Holm correction).}
    \label{fig:forest_vanilla_topological_ate_sgl}
\end{figure}

\Cref{fig:forest_vanilla_topological_ate_sgl} reports \gls{ate} preservation results for vanilla \gls{tabpfn} with topological ordering on \gls{sgl}.
Topological ordering produces \num{5} significant improvements out of \num{6} training sizes, with larger effect sizes than the CSuite datasets (\Cref{fig:forest_vanilla_topological_ate}).

\FloatBarrier
\section{Vanilla TabPFN with Reverse Topological Ordering Results on \gls{ate} Preservation}
\label{sec:appendix_reverse_ate}

\begin{figure}[!htbp]
\centering
\includegraphics[width=\linewidth, height=0.48\textheight, keepaspectratio]{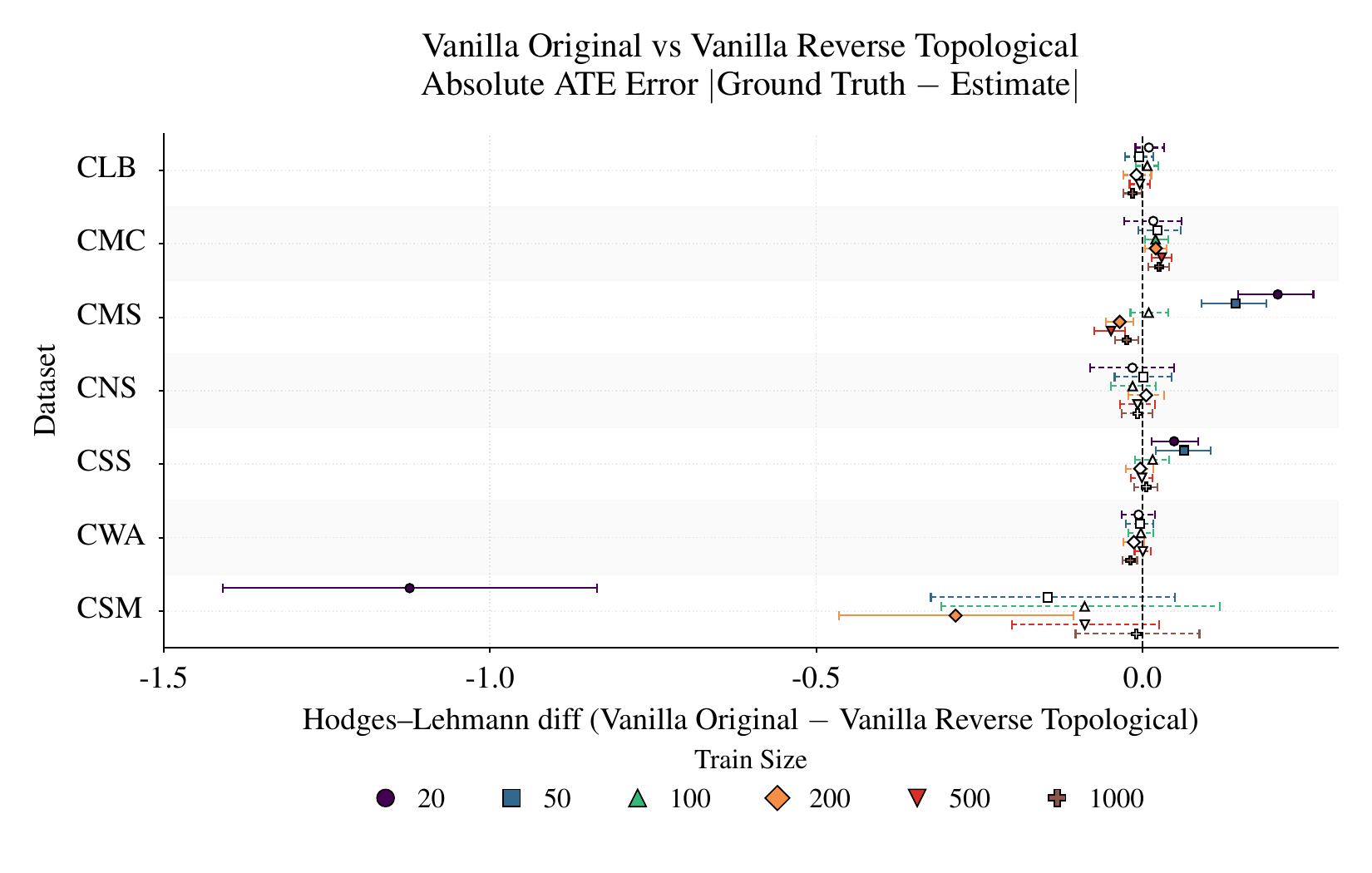}
\caption{Hodges--Lehmann estimates of the reduction in absolute \gls{ate} error ($\Delta_{\text{ATE}}$) when comparing vanilla \gls{tabpfn} with original ordering versus reverse topological ordering. 
Positive values indicate smaller errors (closer to ground truth) for reverse topological ordering; negative values indicate larger errors.
Filled markers with solid error bars indicate significance at $p<0.05$ (Holm correction).}
\label{fig:forest_vanilla_worst_ate}
\end{figure}

\Cref{fig:forest_vanilla_worst_ate} reports results for reverse topological ordering.
Reverse topological ordering shows \num{8} significant improvements and \num{7} degradations.
The largest degradation occurs on \gls{csm} at \num{20} samples.

\FloatBarrier
\section{DAG-Aware Generation vs Vanilla under Topological Ordering on \gls{ate} Preservation}
\label{sec:appendix_dag_vs_vanilla_topo_ate}

We compare \gls{dag}-aware generation and vanilla \gls{tabpfn} on \gls{ate} preservation when both use topological ordering, isolating the contribution of causal parent-based conditioning from the feature ordering itself.

\begin{figure}[!htbp]
\centering
\includegraphics[width=\linewidth, height=0.48\textheight, keepaspectratio]{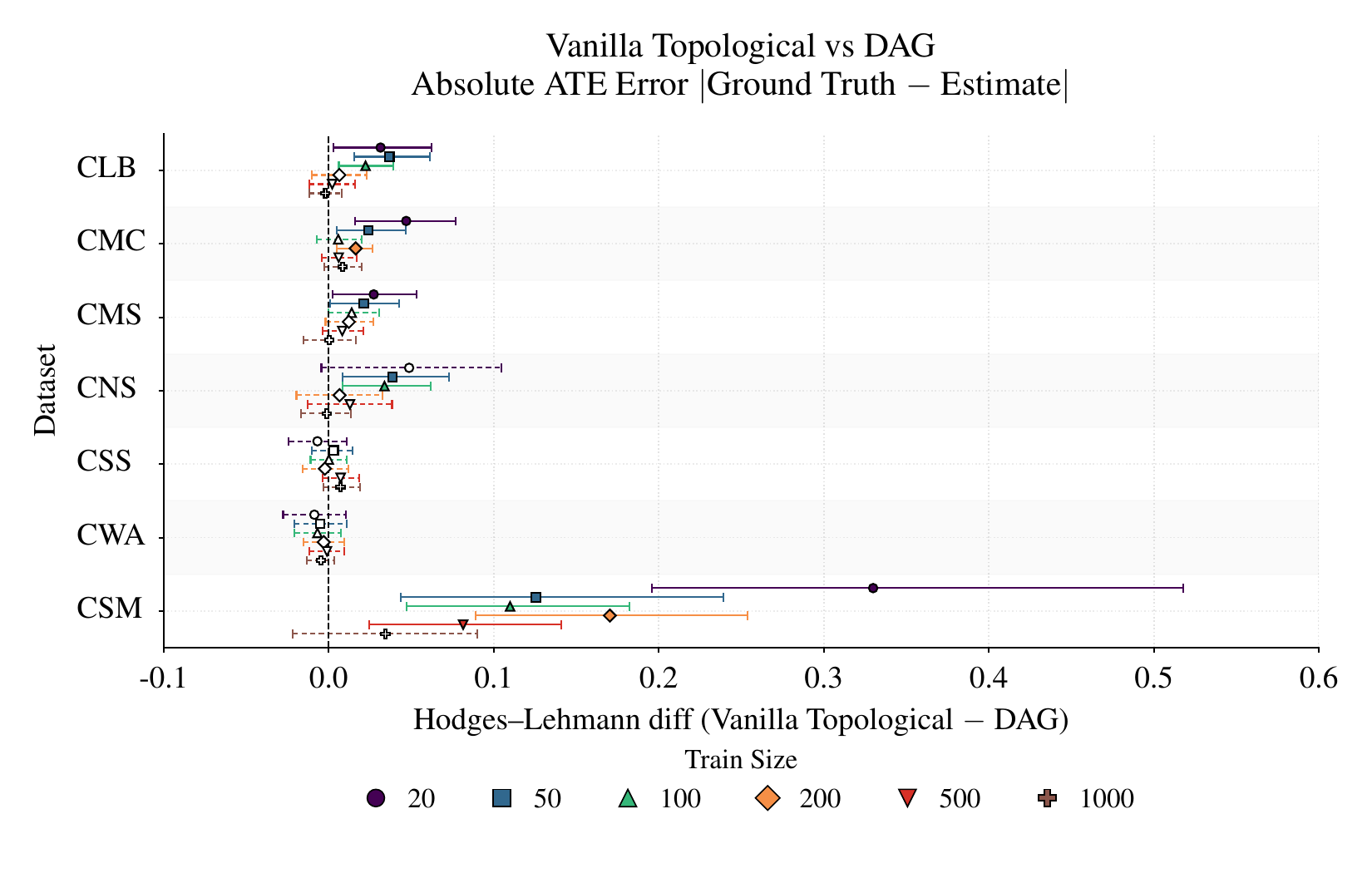}
\caption{Hodges--Lehmann estimates of the reduction in absolute \gls{ate} error ($\Delta_{\text{ATE}}$) when comparing vanilla \gls{tabpfn} with topological ordering versus \gls{dag}-aware generation.
Positive values indicate smaller errors (closer to ground truth) for \gls{dag}-aware generation; negative values indicate larger errors.
Filled markers with solid error bars indicate significance at $p<0.05$ (Holm correction).}
\label{fig:forest_dag_topo_vanilla_topo_ate}
\end{figure}

\gls{dag}-aware generation shows \num{15} significant improvements with no degradations (\cref{fig:forest_dag_topo_vanilla_topo_ate}).
\gls{csm} shows significant improvements across most training sizes, confirming that causal parent conditioning provides benefits beyond feature ordering alone.

\FloatBarrier
\section{Discovered \gls{cpdag} Results on \gls{ate} Preservation}
\label{sec:appendix_cpdag_discovered_ate}

\begin{figure}[!htbp]
\centering
\includegraphics[width=\linewidth, height=0.48\textheight, keepaspectratio]{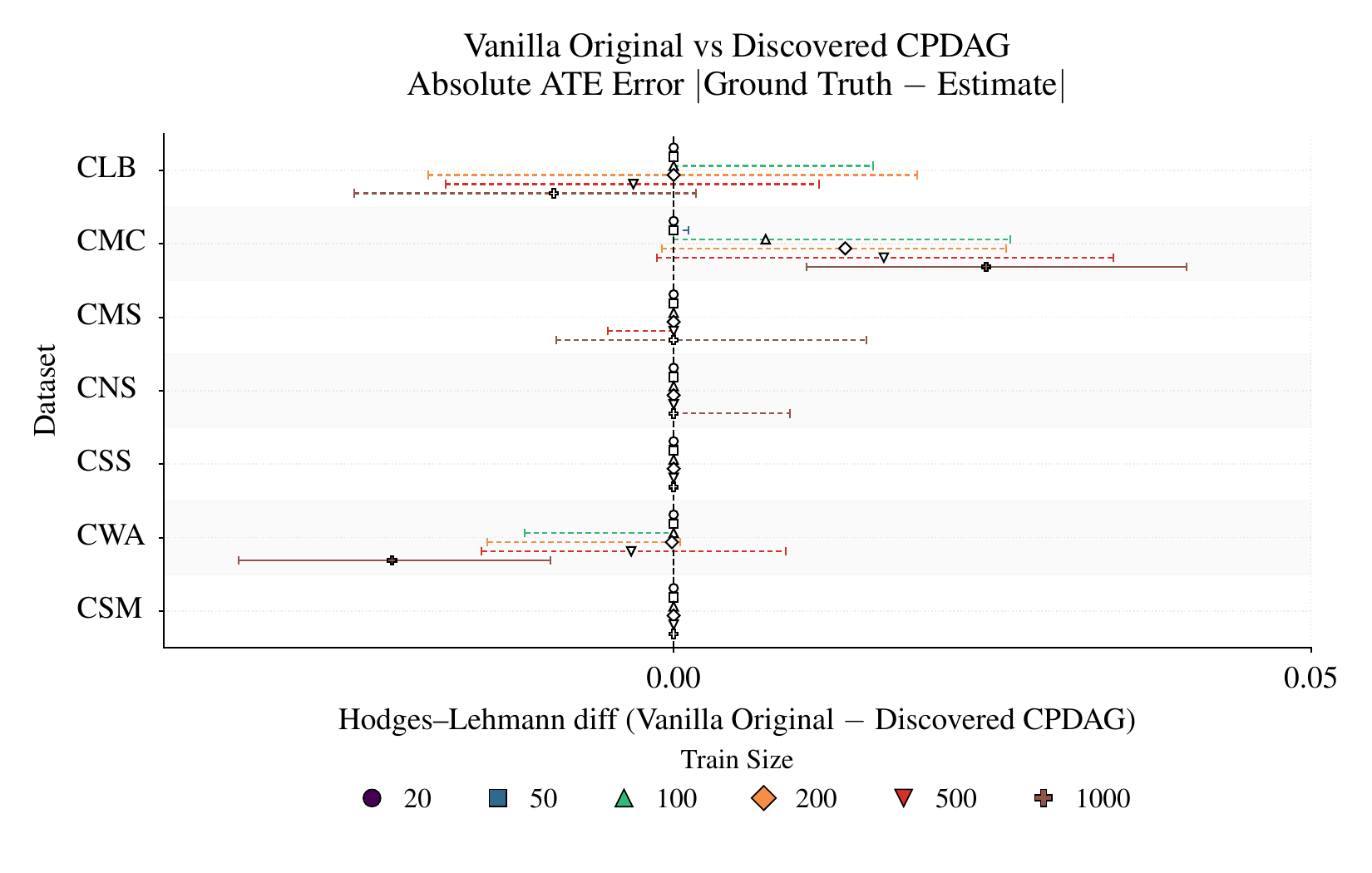}
\caption{Hodges--Lehmann estimates of the reduction in absolute \gls{ate} error ($\Delta_{\text{ATE}}$) when comparing vanilla \gls{tabpfn} with original ordering versus discovered \gls{cpdag}.
Positive values indicate smaller errors (closer to ground truth) for discovered \gls{cpdag}; negative values indicate larger errors.
Filled markers with solid error bars indicate significance at $p<0.05$ (Holm correction).}
\label{fig:forest_cpdag_discovered_ate}
\end{figure}

\Cref{fig:forest_cpdag_discovered_ate} reports \gls{ate} preservation results for discovered \gls{cpdag}.
The discovered \gls{cpdag} shows one significant improvement on \gls{cmc} at $N = \num{1000}$ and one degradation on \gls{cwa} at $N = \num{1000}$.
\Cref{tab:pc_discovery_metrics_ate} reports graph discovery quality against the mutilated \gls{dag} (the causal graph with edges into the treatment variable removed), which is the appropriate reference for interventional data.
On \gls{cmc} at $N = \num{1000}$, PC orients \SI{88}{\percent} of discovered edges with direction precision \num{0.24}, providing enough correct structure around the treatment--outcome path to improve \gls{ate} estimation.
On \gls{cwa} at $N = \num{1000}$, PC orients only \SI{28}{\percent} of edges with direction precision \num{0.02}, causing the few oriented edges to introduce incorrect conditioning around the treatment--outcome relationship.

\begin{table}[!htbp]
\centering
\caption{PC-stable graph discovery quality on CSuite datasets and the custom SCM for the \gls{ate} experiments, averaged over \num{100} repetitions. Metrics are computed against the mutilated \gls{dag} (edges into the treatment variable removed). Skeleton recall measures the fraction of true edges recovered. Direction recall measures the fraction of true edge orientations correctly identified. Oriented fraction reports the proportion of discovered edges that PC orients. Direction precision measures the fraction of oriented edges with correct orientation ({--} indicates no oriented edges).}
\label{tab:pc_discovery_metrics_ate}
\begin{tabular}{l r S[table-format=1.2] S[table-format=1.2] S[table-format=1.2] S[table-format=1.2]}
\toprule
Dataset & {$N$} & {Skel.\ rec.} & {Dir.\ rec.} & {Orient.\ frac.} & {Dir.\ prec.} \\
\midrule
\multirow{6}{*}{CSM}
  &   20 & 0.68 & 0.00 & 0.03 & 0.00 \\
  &   50 & 0.68 & 0.00 & 0.05 & 0.00 \\
  &  100 & 0.68 & 0.00 & 0.03 & 0.00 \\
  &  200 & 0.67 & 0.00 & 0.01 & 0.00 \\
  &  500 & 0.67 & 0.00 & 0.01 & 0.00 \\
  & 1000 & 0.67 & 0.00 & 0.00 & {--} \\
\midrule
\multirow{6}{*}{CLB}
  &   20 & 0.19 & 0.00 & 0.01 & 0.50 \\
  &   50 & 0.53 & 0.01 & 0.08 & 0.29 \\
  &  100 & 0.78 & 0.05 & 0.24 & 0.21 \\
  &  200 & 0.91 & 0.04 & 0.28 & 0.13 \\
  &  500 & 1.00 & 0.04 & 0.32 & 0.07 \\
  & 1000 & 1.00 & 0.03 & 0.31 & 0.03 \\
\midrule
\multirow{6}{*}{CMC}
  &   20 & 0.07 & 0.01 & 0.17 & 0.30 \\
  &   50 & 0.18 & 0.08 & 0.39 & 0.65 \\
  &  100 & 0.22 & 0.11 & 0.60 & 0.52 \\
  &  200 & 0.24 & 0.14 & 0.75 & 0.44 \\
  &  500 & 0.33 & 0.18 & 0.81 & 0.39 \\
  & 1000 & 0.38 & 0.15 & 0.88 & 0.24 \\
\midrule
\multirow{6}{*}{CMS}
  &   20 & 0.27 & 0.01 & 0.19 & 0.20 \\
  &   50 & 0.43 & 0.02 & 0.14 & 0.23 \\
  &  100 & 0.46 & 0.01 & 0.14 & 0.14 \\
  &  200 & 0.64 & 0.02 & 0.38 & 0.07 \\
  &  500 & 0.82 & 0.02 & 0.61 & 0.04 \\
  & 1000 & 0.88 & 0.03 & 0.71 & 0.03 \\
\midrule
\multirow{6}{*}{CNS}
  &   20 & 0.31 & 0.00 & 0.02 & 0.50 \\
  &   50 & 0.40 & 0.00 & 0.00 & {--} \\
  &  100 & 0.56 & 0.00 & 0.06 & 0.13 \\
  &  200 & 0.64 & 0.01 & 0.10 & 0.14 \\
  &  500 & 0.71 & 0.00 & 0.20 & 0.00 \\
  & 1000 & 0.83 & 0.00 & 0.59 & 0.00 \\
\midrule
\multirow{6}{*}{CSS}
  &   20 & 0.38 & 0.01 & 0.05 & 0.67 \\
  &   50 & 0.50 & 0.09 & 0.17 & 1.00 \\
  &  100 & 0.67 & 0.06 & 0.12 & 0.75 \\
  &  200 & 0.67 & 0.00 & 0.02 & 0.00 \\
  &  500 & 0.67 & 0.00 & 0.00 & {--} \\
  & 1000 & 0.67 & 0.00 & 0.01 & 0.00 \\
\midrule
\multirow{6}{*}{CWA}
  &   20 & 0.14 & 0.01 & 0.10 & 0.40 \\
  &   50 & 0.43 & 0.02 & 0.14 & 0.31 \\
  &  100 & 0.50 & 0.01 & 0.15 & 0.12 \\
  &  200 & 0.53 & 0.01 & 0.19 & 0.11 \\
  &  500 & 0.58 & 0.01 & 0.29 & 0.02 \\
  & 1000 & 0.58 & 0.01 & 0.28 & 0.02 \\
\bottomrule
\end{tabular}
\end{table}

\clearpage
\section{Absolute Scale of \gls{ate} Errors}
\label{sec:appendix_ate_scale}
The \gls{ate}-preservation forest plots report paired Hodges--Lehmann reductions in the absolute \gls{ate} error $\Delta_{\text{ATE}}$ without an explicit reference scale. \Cref{tab:ate_scale_reference} complements them by reporting, for each dataset and training size, the absolute value of the true test-set \gls{ate} together with the median $\Delta_{\text{ATE}}$ of vanilla \gls{tabpfn} with original ordering and of the primary comparator. All values are expressed in the native outcome units of each dataset. \Gls{sgl} values are in mg/dL. On the \gls{sgl} simulator, topological ordering lowers the median \gls{ate} error from \num{39.6}--\num{84.9}~mg/dL (vanilla \gls{tabpfn}) to \num{15.9}--\num{64.4}~mg/dL, a paired Hodges--Lehmann reduction of \num{9.06}--\num{34.4}~mg/dL significant at every training size except $N=\num{50}$. These errors remain large relative to the true test-set \gls{ate} of \num{2.29}~mg/dL.

\begin{longtable}{l r r r r r c}
\caption{Absolute-scale reference for \gls{ate} preservation over \num{100} paired repetitions per combination. $|\text{ATE}_{\text{test}}|$ is the absolute true \gls{ate} on the global test set. The error columns report the median absolute \gls{ate} error ($\Delta_{\text{ATE}}$) of vanilla \gls{tabpfn} with original ordering and of the primary comparator. The comparator is \gls{dag}-aware generation with topological ordering, except on \gls{sgl}, where it is vanilla \gls{tabpfn} with topological ordering because no full \gls{dag} is available. HL is the paired Hodges--Lehmann estimate of the error reduction (vanilla minus comparator). The significance flag comes from the paired Wilcoxon signed-rank test (Pratt ties, Holm correction) and matches the corresponding forest plots. A set flag with a negative HL marks a significant worsening.}
\label{tab:ate_scale_reference}\\
\toprule
Dataset & {$N$} & {$|\text{ATE}_{\text{test}}|$} & {Vanilla $\Delta_{\text{ATE}}$} & {Comp.\ $\Delta_{\text{ATE}}$} & {HL} & {Sig.} \\
\midrule
\endfirsthead
\multicolumn{7}{l}{\itshape Table \thetable\ (continued)}\\
\toprule
Dataset & {$N$} & {$|\text{ATE}_{\text{test}}|$} & {Vanilla $\Delta_{\text{ATE}}$} & {Comp.\ $\Delta_{\text{ATE}}$} & {HL} & {Sig.} \\
\midrule
\endhead
\bottomrule
\endlastfoot
\multirow{6}{*}{CSM}
  &   20 & \num{4.98} & \num{1.23} & \num{0.186} & \num{1.23} & Y \\*
  &   50 & \num{4.98} & \num{1.15} & \num{0.183} & \num{1.00} & Y \\*
  &  100 & \num{4.98} & \num{0.826} & \num{0.163} & \num{0.649} & Y \\*
  &  200 & \num{4.98} & \num{0.343} & \num{0.162} & \num{0.268} & Y \\*
  &  500 & \num{4.98} & \num{0.274} & \num{0.152} & \num{0.14} & Y \\*
  & 1000 & \num{4.98} & \num{0.243} & \num{0.151} & \num{0.119} & Y \\
\midrule
\multirow{6}{*}{CLB}
  &   20 & \num{0.593} & \num{0.402} & \num{0.328} & \num{0.0336} & Y \\*
  &   50 & \num{0.593} & \num{0.271} & \num{0.221} & \num{0.0433} & Y \\*
  &  100 & \num{0.593} & \num{0.14} & \num{0.111} & \num{0.03} & Y \\*
  &  200 & \num{0.593} & \num{0.0961} & \num{0.0782} & \num{0.00725} & N \\*
  &  500 & \num{0.593} & \num{0.0556} & \num{0.0533} & \num{0.00443} & N \\*
  & 1000 & \num{0.593} & \num{0.0405} & \num{0.0301} & \num{0.0087} & N \\
\midrule
\multirow{6}{*}{CMC}
  &   20 & \num{1.03} & \num{0.389} & \num{0.479} & \num{-0.063} & Y \\*
  &   50 & \num{1.03} & \num{0.173} & \num{0.105} & \num{0.037} & Y \\*
  &  100 & \num{1.03} & \num{0.129} & \num{0.0873} & \num{0.0286} & Y \\*
  &  200 & \num{1.03} & \num{0.0926} & \num{0.0554} & \num{0.0357} & Y \\*
  &  500 & \num{1.03} & \num{0.0894} & \num{0.0453} & \num{0.0337} & Y \\*
  & 1000 & \num{1.03} & \num{0.0701} & \num{0.0527} & \num{0.0231} & Y \\
\midrule
\multirow{6}{*}{CMS}
  &   20 & \num{0.727} & \num{0.491} & \num{0.133} & \num{0.348} & Y \\*
  &   50 & \num{0.727} & \num{0.445} & \num{0.0742} & \num{0.33} & Y \\*
  &  100 & \num{0.727} & \num{0.192} & \num{0.0677} & \num{0.145} & Y \\*
  &  200 & \num{0.727} & \num{0.094} & \num{0.0496} & \num{0.0511} & Y \\*
  &  500 & \num{0.727} & \num{0.0705} & \num{0.0496} & \num{0.0242} & Y \\*
  & 1000 & \num{0.727} & \num{0.081} & \num{0.0565} & \num{0.0211} & Y \\
\midrule
\multirow{6}{*}{CNS}
  &   20 & \num{1.96} & \num{0.73} & \num{0.289} & \num{0.525} & Y \\*
  &   50 & \num{1.96} & \num{0.419} & \num{0.123} & \num{0.337} & Y \\*
  &  100 & \num{1.96} & \num{0.278} & \num{0.101} & \num{0.199} & Y \\*
  &  200 & \num{1.96} & \num{0.178} & \num{0.142} & \num{0.0684} & Y \\*
  &  500 & \num{1.96} & \num{0.129} & \num{0.126} & \num{0.00424} & N \\*
  & 1000 & \num{1.96} & \num{0.106} & \num{0.0519} & \num{0.0562} & Y \\
\midrule
\multirow{6}{*}{CSS}
  &   20 & \num{0.0371} & \num{0.212} & \num{0.262} & \num{-0.0538} & Y \\*
  &   50 & \num{0.0371} & \num{0.237} & \num{0.219} & \num{-0.00167} & N \\*
  &  100 & \num{0.0371} & \num{0.137} & \num{0.151} & \num{-0.00683} & N \\*
  &  200 & \num{0.0371} & \num{0.0901} & \num{0.0849} & \num{0.00135} & N \\*
  &  500 & \num{0.0371} & \num{0.0673} & \num{0.0458} & \num{0.0201} & Y \\*
  & 1000 & \num{0.0371} & \num{0.0577} & \num{0.0446} & \num{0.0126} & N \\
\midrule
\multirow{6}{*}{CWA}
  &   20 & \num{1.08} & \num{0.491} & \num{0.523} & \num{-0.0118} & N \\*
  &   50 & \num{1.08} & \num{0.144} & \num{0.152} & \num{0.00659} & N \\*
  &  100 & \num{1.08} & \num{0.1} & \num{0.0901} & \num{-0.000419} & N \\*
  &  200 & \num{1.08} & \num{0.0779} & \num{0.0853} & \num{-0.00565} & N \\*
  &  500 & \num{1.08} & \num{0.0605} & \num{0.0454} & \num{0.0124} & N \\*
  & 1000 & \num{1.08} & \num{0.0274} & \num{0.0352} & \num{-0.00562} & N \\
\midrule
\multirow{6}{*}{SGL}
  &   20 & \num{2.29} & \num{84.9} & \num{50.7} & \num{31.3} & Y \\*
  &   50 & \num{2.29} & \num{65.2} & \num{64.4} & \num{14.9} & N \\*
  &  100 & \num{2.29} & \num{79.0} & \num{51.8} & \num{23.2} & Y \\*
  &  200 & \num{2.29} & \num{58.9} & \num{23.9} & \num{34.4} & Y \\*
  &  500 & \num{2.29} & \num{39.6} & \num{15.9} & \num{24.3} & Y \\*
  & 1000 & \num{2.29} & \num{43.5} & \num{28.1} & \num{9.06} & Y \\
\end{longtable}

\FloatBarrier
\section{Empirical Conditional-Independence Preservation}
\label{sec:appendix_ci_preservation}
We test whether the synthetic data preserve the conditional-independence (CI) structure of the real data. For each training split we identify the CI statements that are not rejected on the real data, test the same statements on the matched synthetic data, and report the fraction that remains non-rejected. We use the same conditional-independence test as the discovery pipeline for each dataset. We evaluate \num{100} paired repetitions per dataset and training size and compare conditions with the paper's paired protocol (Wilcoxon signed-rank with Pratt ties, Holm correction).

Under observational generation, preservation is high for every condition. Median fractions range from \num{0.75} to \num{1.00} across datasets and training sizes, and \SI{82}{\percent} of combinations are at or above \num{0.89}. The paired differences between \gls{dag}-aware and vanilla generation are small (absolute Hodges--Lehmann effects below \num{0.05}) and show no consistent direction.

Under interventional generation, causal conditioning matters (\Cref{tab:ci_preservation_interventional}). \Gls{dag}-aware generation preserves significantly more empirical CIs than vanilla \gls{tabpfn} in \num{4} of \num{6} training sizes on \gls{csm} (median Hodges--Lehmann effect $+0.11$) and \num{5} of \num{6} on CSMn2 ($+0.23$), and it is never significantly lower on either. CSMn2 is the custom \gls{scm} with $\sigma = 10^{-2}$ evaluated in \Cref{sec:appendix_noise}. \Gls{opdag} follows the same pattern (\num{4} and \num{5} of \num{6}), while discovered \gls{cpdag} does so in only \num{0} and \num{1} of \num{6}, reflecting discovery quality rather than the \gls{pdag}-based conditioning rule. On \gls{cms}, both \gls{dag}-aware and vanilla generation preserve almost all CIs, with no significant difference between them. On the remaining CSuite datasets vanilla and \gls{dag}-aware generation are essentially at ceiling, and the few significant differences involving \gls{dag}-aware generation are small (absolute Hodges--Lehmann effects below \num{0.011}). They all occur at large training sizes ($N \geq 200$). Two of them fall on \gls{clb}, where the \gls{ate} improvement of \gls{dag}-aware generation in the same interventional runs is also not significant at those sizes, reflecting the same large-sample convergence of vanilla \gls{tabpfn}. The diagnostic confirms the paper's central message that under interventional generation causal-structure-aware conditioning reliably preserves the empirical independence structure of the real data.

\begin{longtable}{l r r r r r}
\caption{Median fraction of empirical conditional independencies preserved under interventional generation, over \num{100} paired repetitions per combination. Reference statements are the CIs not rejected on the real training data, using for each dataset the same conditional-independence test as the discovery pipeline. The same statements are then tested on the synthetic data. * marks medians whose paired difference from vanilla original is significant (Wilcoxon-Pratt with Holm correction within each dataset--size family). A starred median below the corresponding vanilla value indicates a significant decrease relative to vanilla. When the displayed medians coincide, the direction of the difference is not visible at this precision.}
\label{tab:ci_preservation_interventional}\\
\toprule
Dataset & {$N$} & {Vanilla} & {\gls{dag}-aware} & {\Gls{opdag}} & {Disc.\ \gls{cpdag}} \\
\midrule
\endfirsthead
\multicolumn{6}{l}{\itshape Table \thetable\ (continued)}\\
\toprule
Dataset & {$N$} & {Vanilla} & {\gls{dag}-aware} & {\Gls{opdag}} & {Disc.\ \gls{cpdag}} \\
\midrule
\endhead
\bottomrule
\endlastfoot
\multirow{6}{*}{CSM}
  &   20 & 0.91 & 0.91 & 0.91 & 0.91 \\*
  &   50 & 0.89 & 0.89 & 0.89\textsuperscript{*} & 0.82 \\*
  &  100 & 0.78 & 0.89\textsuperscript{*} & 0.89\textsuperscript{*} & 0.78 \\*
  &  200 & 0.78 & 0.89\textsuperscript{*} & 0.89\textsuperscript{*} & 0.78 \\*
  &  500 & 0.67 & 0.89\textsuperscript{*} & 0.67 & 0.67 \\*
  & 1000 & 0.63 & 0.75\textsuperscript{*} & 0.75\textsuperscript{*} & 0.59 \\
\midrule
\multirow{6}{*}{CSMn2}
  &   20 & 0.91 & 0.91 & 0.91 & 0.91 \\*
  &   50 & 0.67 & 0.89\textsuperscript{*} & 0.89\textsuperscript{*} & 0.73\textsuperscript{*} \\*
  &  100 & 0.67 & 0.89\textsuperscript{*} & 0.89\textsuperscript{*} & 0.67 \\*
  &  200 & 0.67 & 0.89\textsuperscript{*} & 0.89\textsuperscript{*} & 0.67 \\*
  &  500 & 0.67 & 0.89\textsuperscript{*} & 0.89\textsuperscript{*} & 0.67 \\*
  & 1000 & 0.67 & 1.00\textsuperscript{*} & 1.00\textsuperscript{*} & 0.67 \\
\midrule
\multirow{6}{*}{CLB}
  &   20 & 1.00 & 1.00 & 1.00 & 1.00 \\*
  &   50 & 0.97 & 0.97 & 0.98 & 0.97 \\*
  &  100 & 0.96 & 0.96 & 0.96 & 0.96 \\*
  &  200 & 0.97 & 0.96\textsuperscript{*} & 0.98\textsuperscript{*} & 0.97 \\*
  &  500 & 0.99 & 0.98\textsuperscript{*} & 0.99\textsuperscript{*} & 0.98\textsuperscript{*} \\*
  & 1000 & 0.99 & 0.99 & 0.99 & 0.96\textsuperscript{*} \\
\midrule
\multirow{6}{*}{CMC}
  &   20 & 1.00 & 1.00 & 1.00 & 1.00 \\*
  &   50 & 1.00 & 1.00 & 1.00 & 1.00 \\*
  &  100 & 1.00 & 1.00 & 1.00 & 1.00 \\*
  &  200 & 1.00 & 1.00 & 1.00 & 1.00 \\*
  &  500 & 1.00 & 0.99\textsuperscript{*} & 1.00\textsuperscript{*} & 1.00\textsuperscript{*} \\*
  & 1000 & 0.99 & 0.99\textsuperscript{*} & 0.99\textsuperscript{*} & 0.99\textsuperscript{*} \\
\midrule
\multirow{6}{*}{CMS}
  &   20 & 1.00 & 1.00 & 1.00 & 1.00 \\*
  &   50 & 1.00 & 1.00 & 1.00 & 1.00 \\*
  &  100 & 1.00 & 0.91 & 0.91 & 1.00 \\*
  &  200 & 0.89 & 0.88 & 0.88 & 0.85 \\*
  &  500 & 0.80 & 0.78 & 0.80 & 0.67\textsuperscript{*} \\*
  & 1000 & 1.00 & 1.00 & 1.00 & 0.50\textsuperscript{*} \\
\midrule
\multirow{6}{*}{CNS}
  &   20 & 0.95 & 0.96 & 0.95 & 0.95 \\*
  &   50 & 1.00 & 1.00 & 1.00 & 1.00 \\*
  &  100 & 1.00 & 1.00 & 1.00 & 0.88 \\*
  &  200 & 1.00 & 1.00\textsuperscript{*} & 1.00 & 1.00 \\*
  &  500 & 1.00 & 1.00 & 1.00 & 1.00 \\*
  & 1000 & 1.00 & 1.00 & 1.00 & 0.83\textsuperscript{*} \\
\midrule
\multirow{6}{*}{CSS}
  &   20 & 1.00 & 1.00 & 1.00 & 1.00 \\*
  &   50 & 1.00 & 1.00 & 1.00 & 1.00 \\*
  &  100 & 0.93 & 0.90 & 0.93 & 0.93 \\*
  &  200 & 1.00 & 1.00 & 1.00 & 1.00 \\*
  &  500 & 1.00 & 1.00 & 1.00 & 1.00 \\*
  & 1000 & 1.00 & 1.00 & 1.00 & 1.00 \\
\midrule
\multirow{6}{*}{CWA}
  &   20 & 1.00 & 1.00 & 1.00 & 1.00 \\*
  &   50 & 0.96 & 0.96 & 0.96 & 0.96 \\*
  &  100 & 0.95 & 0.95 & 0.95 & 0.96\textsuperscript{*} \\*
  &  200 & 0.95 & 0.94 & 0.97\textsuperscript{*} & 0.95 \\*
  &  500 & 0.95 & 0.95 & 0.98\textsuperscript{*} & 0.95 \\*
  & 1000 & 0.97 & 0.96\textsuperscript{*} & 0.97 & 0.94\textsuperscript{*} \\
\end{longtable}

\FloatBarrier
\section{Robustness to Higher Noise}
\label{sec:appendix_noise}

To verify that the findings reported in the main text are not specific to the near-deterministic regime ($\sigma = 10^{-5}$), we repeat our experiments on the custom \gls{scm} with $\sigma = 10^{-2}$ (denoted CSMn2).

\subsection*{Synthetic Data Quality}

\begin{figure}[!htbp]
  \centering
  \begin{adjustbox}{max width=\linewidth, max totalheight=0.48\textheight}
    \includegraphics{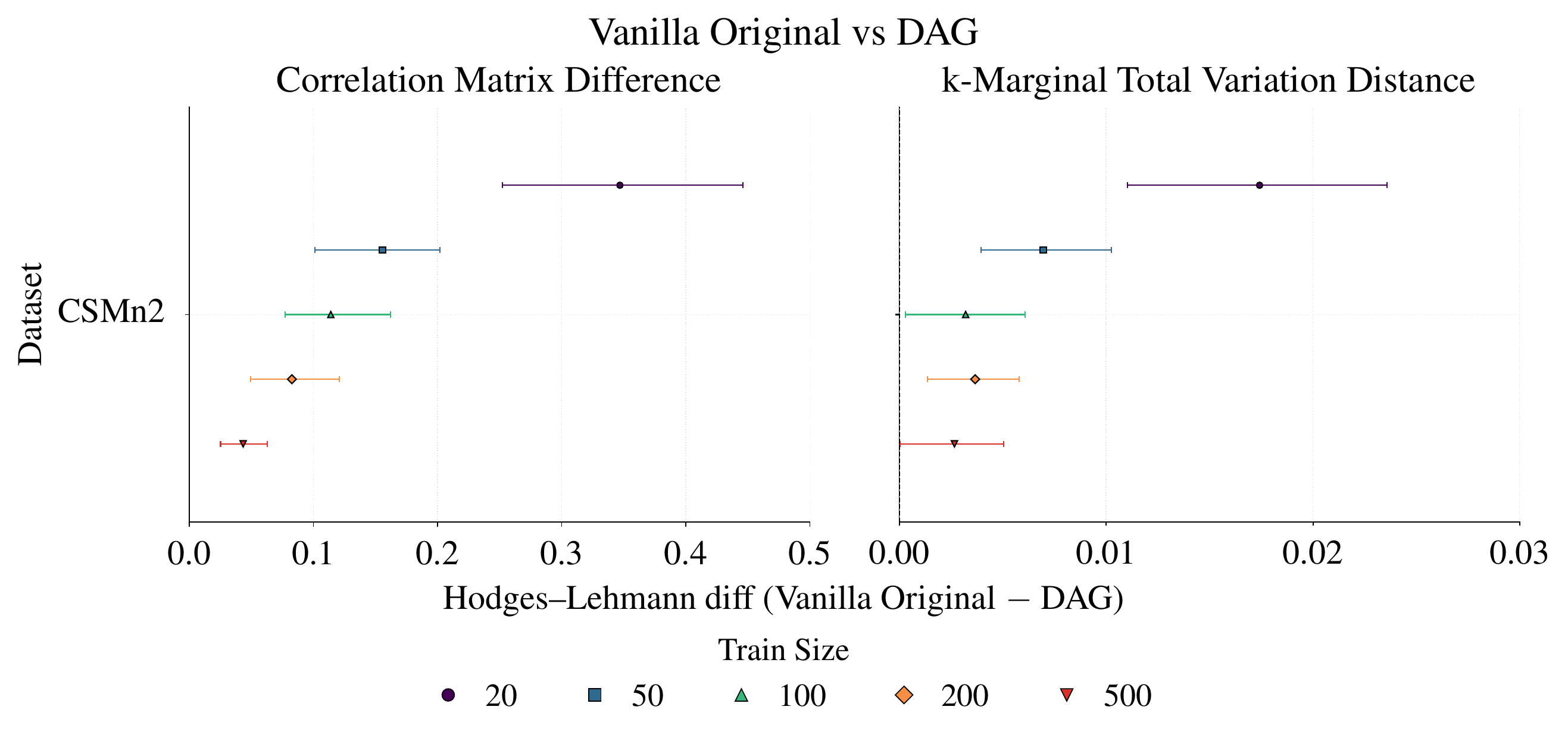}
  \end{adjustbox}
  \caption{Hodges--Lehmann estimates comparing \gls{dag}-aware generation and vanilla \gls{tabpfn} with original ordering on the custom \gls{scm} with $\sigma = 10^{-2}$, in \gls{cmd} (left) and \gls{kmtvd} ($k=2$, right).
  Positive values indicate that \gls{dag}-aware generation achieves lower metric values (i.e., better synthetic data quality). Filled markers with solid error bars indicate significance at $p<0.05$ (Holm correction).}
  \label{fig:forest_noise_dag_combined}
\end{figure}

\begin{figure}[!htbp]
  \centering
  \includegraphics[width=\linewidth, height=0.48\textheight, keepaspectratio]{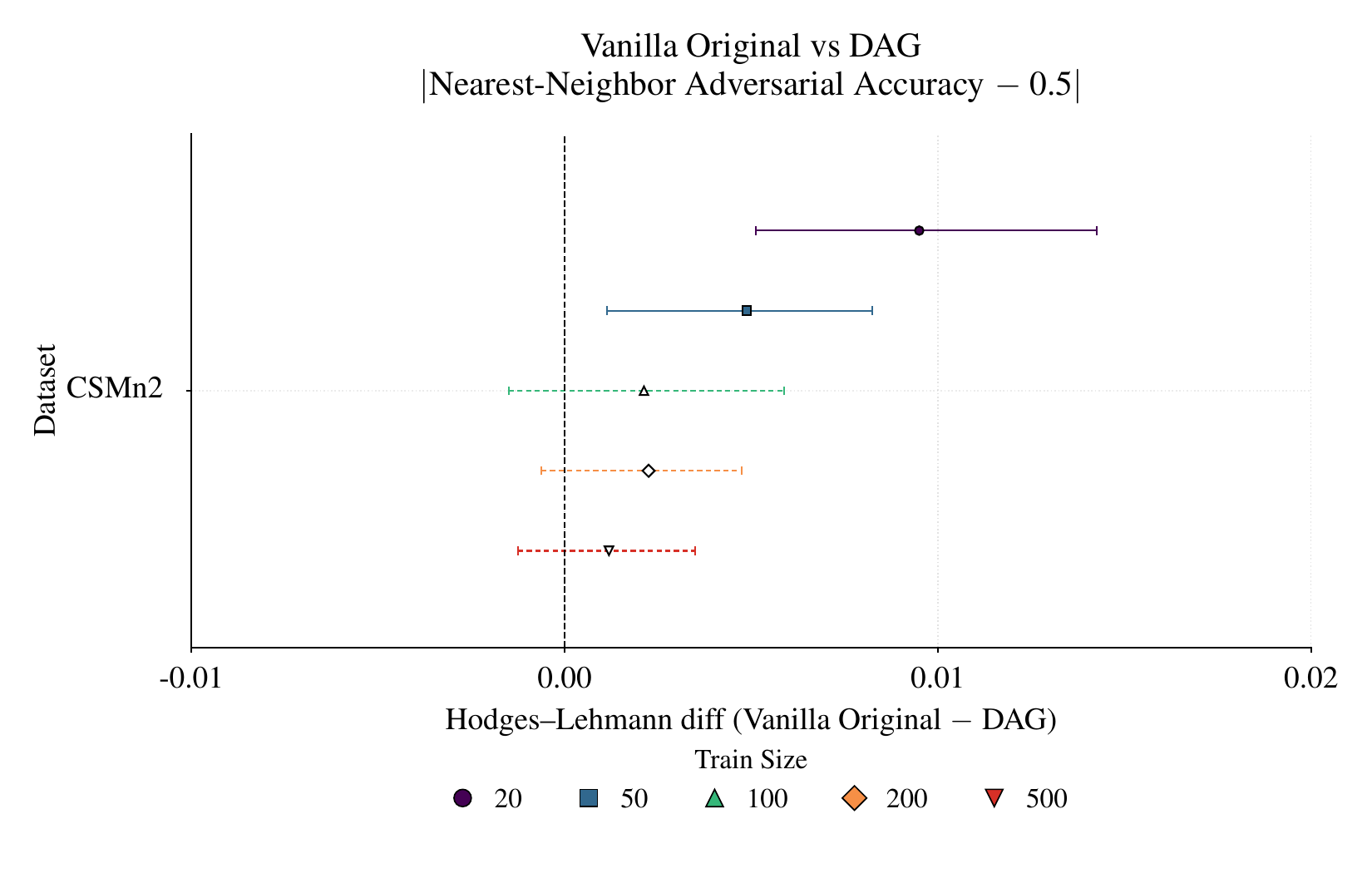}
  \caption{Hodges--Lehmann estimates comparing \gls{dag}-aware generation and vanilla \gls{tabpfn} on the custom \gls{scm} with $\sigma = 10^{-2}$, in \gls{nnaa} (reported as the distance from the ideal value, $\lvert\mathrm{NNAA} - 0.5\rvert$).
    Positive values indicate that \gls{dag}-aware generation achieves lower values (i.e., greater indistinguishability between synthetic and real data).
    Filled markers with solid error bars indicate significance at $p<0.05$ (Holm correction).}
  \label{fig:forest_noise_dag_nnaa}
\end{figure}

\gls{dag}-aware generation shows significant improvements over vanilla \gls{tabpfn} across all training sizes in both \gls{cmd} and \gls{kmtvd} (\Cref{fig:forest_noise_dag_combined}). In \gls{nnaa}, improvements are significant at $N = 20$ and $N = 50$ but not at larger training sizes (\Cref{fig:forest_noise_dag_nnaa}), consistent with the $\sigma = 10^{-5}$ results.

\begin{figure}[!htbp]
  \centering
  \begin{adjustbox}{max width=\linewidth, max totalheight=0.48\textheight}
    \includegraphics{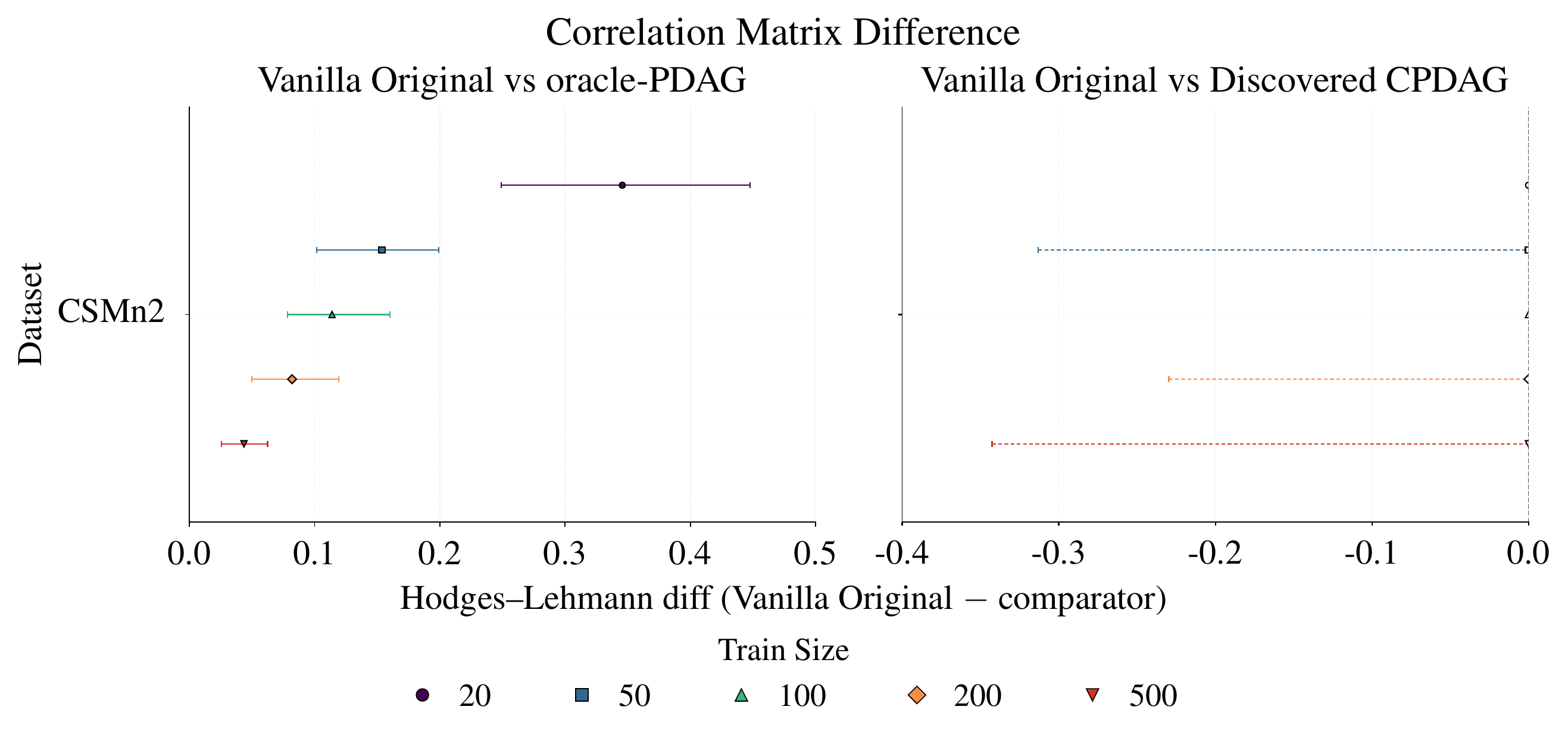}
  \end{adjustbox}
  \caption{Hodges--Lehmann estimates comparing \gls{pdag}-based generation and vanilla \gls{tabpfn} on the custom \gls{scm} with $\sigma = 10^{-2}$, in \gls{cmd}, with \gls{opdag} (left) and discovered \gls{cpdag} (right).
  Positive values indicate that \gls{pdag}-based generation achieves lower metric values (i.e., better synthetic data quality).
  Filled markers with solid error bars indicate significance at $p<0.05$ (Holm correction).}
  \label{fig:forest_noise_cpdag_combined_cmd}
\end{figure}

\begin{figure}[!htbp]
  \centering
  \begin{adjustbox}{max width=\linewidth, max totalheight=0.48\textheight}
    \includegraphics{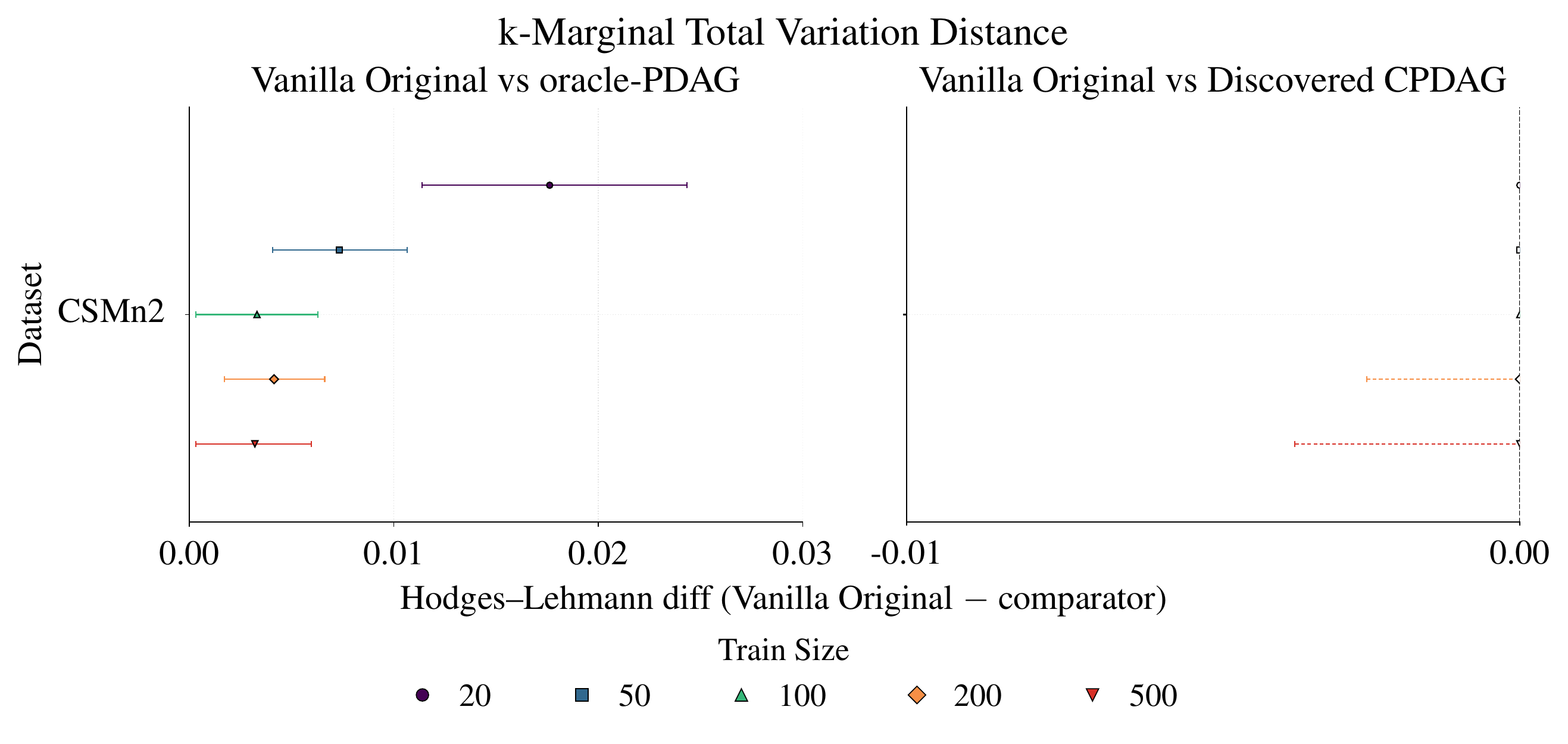}
  \end{adjustbox}
  \caption{Hodges--Lehmann estimates comparing \gls{pdag}-based generation and vanilla \gls{tabpfn} with original ordering on the custom \gls{scm} with $\sigma = 10^{-2}$, in \gls{kmtvd} ($k=2$), with \gls{opdag} (left) and discovered \gls{cpdag} (right).
  Positive values indicate that \gls{pdag}-based generation achieves lower metric values (i.e., better synthetic data quality).
  Filled markers with solid error bars indicate significance at $p<0.05$ (Holm correction).}
  \label{fig:forest_noise_cpdag_combined_2marg}
\end{figure}

\begin{figure}[!htbp]
  \centering
  \begin{adjustbox}{max width=\linewidth, max totalheight=0.48\textheight}
    \includegraphics{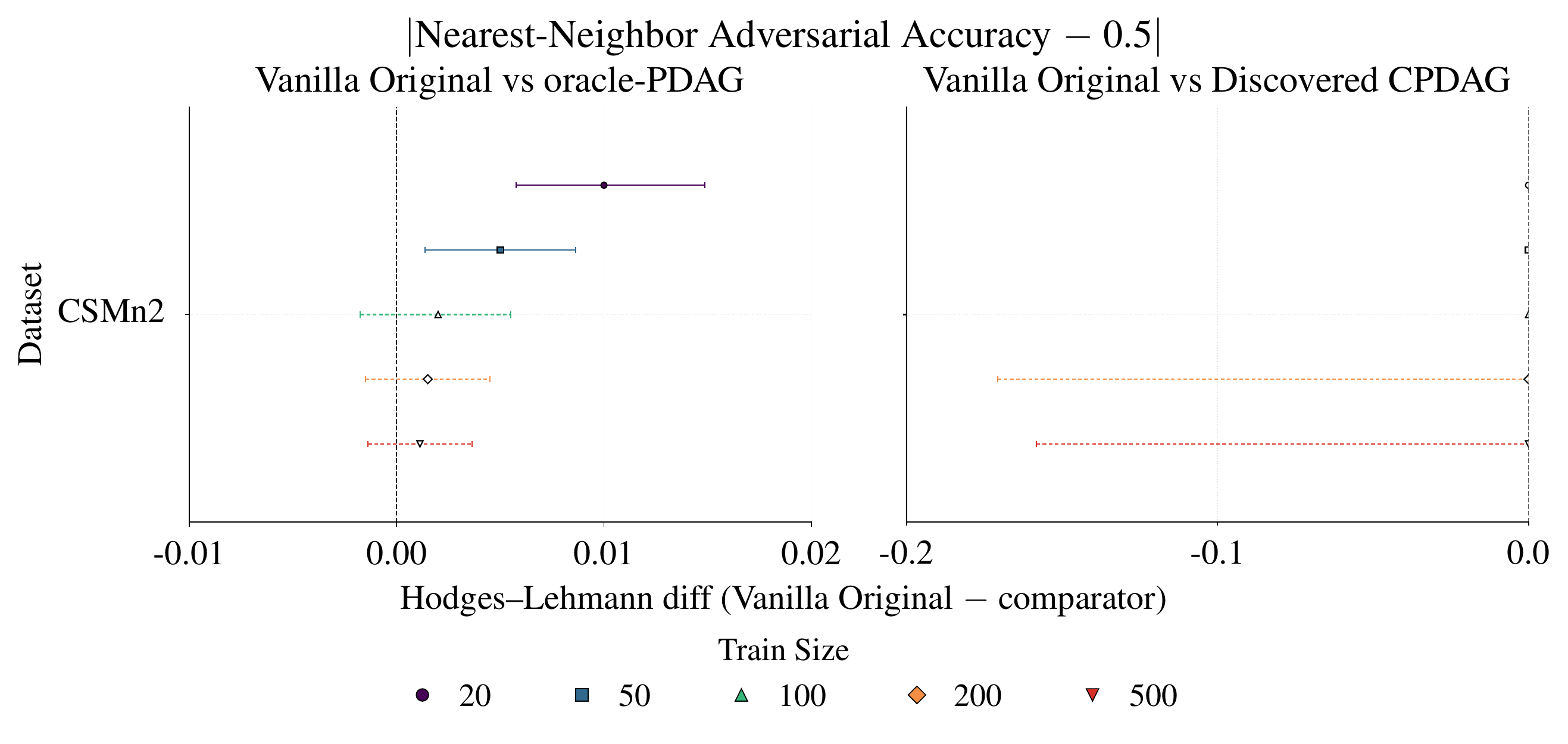}
  \end{adjustbox}
  \caption{Hodges--Lehmann estimates comparing \gls{pdag}-based generation and vanilla \gls{tabpfn} on the custom \gls{scm} with $\sigma = 10^{-2}$, in \gls{nnaa} (reported as the distance from the ideal value, $\lvert\mathrm{NNAA} - 0.5\rvert$), with \gls{opdag} (left) and discovered \gls{cpdag} (right).
  Positive values indicate that \gls{pdag}-based generation achieves lower values (i.e., greater indistinguishability between synthetic and real data).
  Filled markers with solid error bars indicate significance at $p<0.05$ (Holm correction).}
  \label{fig:forest_noise_cpdag_combined_nnaa}
\end{figure}

\Gls{opdag} shows the same pattern: significant improvements in \gls{cmd} and \gls{kmtvd} across all training sizes (\Cref{fig:forest_noise_cpdag_combined_cmd,fig:forest_noise_cpdag_combined_2marg}), with \gls{nnaa} significant only at small $N$ (\Cref{fig:forest_noise_cpdag_combined_nnaa}). 
Discovered \gls{cpdag} shows no meaningful differences across metrics and training sizes, consistent with the $\sigma = 10^{-5}$ results.

These results confirm that the benefits of causal conditioning extend beyond the near-deterministic regime.

\Cref{tab:spurious_correlations_noise} confirms that spurious correlations under $\sigma = 10^{-2}$ are comparable in magnitude to those observed under $\sigma = 10^{-5}$ (\Cref{tab:spurious_correlations,tab:spurious_correlations_main}). \Gls{opdag} shows reduced correlations under higher noise, suggesting increased robustness to imprecise conditioning from undirected edges.

\begin{table}[!htbp]
\centering
\caption{Mean Pearson correlation coefficients for independent variable pairs in the custom collider \gls{scm} with $\sigma = 10^{-2}$.
Values close to zero indicate correct independence preservation; the best value per column and training size is in \textbf{bold}.
Standard deviations across \num{100} repetitions in parentheses.}
\label{tab:spurious_correlations_noise}
\begin{tabular}{l
  S[table-format=-1.3,detect-weight,detect-family] @{\hspace{1mm}} l
  S[table-format=-1.3,detect-weight,detect-family] @{\hspace{1mm}} l}
\toprule
Method & {$\rho(X_0, X_3)$} & & {$\rho(X_0, X_2)$} & \\
\midrule
\multicolumn{5}{l}{\textit{Train size $N = 20$}} \\
Vanilla original & -0.147 & (0.205) & -0.145 & (0.206) \\
Vanilla topological & -0.015 & (0.116) & -0.014 & (0.115) \\
Vanilla reverse top. & -0.121 & (0.202) & -0.119 & (0.204) \\
\gls{dag}-aware & 0.004 & (0.021) & \bfseries 0.003 & (0.021) \\
\Gls{opdag} & \bfseries 0.003 & (0.023) & 0.004 & (0.021) \\
Discovered \gls{cpdag} & -0.136 & (0.215) & -0.134 & (0.216) \\
\midrule
\multicolumn{5}{l}{\textit{Train size $N = 50$}} \\
Vanilla original & -0.026 & (0.144) & -0.026 & (0.144) \\
Vanilla topological & 0.005 & (0.075) & 0.006 & (0.076) \\
Vanilla reverse top. & -0.033 & (0.148) & -0.034 & (0.146) \\
\gls{dag}-aware & \bfseries 0.000 & (0.023) & \bfseries 0.000 & (0.024) \\
\Gls{opdag} & -0.001 & (0.024) & \bfseries 0.000 & (0.023) \\
Discovered \gls{cpdag} & 0.005 & (0.194) & 0.005 & (0.193) \\
\midrule
\multicolumn{5}{l}{\textit{Train size $N = 100$}} \\
Vanilla original & -0.043 & (0.113) & -0.043 & (0.113) \\
Vanilla topological & -0.017 & (0.064) & -0.017 & (0.063) \\
Vanilla reverse top. & -0.049 & (0.102) & -0.047 & (0.102) \\
\gls{dag}-aware & \bfseries 0.000 & (0.020) & 0.001 & (0.020) \\
\Gls{opdag} & -0.002 & (0.025) & \bfseries 0.000 & (0.020) \\
Discovered \gls{cpdag} & -0.022 & (0.143) & -0.022 & (0.145) \\
\midrule
\multicolumn{5}{l}{\textit{Train size $N = 200$}} \\
Vanilla original & -0.029 & (0.099) & -0.034 & (0.080) \\
Vanilla topological & -0.012 & (0.042) & -0.012 & (0.042) \\
Vanilla reverse top. & -0.035 & (0.082) & -0.028 & (0.109) \\
\gls{dag}-aware & \bfseries -0.002 & (0.025) & \bfseries -0.001 & (0.026) \\
\Gls{opdag} & -0.003 & (0.029) & -0.002 & (0.025) \\
Discovered \gls{cpdag} & 0.019 & (0.178) & 0.017 & (0.175) \\
\midrule
\multicolumn{5}{l}{\textit{Train size $N = 500$}} \\
Vanilla original & -0.027 & (0.043) & -0.027 & (0.042) \\
Vanilla topological & -0.004 & (0.025) & -0.003 & (0.025) \\
Vanilla reverse top. & -0.033 & (0.039) & -0.032 & (0.039) \\
\gls{dag}-aware & \bfseries 0.000 & (0.023) & 0.001 & (0.023) \\
\Gls{opdag} & \bfseries 0.000 & (0.023) & \bfseries 0.000 & (0.023) \\
Discovered \gls{cpdag} & 0.016 & (0.146) & 0.019 & (0.155) \\
\midrule
Test set & 0.022 & {--} & 0.022 & {--} \\
\bottomrule
\end{tabular}
\end{table}

\subsection*{Treatment Effect Preservation}

\begin{figure}[!htbp]
  \centering
  \includegraphics[width=\linewidth, height=0.48\textheight, 
    keepaspectratio]{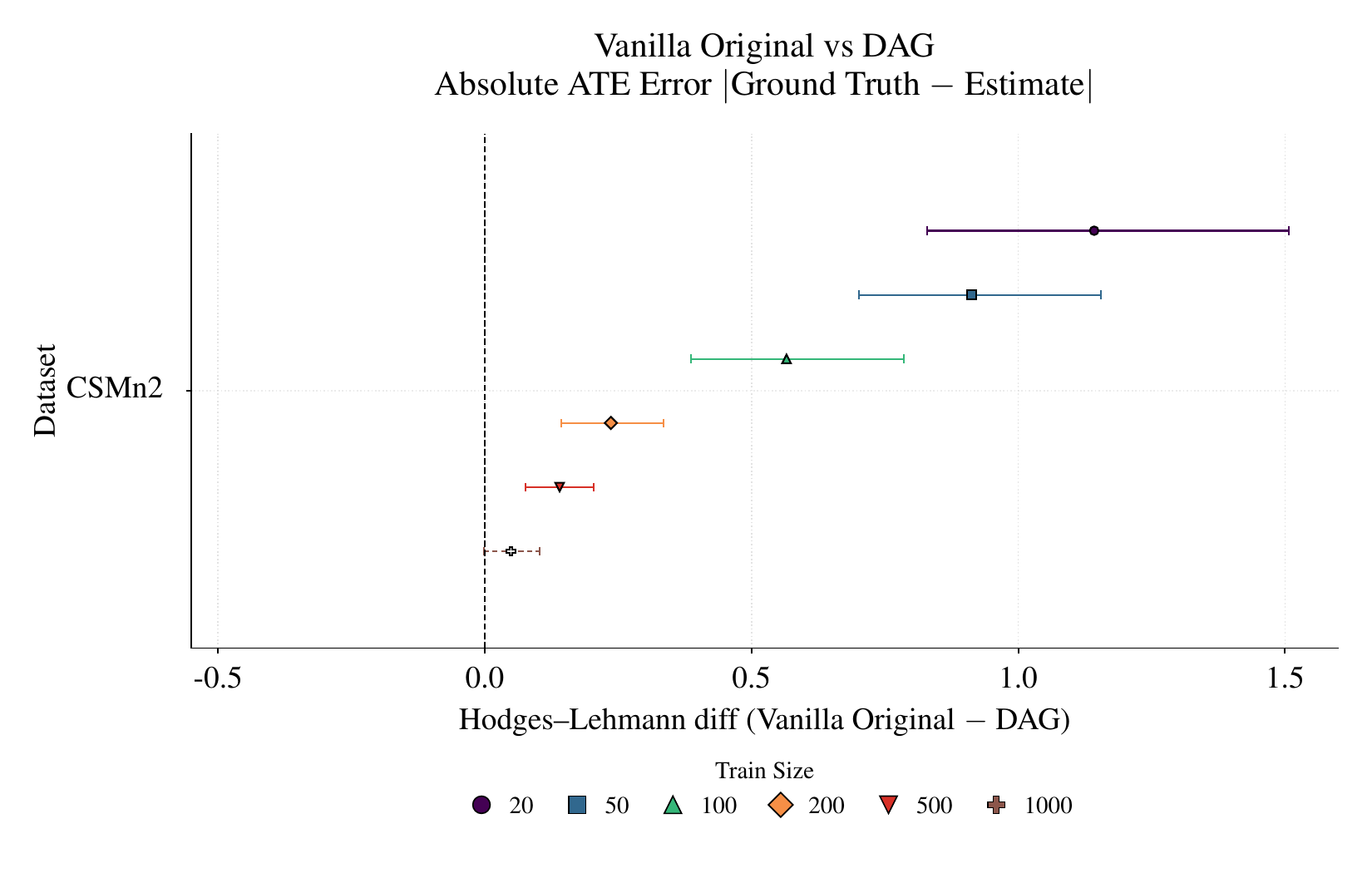}
  \caption{Hodges--Lehmann estimates of the reduction in absolute \gls{ate} 
  error ($\Delta_{\text{ATE}}$) when comparing vanilla \gls{tabpfn} with 
  original ordering versus \gls{dag}-aware generation on the custom \gls{scm} 
  with $\sigma = 10^{-2}$.
  Positive values indicate smaller errors (closer to ground truth) for \gls{dag}-aware generation; negative values indicate larger errors.
  Filled markers with solid error bars indicate significance at $p<0.05$ (Holm correction).}
  \label{fig:forest_noise_dag_ate}
\end{figure}

\begin{figure}[!htbp]
  \centering
  \begin{adjustbox}{max width=\linewidth, max totalheight=0.48\textheight}
    \includegraphics{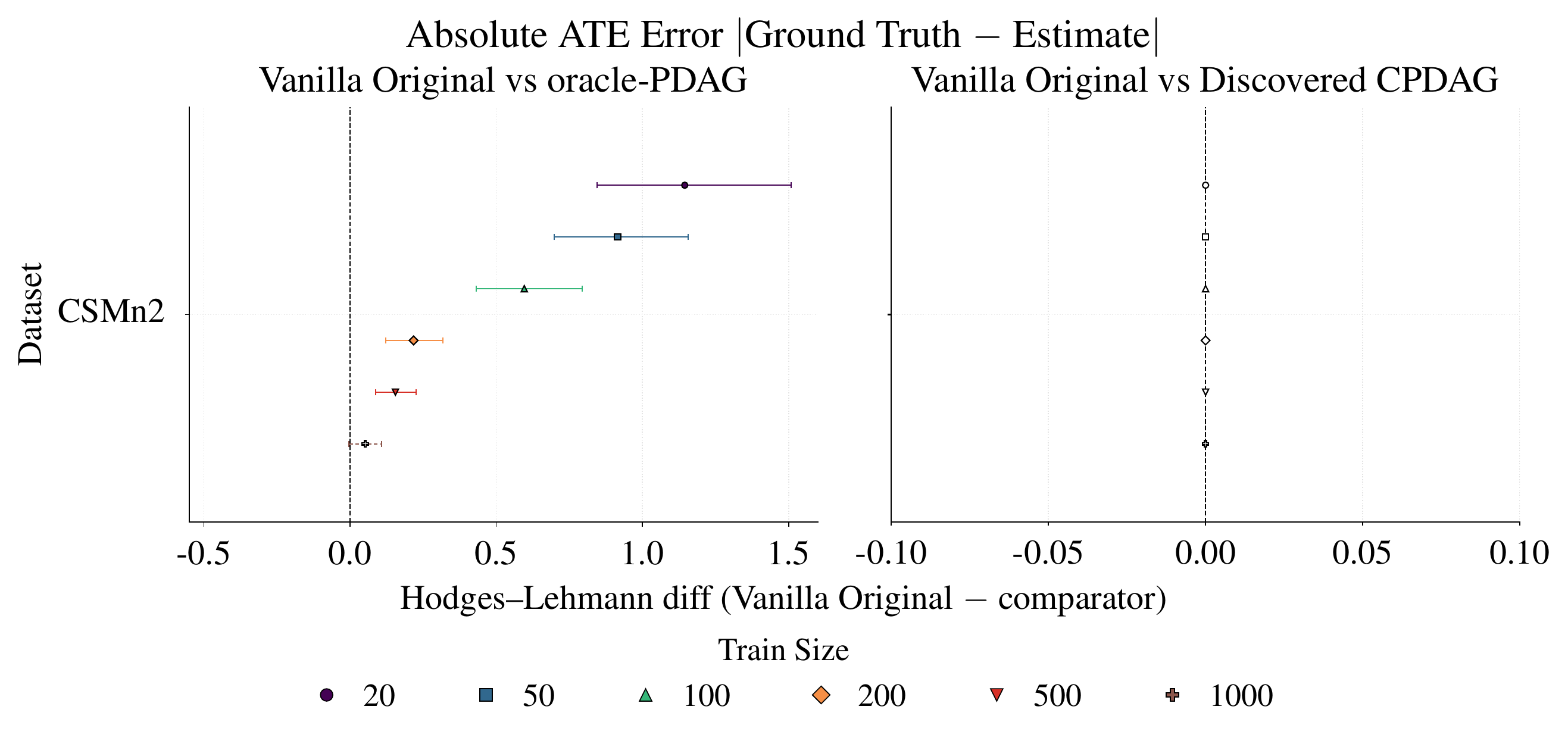}
  \end{adjustbox}
  \caption{Hodges--Lehmann estimates of the reduction in absolute \gls{ate} 
  error ($\Delta_{\text{ATE}}$) when comparing vanilla \gls{tabpfn} with 
  original ordering versus \gls{pdag}-based generation on the custom 
  \gls{scm} with $\sigma = 10^{-2}$, with \gls{opdag} (left) and 
  discovered \gls{cpdag} (right).
Positive values indicate smaller errors (closer to ground truth) for the respective method (\gls{opdag} or discovered \gls{cpdag}); negative values indicate larger errors.
Filled markers with solid error bars indicate significance at $p<0.05$ (Holm correction).}
  \label{fig:forest_noise_cpdag_ate}
\end{figure}

\gls{dag}-aware generation significantly improves \gls{ate} preservation across $N \in \{20, 50, 100, 200, 500\}$ compared to vanilla \gls{tabpfn} (\Cref{fig:forest_noise_dag_ate}). At $N = 1000$, the direction remains favorable but not statistically significant.
\Gls{opdag} shows the same pattern (\Cref{fig:forest_noise_cpdag_ate}). 
Discovered \gls{cpdag} shows no differences across training sizes.

\section{Causal Discovery with ReX}
\label{sec:appendix_rex}
To assess the sensitivity of our approach to the choice of causal discovery algorithm, we tested ReX~\citep{renero2026rex}, a method that employs machine learning regressors for skeleton discovery and exploits additive noise model assumptions to orient edges, on the four CSuite datasets that satisfy these assumptions (CLB, CNS, CSS, CWA). We ran ReX with default hyperparameters, as preliminary tests showed no meaningful difference with tuning at these sample sizes.
Given the consistent degradations observed across all distributional metrics, we did not extend the evaluation to ATE preservation.

\Cref{tab:rex_graph_quality} reports graph quality metrics averaged over \num{100} repetitions. ReX exhibits low direction precision at small and medium training sizes, particularly on CLB, CNS, and CWA (below \SI{50}{\percent} for $N \leq 100$). Graph quality improves at $N = 500$ for CLB (\num{0.96} skeleton recall, \num{0.95} direction precision) and CSS (\num{1.00} direction precision, \num{0.60} recall). CNS achieves high direction precision at $N = 500$ but skeleton recall remains at \num{0.50}. CWA maintains \num{0.75} direction precision and \num{0.57} recall even at $N = 500$.

\begin{figure}[!htbp]
  \centering
  \begin{adjustbox}{max width=\linewidth, max totalheight=0.48\textheight}
    \includegraphics{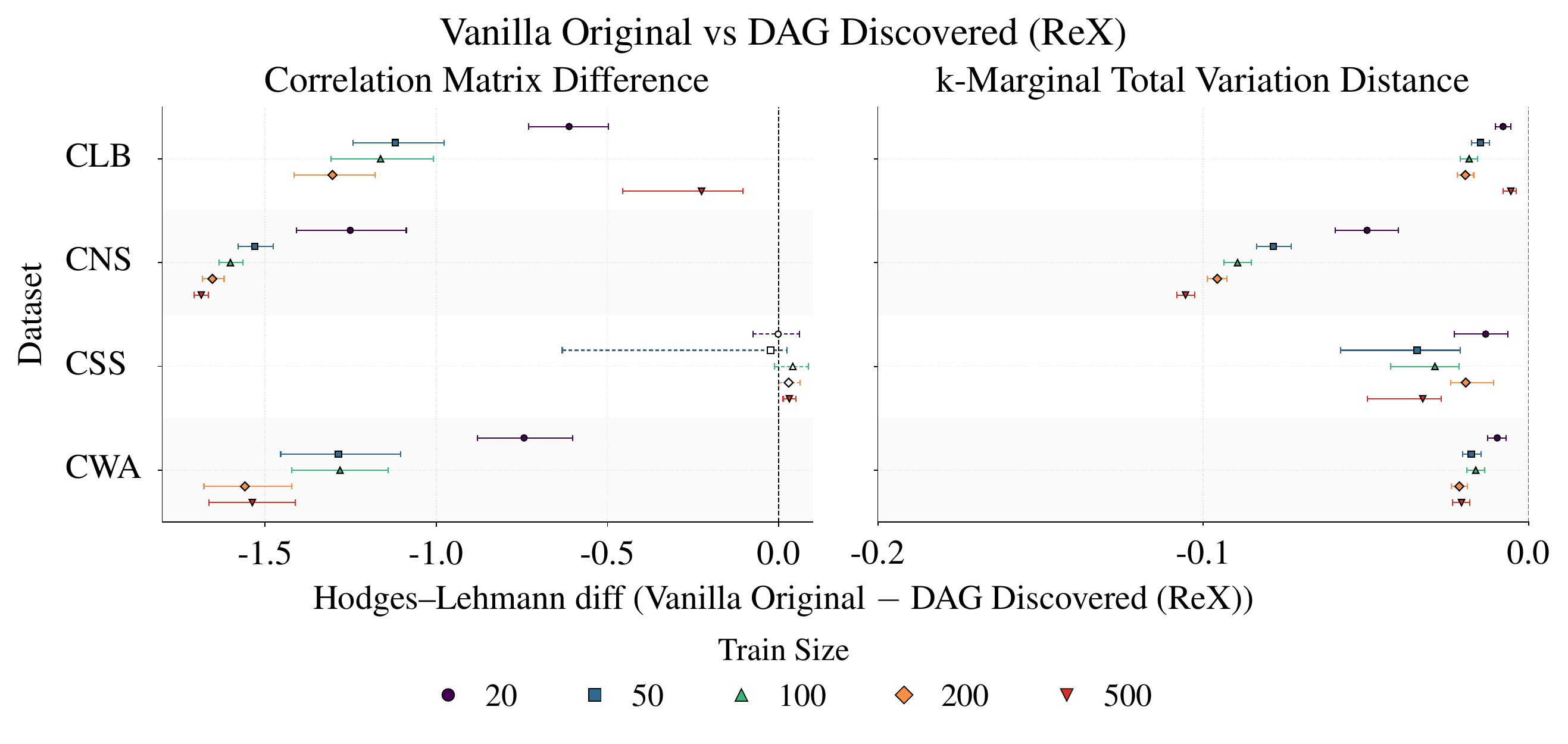}
  \end{adjustbox}
  \caption{Hodges--Lehmann estimates comparing \gls{dag}-aware generation with ReX-discovered graphs and vanilla \gls{tabpfn} with original ordering, in \gls{cmd} (left) and \gls{kmtvd} ($k=2$, right).
  Positive values indicate that the ReX-based approach achieves lower metric values (i.e., better synthetic data quality).
  Filled markers with solid error bars indicate significance at $p<0.05$ (Holm correction).}
  \label{fig:forest_rex_dag_combined}
\end{figure}

\begin{figure}[!htbp]
  \centering
  \includegraphics[width=\linewidth, height=0.48\textheight, 
    keepaspectratio]{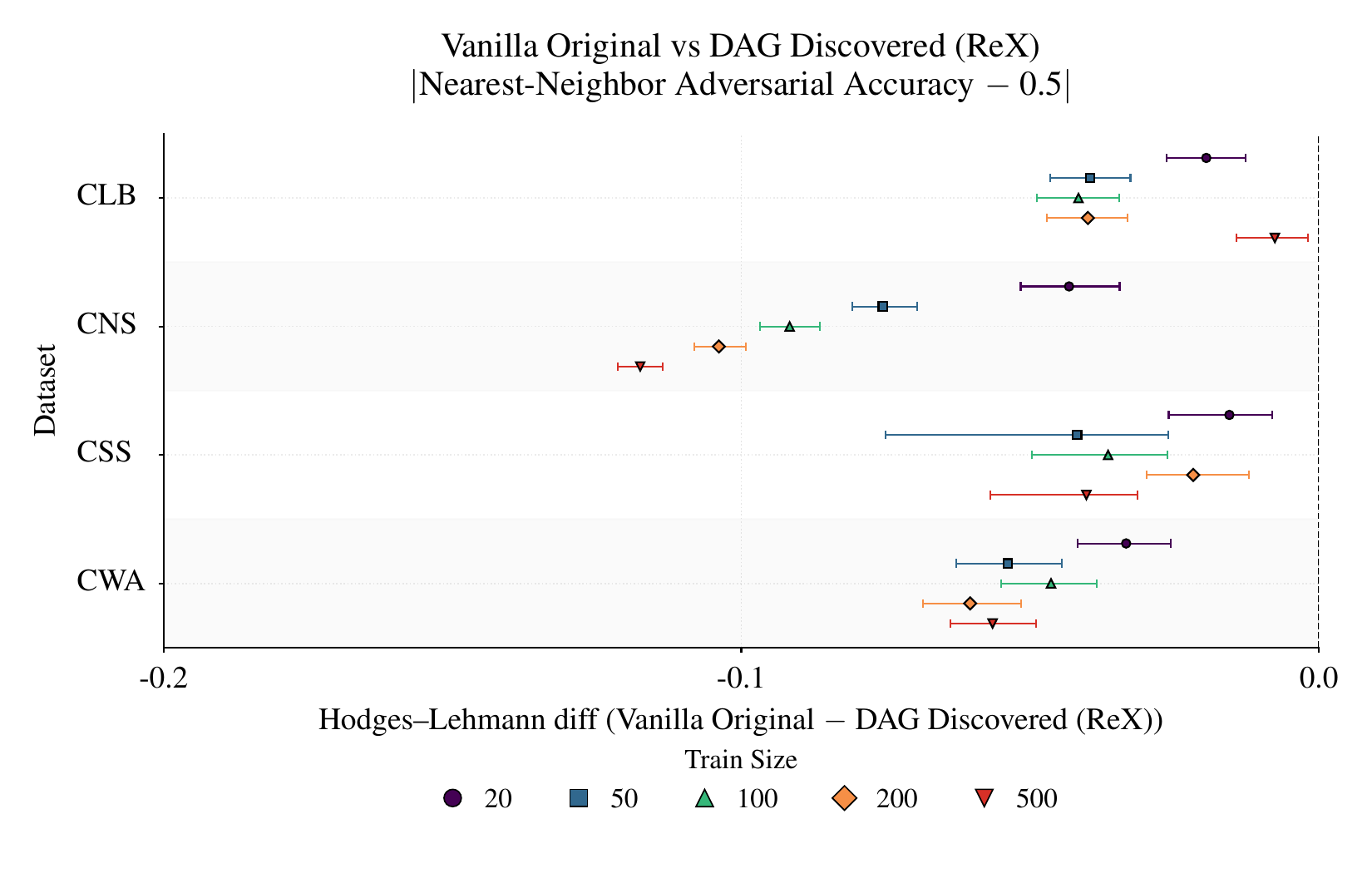}
  \caption{Hodges--Lehmann estimates comparing \gls{dag}-aware generation with ReX-discovered graphs and vanilla \gls{tabpfn} with original ordering, in \gls{nnaa} (reported as the distance from the ideal value, $\lvert\mathrm{NNAA} - 0.5\rvert$).
  Positive values indicate that the ReX-based approach achieves lower values (i.e., greater indistinguishability between synthetic and real data).
  Filled markers with solid error bars indicate significance at $p<0.05$ (Holm correction).}
  \label{fig:forest_rex_dag_nnaa}
\end{figure}

Using ReX-discovered DAGs for causal conditioning leads to significant degradations compared to vanilla \gls{tabpfn} across nearly all dataset--training size combinations: \num{15} significant degradations and one significant improvement (CSS at $N = 500$, where ReX reaches \num{1.00} direction precision) in \gls{cmd}, and \num{20} significant degradations with no improvements in both \gls{kmtvd} and \gls{nnaa} (\Cref{fig:forest_rex_dag_combined,fig:forest_rex_dag_nnaa}).

These results confirm that the quality of the discovered causal graph is critical: incorrect graphs actively harm synthetic data quality. This motivates the use of conservative algorithms such as PC combined with our hybrid conditioning strategy, which falls back to vanilla sequential generation for undirected edges rather than committing to potentially wrong directions.

\begin{table}[!htbp]
\centering
\caption{ReX graph quality on CSuite datasets, averaged over \num{100} repetitions. Since ReX returns fully oriented DAGs, all discovered edges are directed. Skeleton recall measures the fraction of true edges recovered. Direction precision measures the fraction of correctly oriented edges among recovered true-skeleton edges.}
\label{tab:rex_graph_quality}
\begin{tabular}{l r S[table-format=1.2] S[table-format=1.2]}
\toprule
Dataset & {$N$} & {Skel.\ recall} & {Dir.\ prec.} \\
\midrule
\multirow{5}{*}{CLB} 
  &  20 & 0.70 & 0.47 \\
  &  50 & 0.73 & 0.45 \\
  & 100 & 0.86 & 0.47 \\
  & 200 & 0.91 & 0.62 \\
  & 500 & 0.96 & 0.95 \\
\midrule
\multirow{5}{*}{CNS} 
  &  20 & 0.57 & 0.36 \\
  &  50 & 0.53 & 0.36 \\
  & 100 & 0.51 & 0.32 \\
  & 200 & 0.51 & 0.68 \\
  & 500 & 0.50 & 0.99 \\
\midrule
\multirow{5}{*}{CSS} 
  &  20 & 0.51 & 0.57 \\
  &  50 & 0.47 & 0.65 \\
  & 100 & 0.51 & 0.88 \\
  & 200 & 0.63 & 0.99 \\
  & 500 & 0.60 & 1.00 \\
\midrule
\multirow{5}{*}{CWA} 
  &  20 & 0.56 & 0.44 \\
  &  50 & 0.55 & 0.41 \\
  & 100 & 0.62 & 0.39 \\
  & 200 & 0.58 & 0.51 \\
  & 500 & 0.57 & 0.75 \\
\bottomrule
\end{tabular}
\end{table}

